\documentclass{article}

\usepackage[main, final]{neurips_2026}
\usepackage{xcolor}         % colors
\usepackage[table]{xcolor}
\usepackage{array}
\usepackage{arydshln}
\usepackage{multirow}
\usepackage{makecell}
\usepackage{algorithm}
\usepackage{algorithmic}
\usepackage{amsmath}
\usepackage{enumitem}
\usepackage{graphicx}
\usepackage{pifont}
\usepackage{float}
\usepackage{wrapfig}
\usepackage{url}            % simple URL typesetting
\usepackage{booktabs}       % professional-quality tables
\usepackage{amsfonts}       % blackboard math symbols
\usepackage{nicefrac}       % compact symbols for 1/2, etc.
\usepackage{microtype}      % microtypography
\usepackage{multirow}
\usepackage{makecell}
\usepackage{diagbox} 
\usepackage{caption}

\usepackage{xspace}

\newcommand{\method}{\textsc{I2V-Detach}\xspace}
\newcommand{\svloss}{source-video decoupling}
\newcommand{\vvloss}{intra-video coupling}

\usepackage[utf8]{inputenc} % allow utf-8 input
\usepackage[T1]{fontenc}    % use 8-bit T1 fonts
\usepackage{hyperref}       % hyperlinks
\usepackage{url}            % simple URL typesetting
\usepackage{booktabs}       % professional-quality tables
\usepackage{amsfonts}       % blackboard math symbols
\usepackage{nicefrac}       % compact symbols for 1/2, etc.
\usepackage{microtype}      % microtypography
\usepackage{xcolor}         % colors
\usepackage{xspace}
\usepackage{xpatch}
\title{\method: Source Grounding Detachment for Unauthorized Image-to-Video Generation}

\newcommand{\eg}{\emph{e.g.}\xspace}

\newcommand{\ie}{\emph{i.e.}\xspace}

\newcommand{\notocsection}[1]{%
  {\let\addcontentsline\gobblethree\section{#1}}
}

\newcommand{\jeany}[1]{\textcolor{blue}{{#1}}}
\newcommand{\chan}[1]{\textcolor{red}{{#1}}}
\newcommand{\note}[1]{\textcolor{orange}{{#1}}}
\author{%
  \textbf{Chanhui Lee}$^{1}$ \quad
  \textbf{Yeonghwan Song}$^{2}$ \quad
  \textbf{Yewon Kang}$^{1}$ \quad
  \textbf{Jeany Son}$^{1}$ \\[2pt]
  $^{1}$POSTECH \quad $^{2}$GIST \\[2pt]
  \texttt{\{chanlee, yewon.kang, jeany\}@postech.ac.kr} \ \ \ \ \
  \texttt{yeonghwan.song@gm.gist.ac.kr}
}

\begin{document}

\maketitle
\addtocontents{toc}{\protect\setcounter{tocdepth}{0}}
% \vspace{-0.7cm}
\begin{figure*}[h]
\centering
    \includegraphics[width=\linewidth]{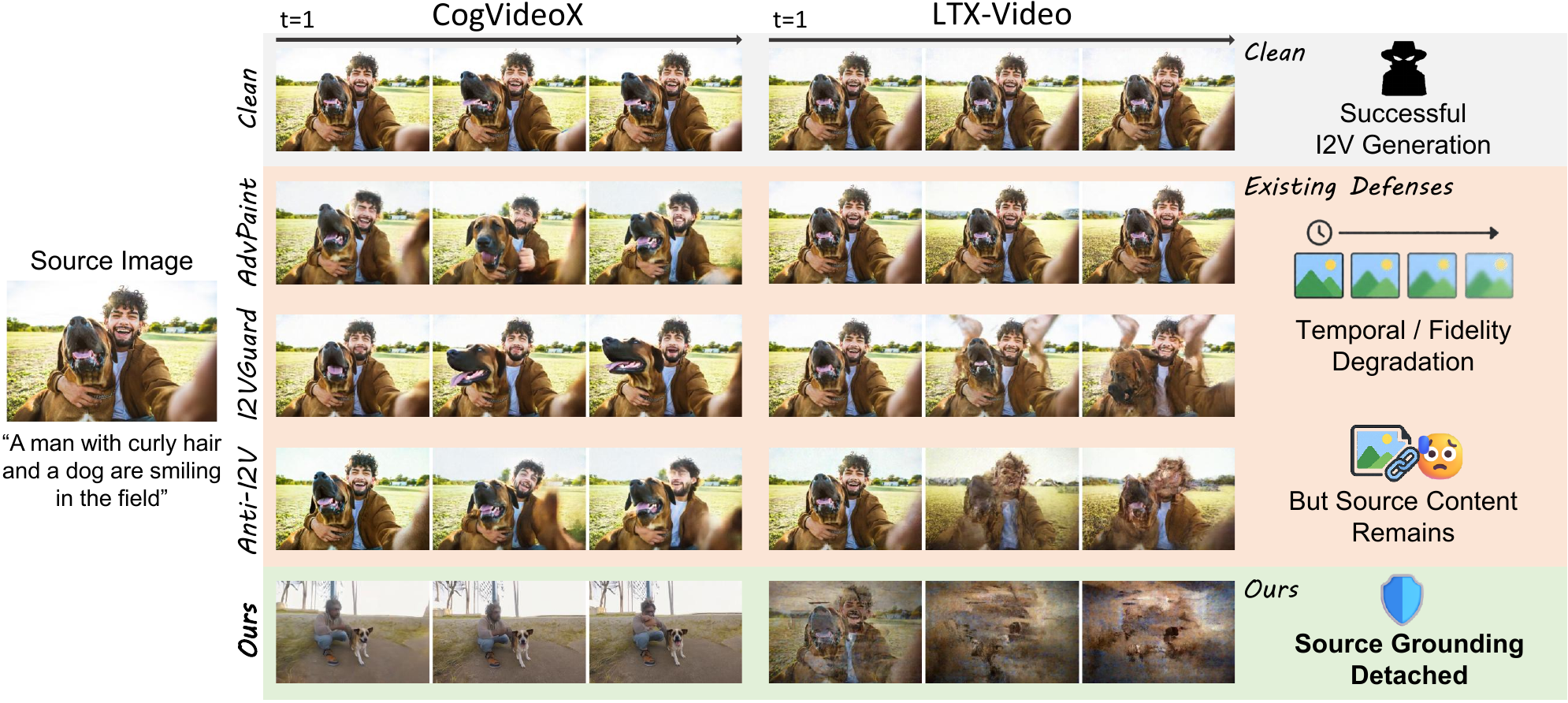}
    \vspace{-0.3cm}
    \caption{Goal of \method against diffusion-based I2V generation. Unlike prior baselines that often result in temporal degradation or fidelity reduction, \emph{source grounding detachment} objective directly drives video generation that no longer faithfully preserves visual content of source-image.}
    \vspace{-0.2cm}
    \label{fig/teaser}
\end{figure*}

\begin{abstract}
Recent diffusion models have advanced image-conditioned video generation, but their accessibility raises risks of unauthorized manipulation.
Image immunization has emerged as a promising defense for Image-to-Video (I2V) protection, yet existing methods often rely on degrading visual quality or temporal coherence.
In this work, we revisit I2V immunization as reducing source-content preservation, shifting the focus from output degradation to weakening source-conditioned content propagation.
To this end, we propose \method, an image immunization method that prevents unauthorized I2V generation through source grounding detachment.
I2V models preserve source content by propagating source-image cues into video latents during denoising.
Accordingly, \method disrupts this process in two complementary ways: it suppresses source-video grounding to reduce source cue propagation, and strengthens intra-video interactions to shift denoising toward video-side dynamics with reduced reliance on the source image.
To obtain informative optimization signals, we initialize auxiliary video latents with source-oriented states that remain within the source-conditioned generation regime while deviating from source-preserving trajectories.
Extensive experiments across diverse diffusion architectures show that \method substantially reduces source-content preservation, improves over prior immunization methods, and exhibits strong black-box transferability.

\end{abstract}
\section{Introduction}

Diffusion models~\cite{ddpm, diffusionsurvey} have enabled high-fidelity visual content manipulation under natural-language guidance, spanning image synthesis~\cite{stable, esser2024scaling, flux, flux-2} and video generation~\cite{cogvideo,cogvideox,hunda,wan,opensora,ltx, shin2026trimotion}.
In particular, image-conditioned video generation allows users to preserve and animate the visually grounded content of a reference image, but also increases the risk of unauthorized image use, including portrait misuse, privacy violations, and malicious visual manipulation~\cite{deepfake1, deepfake2, deepfake3}.
Protecting private images from such misuse has therefore become an urgent problem.

Image immunization provides a proactive defense by adding imperceptible perturbations to source images before they are used by generative models. 
Early studies mainly focused on image manipulation tasks such as image editing~\cite{raising, semantic, blurguard, dct, advcross}, inpainting~\cite{diffvax, diffguard, advpaint}, and instruction-guided image editing~\cite{editshield, myface}.
I2VGuard~\cite{i2vguard} contrastively aligns intermediate representations with a target video, VidFreeze~\cite{vidfreeze} suppresses motion synthesis by manipulating attention dynamics, and Anti-I2V~\cite{antii2v} enhances robustness against purification by optimizing perturbations jointly in the CIE Lab color space and the frequency domain.

While effective, existing defenses often overlook a key property of I2V generation: videos are generated under strong reference conditioning, where the source image serves as a persistent and dominant reference throughout denoising~\cite{identifying, ditctrl, sparse}.
% Under this strong source dependency, perturbing intermediate representations or attention dynamics can disrupt generation, but does not necessarily target the source-to-video grounding pathway that carries protected source cues into video latents.
As a result, prior defenses often manifest as perceptual degradation, temporal artifacts, or unstable generation, while the generated video can still retain recognizable source identity, structure, and appearance, as shown in Figure~\ref{fig/teaser}.
% }
%
% \jeany{
% We therefore revisit image immunization for I2V generation as source-grounding attenuation.
% We therefore revisit image immunization for I2V generation as reducing source-conditioned content preservation.
We therefore shift the goal of I2V immunization from degrading generated videos to reducing the preservation of protected source content.
% Rather than merely degrading visual quality or temporal consistency, effective immunization should weaken the source-conditioned pathway that carries protected source content into the generated video.
Importantly, this does not require eliminating all source-related cues; instead, the goal is to reduce source preservation rather than treating generation degradation itself as the primary objective.
To this end, we propose \method, a novel image immunization method for diffusion-based I2V models based on \emph{source grounding detachment}.
\method directly targets the source-video interactions that act as the main pathway for propagating source-specific cues into video latents during denoising.
Specifically, it learns an imperceptible perturbation that weakens the grounding between the protected source image and source-conditioned video latents, thereby reducing the generated video's dependence on the protected source.
% }

% \jeany{
\method optimizes two complementary objectives for source grounding detachment.
A \textbf{\emph{source-video decoupling loss}} suppresses source-to-video interactions, while an \textbf{\emph{intra-video coupling loss}} strengthens interactions among video latents, increasing the degree to which video latents evolve through video-side interactions rather than being anchored to the source image.
% A \emph{source-video decoupling loss} suppresses source-to-video interactions, while an \emph{intra-video coupling loss} strengthens interactions among video latents, encouraging denoising to rely less on source-dominated conditioning and more on video-side interactions.
% A \emph{source-video decoupling loss} suppresses source-to-video interactions, while an \emph{intra-video coupling loss} strengthens interactions among video latents, shifting denoising from source-dominated conditioning toward video-side dynamics.
% A \emph{source-video decoupling loss} suppresses source-to-video interactions, while an \emph{intra-video coupling loss} amplifies interactions among video latents, reducing the role of the source image as a persistent conditioning anchor and allowing video latents to evolve with greater freedom in their temporal dynamics.
To provide informative optimization signals, we further introduce source-divergent video initialization, where auxiliary video latents are drawn from the source-conditioned generation process rather than random initialization.
These latents remain source-related while deviating from source-preserving trajectories, enabling effective detachment without merely destabilizing generation.
As a result, \method makes generated videos less strongly grounded in the protected source, as shown in Figure~\ref{fig/teaser}.

We perform extensive experiments to compare our method with existing I2I immunization methods~\cite{editshield, advpaint}, as well as I2V immunization methods~\cite{antii2v, i2vguard} whose objectives are closely related to ours, across diffusion-based I2V models with diverse architectures~\cite{cogvideox, ltx, opensora}.
Furthermore, we conduct comprehensive evaluations from multiple perspectives, including quantitative metrics and human evaluation, to validate the effectiveness of the proposed method.
The results demonstrate the superiority of our method compared with existing I2V immunization baselines, which we attribute to explicitly tailoring source-image grounding disruption to the distinctive characteristics of I2V generation.

The main contributions can be summarized as follows:
% \jeany{
% \begin{itemize}
%     % \item We identify a key limitation of existing I2V image immunization methods: perceptual or temporal degradation alone does not necessarily prevent generated videos from preserving protected source content under strong reference conditioning.
%     % \item We revisit image immunization for diffusion-based I2V generation from the perspective of \emph{source-content non-preservation}. 
%     \item We revisit I2V image immunization as \emph{source grounding detachment}, and propose \method that directly weakens source-to-video interactions to prevent faithful propagation of protected source cues during denoising.
%     \item We propose source-video decoupling and intra-video coupling objectives, together with source-oriented video latent initialization, to explicitly target the source-conditioned pathway through which I2V models propagate source cues into video latents, enabling effective source grounding detachment.
%     \item Extensive experiments demonstrate that our method consistently achieves stronger image immunization than existing I2V methods and adapted I2I baselines across diverse diffusion-based I2V models. 
%     Source-related consistency metrics and human evaluation further validate the effectiveness and generalizability of our visual grounding detachment approach.
% \end{itemize}
\begin{itemize}
    % \item We revisit I2V immunization from the perspective of source-conditioned content preservation, highlighting that generation degradation and source-content reduction are not necessarily equivalent under strong reference conditioning.
    % \item We shift the focus of I2V immunization from degrading generated videos to reducing source-conditioned content preservation, highlighting that protected source content can remain recognizable even when generation quality or temporal dynamics are degraded.
    \item We reformulate I2V immunization as reducing source-conditioned content preservation, showing that protected source content can remain recognizable even when output quality or temporal dynamics are degraded.
    \item We propose \method, a source-grounding detachment framework that directly intervenes in the source-to-video pathway of I2V denoising by separating source-video grounding from intra-video interactions.
    \item We introduce source-video decoupling and intra-video coupling objectives with source-oriented video latent initialization, enabling targeted suppression of source cue retrieval while promoting video-side dynamics.
    \item We demonstrate consistent source-preservation reduction across diverse I2V architectures, adapted I2I baselines, and I2V immunization methods, with strong robustness and black-box transferability.
\end{itemize}
\section{Related Works}

\paragraph{Image-to-Video Diffusion Models.}
Modern image-to-video (I2V) generation has largely evolved by augmenting text-to-image diffusion models with temporal modeling modules, enabling the synthesis of temporally coherent videos from a single image.
Early methods such as Tune-A-Video~\cite{tav} and AnimateDiff~\cite{anidiff} showed that even lightweight temporal extensions can produce convincing image animation.
Stable Video Diffusion~\cite{svd} further advanced latent video diffusion at scale and has since become a widely adopted open-source benchmark.
More recent models have improved generation quality and controllability through more powerful spatio-temporal architectures, including Lumiere~\cite{lumiere} and diffusion transformer-based models such as CogVideoX~\cite{cogvideox}.
At the same time, LTX-Video~\cite{ltx} emphasizes efficient generation, while Wan~\cite{wan} lowers the barrier to high-quality I2V synthesis as a large-scale open-source model.
Community-led projects such as VideoCrafter2~\cite{videocrafter2} and Open-Sora~\cite{opensora} have further democratized access to I2V technology.
As I2V models become increasingly realistic and accessible, protecting source images from misuse in video synthesis has emerged as an important problem.
\vspace{-0.2cm}

\paragraph{Image Immunization against Diffusion Models.}
Adversarial attack-based image immunization has emerged as an active research direction for protecting images against unauthorized use and malicious generative manipulation.
In image-to-image pipelines, prior works have demonstrated strong protection performance by perturbing VAE encoder latents~\cite{raising, blurguard, dct, editshield}, disrupting cross-attention mechanisms~\cite{semantic, advcross}, or exploiting denoising outputs~\cite{raising, diffguard, diffvax} to break faithful image editing.
With the rapid progress of image-to-video generation, recent attention has naturally shifted toward defending images from misuse in video synthesis.
Among the first works to explicitly identify this problem, I2VGuard~\cite{i2vguard} induced temporal degradation by forcing encoder-level latents, attention outputs, and denoised latents to resemble those of a fixed target video.
% \chan{
VidFreeze~\cite{vidfreeze} primarily promotes temporal freezing by imposing temporally static attention patterns.
Anti-I2V~\cite{antii2v} extended this line of work by encouraging internal attention features to collapse into uninformative representations.
While these approaches provide effective protection, generated videos can still preserve the identity and visual content of the source image due to the source-anchored prior of I2V models~\cite{identifying, ditctrl, sparse}.
In contrast, our method explicitly separates \emph{two attention pathways} to suppress source-video grounding and reinforce video-side interactions, directly breaking this prior.
% While these approaches provide effective protection, they do not fully account for the source-dominant nature of I2V generation, often causing fidelity reduction or temporal degradation while leaving source-image content reflected in the outputs.
% }
 
\section{Motivation: Source Grounding in I2V Diffusion Models}

Diffusion-based I2V models generate videos by denoising video latents conditioned on a source image and a text prompt.
Let $Z_{\tau}$ denote the noisy video latents at timestep $\tau$, $c$ the source-image conditioning states, and $t$ the text embedding.
At each denoising step, the model predicts
\begin{equation}
    \hat{\epsilon}_{\tau}
    =
    \epsilon_{\theta}(Z_{\tau}, \tau, t, c).
\end{equation}
Since source conditioning is repeatedly injected throughout the denoising process, the source image serves as a persistent visual reference in I2V generation.
Consequently, generated videos can remain strongly grounded in the source image even when perturbations mainly induce temporal instability or perceptual degradation.

Although different I2V architectures implement source conditioning in different ways, source information is commonly propagated through attention interactions between video latent states and source-conditioning states.
We denote the source-conditioning tokens by $\mathcal{C}$ and the video tokens by $\mathcal{V}$.
For query and key features $Q$ and $K$, we define the pre-softmax attention logit function $S$ as
\begin{equation}
    S(Q,K)
    =
    \frac{QK^{\top}}{\sqrt{d_k}}, 
    \quad 
    [S(Q,K)]_{I,J} =
    \bigl[S(Q,K)_{ij}\bigr]_{i\in I,\,j\in J}.
    % S_{\mathcal{I},\mathcal{J}}=\left[S_{i,j}\right]_{i\in\mathcal{I},\,j\in\mathcal{J}}.
    \label{eq:logit}
\end{equation}
Here, rows correspond to query tokens and columns correspond to key tokens.
For index sets \(\mathcal I\) and \(\mathcal J\), 
\([S(Q,K)]_{\mathcal I,\mathcal J}\) denotes the submatrix obtained by selecting query rows in \(\mathcal I\) and key columns in \(\mathcal J\).
Under this notation, \([S(Q,K)]_{\mathcal V,\mathcal C}\) measures how strongly video queries retrieve source-conditioning keys.
We refer to this interaction as the source-retrieval grounding pathway, which serves as the main source-to-video grounding mechanism.
In contrast, \([S(Q,K)]_{\mathcal V,\mathcal V}\) captures interactions among video tokens themselves, corresponding to the video-side interaction pathway.
% Here, rows correspond to query tokens and columns correspond to key tokens; 
% $S_{\mathcal{I},\mathcal{J}}$ denotes the submatrix of $S$ obtained by selecting query rows in $\mathcal{I}$ and key columns in $\mathcal{J}$.
% % where rows correspond to query tokens and columns correspond to key tokens, and the submatrix of $S$ obtained by selecting query rows in \(\mathcal{I}\) and key columns in \(\mathcal{J}\).
% Under this notation, $S_{\mathcal{V},\mathcal{C}}$ measures how strongly video tokens retrieve information from source-conditioning tokens.
% We refer to this interaction as the source-to-video grounding pathway.
% In contrast, $S_{\mathcal{V},\mathcal{V}}$ captures interactions among video tokens themselves, corresponding to the video-side interaction pathway.

Figure~\ref{fig/attention}(a) visualizes these token-level attention structures.
The blue rectangle indicates the source-conditioning region, where video queries assign salient attention to source-conditioning keys, suggesting that the source image remains an important reference during denoising.
The red rectangle indicates the video-side region, where the strong diagonal structure reflects organized interactions among video tokens.
These observations motivate our source grounding detachment objective, which suppresses direct source retrieval and promotes video-side interactions, making the denoising process less dominated by source conditioning.
\vspace{-0.2cm}
\begin{figure*}[t]
\centering
    \includegraphics[width=\linewidth]{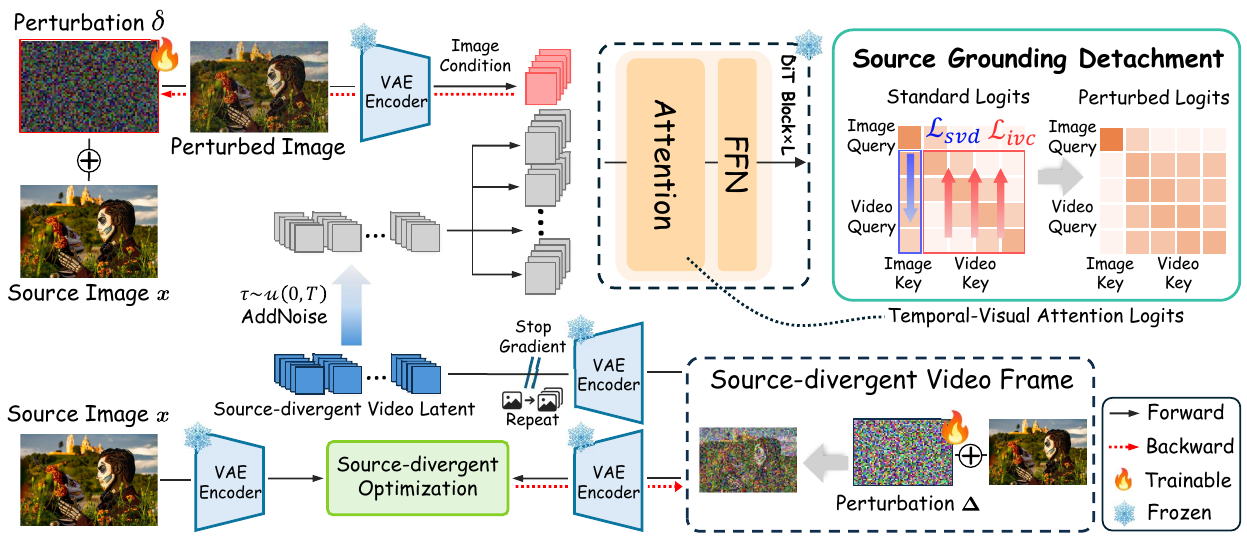}
    \vspace{-0.2cm}
    \caption{Illustration of the framework of \method.
Source grounding detachment is achieved by two complementary losses; \emph{{\svloss}} and \emph{{\vvloss}}.
Jointly optimized with progressively source-divergent video frames, \method drives denoising video latents away from source-bound constraints, disrupting source-content preservation in generated videos.
}

    \label{fig/framework}
\end{figure*}

\section{\method}
\label{sec:03_method}
\vspace{-0.2cm}
In this section, we present \method, an image immunization framework that reduces faithful source-content preservation by detaching the generated video from the protected source image. 
Instead of inducing generic temporal artifacts or reducing visual fidelity, our goal is to weaken the attention-mediated pathway through which source-image cues are repeatedly propagated into video latents during denoising.

% \chan{
% In this section, we present \method, which aims to prevent source-image visual content from being grounded into the generated video. %during diffusion-based I2V generation.
% Our method explicitly intervenes in the attention-mediated conditioning, directly disrupting the dominant image--video anchor that governs denoising video latents in I2V models.
% }

\subsection{Overview}
\label{sec:overview}
\vspace{-0.2cm}
Given a source image $x$ and a frozen I2V diffusion denoiser $\epsilon_\theta$, I2V-DETACH learns an imperceptible perturbation $\delta$ under an $\ell_\infty$ budget $\eta$. 
The resulting immunized image $x_{\mathrm{adv}}=x+\delta^\ast$ is optimized by minimizing a source grounding detachment loss over denoising timesteps:
\begin{equation}
\delta^\ast
=
\arg\min_{\|\delta\|_\infty \le \eta}
\mathbb{E}_{\tau\sim\mathcal{T}}
\left[
\mathcal{L}_{\mathrm{sgd}}(x+\delta,\tau;\epsilon_\theta)
\right].
\end{equation}
Our method directly intervenes in the attention-mediated source-to-video grounding pathway of the I2V denoiser at timestep $\tau$. 
The key idea is to suppress source-conditioned retrieval from video queries to source-conditioning keys, while reinforcing video-side interactions among video tokens. 
% Given a source image $x$, I2V-DETACH learns an imperceptible perturbation $\delta$ under an $\ell_\infty$ budget $\epsilon$, producing an immunized image $x_{\mathrm{adv}} = x + \delta^*$ via a source grounding detachment loss $\mathcal{L}_{sgd}$:
% \begin{equation}
% \delta^*=\arg\min_{||\delta||_\infty  \le \epsilon} \mathbb{E}_{\tau\sim\mathcal{T}} [ \mathcal {L}_{sgd} (x+\delta, \tau)],
% \end{equation}
% Unlike target-video-based defenses, our method does not require a pre-generated target video. 
% Our method directly intervenes in the attention-mediated source-to-video grounding pathway of the denoising process.
% The key idea is to weaken the interaction from video queries to source-conditioning keys, while reinforcing video-side interactions among video tokens. 

% At each optimization step, we encode $x_{\mathrm{adv}}$ as the source conditioning latent and pass it through the I2V denoiser together with auxiliary video latents. 
% These auxiliary latents are constructed to remain source-related while avoiding simple source repetition, providing an informative state for measuring source-video interactions.

Our optimization has two complementary components. 
The \emph{source-video decoupling} objective suppresses attention logits from video queries to condition-image keys, reducing direct retrieval of protected source cues. 
To obtain informative optimization signals, we further construct auxiliary source-divergent video latents that originate from the source image but progressively deviate from simple source repetition.
The \emph{intra-video coupling} objective reinforces attention logits among video tokens, encouraging denoising to proceed through video-side interactions rather than repeatedly returning to the source image as an anchor. 
% During the update of $\delta$, the auxiliary video latents are detached, so only the source-image perturbation is optimized. 
The overall framework is shown in Figure~\ref{fig/framework}.
The following sections define these components and present the final optimization objective.

\begin{figure*}[t]
\centering
    \includegraphics[width=\linewidth]{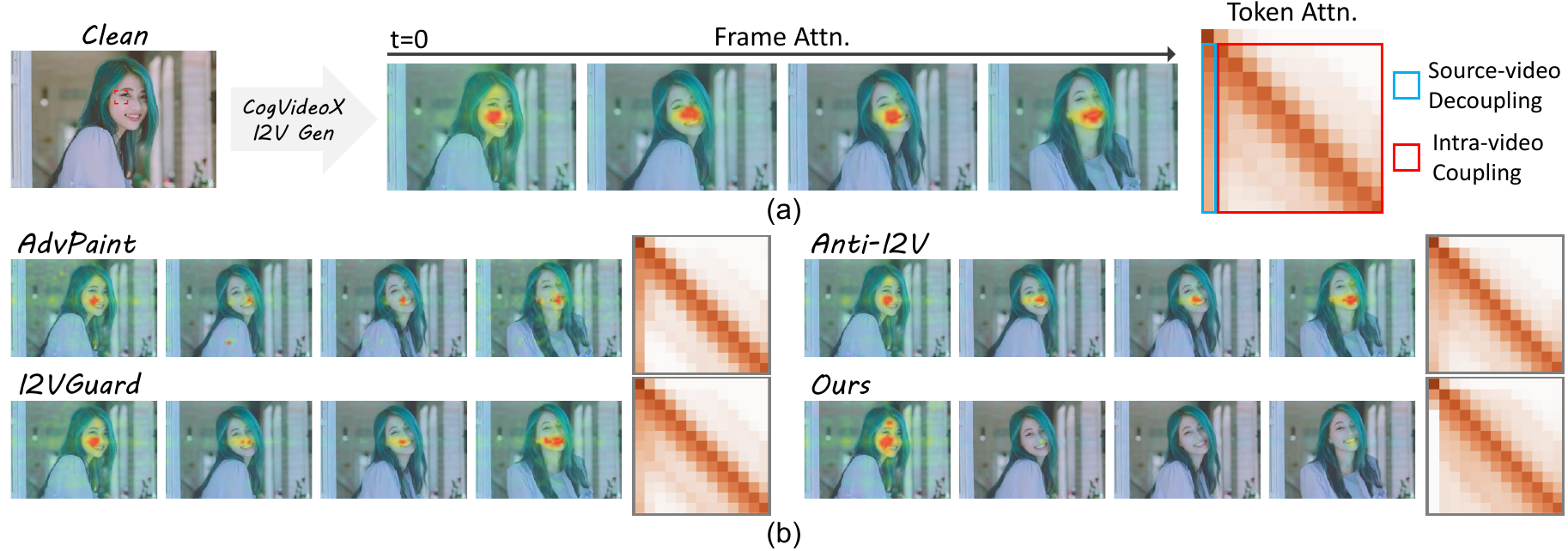}
    \vspace{-6mm}
    \caption{Frame- and token-level attention visualizations from a single denoising pass using video latents generated from the clean source image.
    In the frame-level attention maps, we highlight the most attended video tokens, with their corresponding spatial regions marked on the clean frame.
    (a) shows the result for the clean source image, while (b) shows the results for immunized images obtained by different methods.}
    \label{fig/attention}
    \vspace{-3mm}
\end{figure*}

\subsection{Source Grounding Detachment Optimization}
% Based on the pseudo-video constructed above, we then weaken the condition-dominant attention prior through which the source image maintains visual grounding over generated video latents. 

\paragraph{Source-divergent Video Latent Construction.}
\label{sec:sdv}
% \chan{
% For source-grounding suppression to be meaningful, the video-side latents should be source-related rather than arbitrary.
% Simple source repetition provides source-responsive video queries, but lacks video-side diversity and may induce shortcut attention over redundant copies.
% Therefore, we construct source-divergent video that originate from the source image while gradually diversifying during perturbation optimization.
% }
% While repeating the source image provides source-responsive video queries for {\svloss} loss, it lacks video-side diversity and may induce shortcut attention over redundant tokens.
% To prevent this, we construct source-divergent videos that stay source-oriented while gradually diversifying during perturbation optimization.
Effective source-video decoupling requires video latents that provide meaningful source-conditioned signals. 
Random or source-unrelated latents are weakly coupled with the I2V denoising process in the early timestep, whereas simple source repetition can reinforce shortcut anchoring to the input image. 
We therefore construct source-divergent video latents that start from the source image but gradually move away from the clean source latent.

% For each video frame $j$, we update an image-space perturbation $\Delta^n_j$ at $n$-th iteration by a small PGD step:
% For efficiency, we maintain a single image-space auxiliary perturbation $\Delta^n$. 
At each outer perturbation iteration $n$, we update $\Delta^{n-1}$ once using a small PGD-ascent step with step size $\beta$.
We then encode the source-divergent frame $x+\Delta^n$ into a single latent and replicate it across the video-side positions to obtain source-divergent latents $\bar{Z}^n$:
\begin{equation}
\bar{Z}^n = \!\mathcal{R}_{F-1}(\textrm{sg}[\bar{z}^n]), \ \
\bar{z}^n = \mathcal{E}(x+\Delta^n), \ \
\Delta^{n} =
\Delta^{n-1} + \beta \cdot
\mathrm{sign}
\left(
\nabla_{\Delta^{n-1}}
\hspace{-0.1cm}\left\|
\mathcal E(x+\Delta^{n-1})-\mathcal E(x)
\right\|_2^2
\right),
\label{eq:tail}
\end{equation}
where $\mathcal E$ denotes the 3D-VAE encoder applied to the input after clamping it to $[0,1]$, $F$ is the total number of visual latents, $\mathcal{R}_m(z)$ denotes the operation that repeats the latent $z$ $m$ times along the temporal dimension, and $\textrm{sg}[\cdot]$ denotes the stop-gradient operation.
After sampling a diffusion timestep $\tau\sim \mathcal{T}$, we add forward-process noise to obtain the source-divergent video latents $\bar{Z}^n_\tau$.
During the update of $\delta$, these latents are detached, ensuring that the optimization modifies only the source-image perturbation.
\iffalse
At each outer perturbation iteration $n$, we update $\Delta^n$ once using a small PGD-ascent step with a step size $\beta$.
We then construct the source-divergent frame $x+\Delta^n$, replicate it across the video-side positions, and encode them into source-divergent latents $\bar{Z}^n$:
\begin{equation}
\bar{Z}^n =\textrm{sg}[ \mathcal{E}(\mathrm{Rep}_{F-1}(x+\Delta^n))], \quad \Delta^{n+1} = 
% \Pi_{\|\Delta\|_\infty \leq \epsilon_\Delta}
% \left(
\Delta^n + \beta \cdot
\mathrm{sign}
\left(
\nabla_{\Delta^n}
\left\|
\mathcal E(x+\Delta^n)-\mathcal E(x)
\right\|_2^2
% \right)
\right),
\label{eq:tail}
\end{equation}
where $\mathcal E$ denotes 3D-VAE, $F$ is the number of video frames, $\mathrm{Rep}_m(x)$ denotes the operation that repeats the image $x$ $m$ times along the temporal dimension, and $\textrm{sg}(\cdot)$ denotes stop-gradient.
After sampling a diffusion timestep $\tau\sim \mathcal{T}$, we add forward-process noise to obtain the source-divergent video latents $\bar{Z}^n_\tau$.
% The resulting frames are encoded by the 3D-VAE and used as auxiliary video latents. 
During the update of $\delta$, these latents are detached, ensuring that the optimization modifies only the source-image perturbation. 
\fi

The step-wise update of $\Delta$ prevents the auxiliary video latents from 
becoming either source-unrelated or trivially source-repeated. 
By keeping the latents source-conditioned while gradually deviating from exact source copies, it provides informative grounding signals for $\mathcal{L}_{\mathrm{svd}}$ across a broader range of source-related states.
Given the auxiliary video latents $\bar{Z}_\tau^n$, we run the frozen denoiser and compute two complementary losses: source-video decoupling and intra-video coupling.
\vspace{-0.2cm}
% In practice, we optimize a single source-divergent frame and replicate it across the remaining video positions for efficiency; Appendix~\ref{app:pseudo} shows that independently optimizing all auxiliary frames gives only marginal additional gains.

\paragraph{Source-Video Decoupling Loss.}
To disrupt source-guided attention in I2V models, we directly suppress the pre-softmax attention logits from video queries to condition-image keys across selected layers ${L}$.
Operating at the logit level allows us to isolate this query-key alignment before attention aggregation mixes in responses from other frame keys, as follows:
\begin{equation}
\mathcal{L}_{\mathrm{svd}}
=
\frac{1}{|{L}|}
\sum_{\ell \in {L}}
\frac{1}{|{\mathcal{V}}|\,|{\mathcal{C}}|}
\sum_{i\in{\mathcal{V}}}
\sum_{c\in{\mathcal{C}}}
\left[
S\!\left(
% Q_\ell(z_{\mathrm{adv}}\cup Z_{\tau}),
% Q^\ell(z_{\mathrm{adv}}, \bar{Z}_{\tau}),
Q^\ell,
K^\ell
\right)
\right]_{ic},
% \bigl[S(Q^{\ell},K^{\ell})\bigr]_{ij},
\label{eq:loss_svd}
\end{equation}
%  $Q^\ell([z_{\mathrm{adv}}; \bar{Z}_{\tau}])$
% \begin{equation}
% \mathcal{L}_{\mathrm{svd}}
% =
% \frac{1}{|{L}|}
% \sum_{\ell \in {L}}
% \frac{1}{|{V}|\,|{C}|}
% \sum_{i\in{V}}
% \sum_{c\in{C}}
% \left[
% S\!\left(
% % Q_\ell(z_{\mathrm{adv}}\cup Z_{\tau}),
% Q^\ell_V([z_{\mathrm{adv}}; \bar{Z}_{\tau}]),
% K^\ell_C(z_{\mathrm{adv}})
% \right)
% \right]_{ic},
% \label{eq:loss_svd}
% \end{equation}
%
% where $S(\cdot,\cdot)$ denotes the pre-softmax attention logit in Eq.~\eqref{eq:logit}, and $Q$ and $K$ are query and key embeddings for immunized source latent $z_{\mathrm{adv}}$ and source-divergent video latents $\bar{Z}_\tau$ at timestep $\tau$, respectively.
where $S(\cdot,\cdot)$ denotes the pre-softmax attention logit function defined in Eq.~\eqref{eq:logit}. 
Both $Q^\ell$ and $K^\ell$ are obtained from the joint latent sequence $[z_{\mathrm{adv}};\bar{Z}_{\tau}^n]$ and at layer $\ell$ and timestep $\tau$ and iteration $n$. 
The subscripts $\mathcal{C}$ and $\mathcal{V}$ indicate indexing over the source-token and video-token subsets, respectively.
For architectures without a unified full-attention sequence, $Q^\ell$ and $K^\ell$ are taken from the corresponding source-to-video or video-side attention module.
%denote the immunized source latent and source-divergent video latents at timestep $\tau$, respectively.
% The regions ${V}$ and ${C}$ correspond to video queries and condition-image keys in the projected attention tensors.
By suppressing these responses, we prevent video queries from retrieving source-image information from condition-image keys, weakening source-guided visual grounding during denoising.
\vspace{-0.2cm}
\paragraph{Intra-video Coupling Loss.}
We further introduce {\vvloss} loss, which encourages video-side self-interactions to weaken  temporal anchoring to the conditioning image.
Specifically, we maximize the attention logits among the video-side latents:
\begin{equation}
\mathcal{L}_{\mathrm{ivc}}
=-
\frac{1}{|{L}|}
\sum_{\ell \in {L}}
\frac{1}{|{\mathcal{V}}|^2}
\sum_{i\in{\mathcal{V}}}
\sum_{j\in{\mathcal{V}}}
\left[
S\!\left(
% Q^\ell_V\!\left(\bar{Z}_\tau\right),
Q^\ell,
K^\ell
\right)
\right]_{ij}.
\label{eq:loss_vse}
\end{equation}
% \begin{equation}
% \mathcal{L}_{\mathrm{ivc}}
% =-
% \frac{1}{|{L}|}
% \sum_{\ell \in {L}}
% \frac{1}{|{V}|^2}
% \sum_{i\in{V}}
% \sum_{j\in{V}}
% \left[
% S\!\left(
% % Q^\ell_V\!\left(\bar{Z}_\tau\right),
% \big(Q^\ell_c ;Q^\ell_V\!(\bar{Z}_\tau) \big),
% K^\ell_V\!\left(\bar{Z}_\tau\right)
% \right)
% \right]_{ij}.
% \label{eq:loss_vse}
% \end{equation}
%
% where $c$ is condition image. 
This reinforces query--key matching among video tokens, increasing the response of video queries to video keys and countering the condition-dominant attention pattern.
Denoising is thus steered toward video-side interactions, hindering a return to the source-guided trajectory.
\vspace{-0.2cm}

% \input{sec/4.2_joint}

% \subsection{Overall Objective}
\paragraph{Overall Objective.}
We optimize the adversarial perturbation $\delta$ by jointly minimizing the source-video decoupling loss and the video-side interaction loss:
\begin{align}
    \mathcal{L}_{\mathrm{sgd}}=\mathcal{L}_{\mathrm{svd}}+ \lambda \mathcal{L}_{\mathrm{ivc}},
    \label{eq:objective}
\end{align}
where $\lambda$ is a hyperparameter set to $1$ in all experiments.
\iffalse
\begin{align}
   \delta^*=\arg\min_{\delta}\left[\mathcal{L}_{\mathrm{svd}}(z_{\mathrm{adv}},Z_\tau)+ \lambda \mathcal{L}_{\mathrm{ivc}}(z_{\mathrm{adv}},Z_\tau)\right]
    \quad
    \text{s.t.} \quad \| \delta \|_{\infty} \leq \epsilon,
    \label{eq:objective}
\end{align}
where $\eta$ denotes the perturbation budget and $\lambda$ is a hyperparameter set to $1$ in all experiments.
\fi
%
The effect of the proposed loss is illustrated in Figure~\ref{fig/attention}(b).
Compared with prior methods~\cite{editshield, advpaint, i2vguard, antii2v}, \method substantially weakens salient attention between the protected source image and clean-source generated video latents, indicating effective detachment from the source-video anchor.
The token-level attention maps also show strengthened video-side interactions, consistent with our intra-video coupling objective.
This trend is further quantified by the attention mass in Table~\ref{tab:att-mass} of Appendix~\ref{app/att-mass}, where \method markedly reduces source-video attention mass while increasing video-video attention mass.
Together, these results show that denoising is shifted away from source-dominated retrieval and toward source-suppressed video-side interactions, leading to visually source-detached videos.

\begin{wrapfigure}{r}{0.55\columnwidth}
\vspace{-1.7cm}
\begin{minipage}{0.54\columnwidth}
\begin{algorithm}[H]
\caption{\method}
\label{alg:sigguard}
\footnotesize
\textbf{Input:} Source image $x$, I2V model $\epsilon_\theta$, perturbation budget $\eta$, step size $\alpha$, iterations $N$, diffusion timestep set $\mathcal{T}$, visual latent count $F$\\
\textbf{Output:} Immunized image $x_{\mathrm{adv}}$. \\% and perturbation $\delta$.\\
\vspace{-0.4cm}
\begin{algorithmic}[1]
    \STATE Initialize $\delta \leftarrow 0$, \quad $\Delta^0 \sim \mathcal{N}(0,0.01^2)$
    \FOR{$n = 1, \dots, N$}
        % \STATE // Encode immunized image \\
        % \STATE $x_{\mathrm{adv}} \leftarrow \mathrm{Clip}(x+\delta, 0, 1)$
        % \STATE // encode immunized image \\
        \STATE  $z_{adv} \leftarrow \mathcal{E}(\mathrm{Clip}(x+\delta, 0, 1))$ // Encode immunized image

         \STATE Update $\Delta^n$ \& source-divergent video latent by Eq.~(\ref{eq:tail}) \\
         % $\Delta^n \leftarrow \Delta^{n-1} + \beta$\\
         % $\bar{Z}^n \leftarrow \textrm{sg}[\mathcal E (\text{Clip}(\textrm{Rep}_{F-1}(x+\Delta^{n}), 0, 1))]$\\
         $\bar{Z}^n \leftarrow \mathcal{R}_{F-1}(\mathrm{sg}[\mathcal{E}(\mathrm{Clip}(x+\Delta^n,0,1))])$\\
        % \STATE Construct source-divergent video frames by Eq.~(\ref{eq:tail})\\% and encode to latent\\
        % $Z^\mathrm{sdv} = \mathcal{E}(V^\mathrm{sdv}), \quad V^\mathrm{sdv}=\{v_1,\dots,v_{F-1}\}$ 
        \STATE Sample diffusion timestep $\tau \sim \mathcal{T}$
        % \STATE Construct source-deviated tail latent group\\
        % $Z^{\mathrm{tail}}=\{z_{k+1}^{\mathrm{tail}},\dots,z_F^{\mathrm{tail}}\}$ by Eq.~(\ref{eq:tail})
        
        \STATE Add the forward-process noise to video latents\\
        ${\bar Z_\tau^\mathrm{n}} \leftarrow \mathrm{AddNoise}(\bar Z^\mathrm{n}, \tau; \epsilon_\theta)$
        
        \STATE %// Compute loss using Eq.~\eqref{eq:loss_svd} and~\eqref{eq:loss_vse}
        $L_\mathrm{sgd} \leftarrow L_{\mathrm{svd}} + \lambda L_{\mathrm{ivc}}$ // Compute loss using Eq.~\eqref{eq:loss_svd} and~\eqref{eq:loss_vse}
        
        \STATE %// Update perturbation
        $\delta \leftarrow \delta - \alpha \cdot \mathrm{sign}(\nabla_\delta L_\mathrm{sgd})$ // Update perturbation
        
        \STATE
        $\delta \leftarrow \mathrm{Clip}(\delta, -\eta, \eta)$  // Project $\delta$ into $\ell_\infty$ ball
    \ENDFOR
    \STATE \textbf{return} $x_{\mathrm{adv}}=\mathrm{Clip}(x+\delta,0,1)$
\end{algorithmic}
\end{algorithm}
\vspace{-1.5cm}
\end{minipage}
\end{wrapfigure}
\vspace{-1mm}
\subsection{Overall Algorithm}

The overall \method framework and optimization procedure are shown in Figure~\ref{fig/framework} and Algorithm~\ref{alg:sigguard}, respectively.
We first perform a one-step update of the source-divergent video latents using Eq.~\eqref{eq:tail} with a small step size.
We then sample a diffusion timestep and add the corresponding forward-process noise to the video latents.
The perturbation $\delta$ is then updated using our objective in Eq.~\eqref{eq:objective}, which disrupts source grounding and prevents the generated video from faithfully preserving the visual content of the source image.
% \begin{figure*}[t]
% \centering
%     \includegraphics[width=\linewidth]{fig/qual_3.pdf}
%     \vspace{-0.6cm}
%     \caption{Qualitative results for our method across diverse diffusion-based I2V architectures, including CogVideoX~\cite{cogvideox}, LTX-Video~\cite{ltx}, and OpenSora1.2~\cite{opensora}. Regardless of the underlying model architecture, our method consistently produces outputs that are no longer reliably grounded in the source image.}
%     \label{fig/qual_2}
%     \vspace{-0.4cm}
% \end{figure*}

\vspace{-2mm}
\section{Experiments}
\vspace{-2mm}
\subsection{Experimental Setup}
\vspace{-2mm}
% \vspace{-1mm}
\paragraph{Implementation details.}
% We select three representative diffusion-based I2V models that can be realistically used for video generation and perturbation optimization within an 80GB memory budget, covering different attention architectures: CogVideoX-5B~\cite{cogvideox} with MMDiT full-attention, LTX-Video-2B~\cite{ltx} with cross-attention, and DiT-based OpenSora 1.2~\cite{opensora}.
To examine the generality of our method across diverse DiT-based I2V architectures, we evaluate three representative models feasible for video generation and perturbation optimization within an 80GB memory budget: CogVideoX-5B~\cite{cogvideox} with MMDiT full-attention, LTX-Video-2B~\cite{ltx} with cross-attention, and DiT-based OpenSora 1.2~\cite{opensora}.
For our method, we sample diffusion timestep from the first 10\% of the denoising step, set the step sizes $\alpha=1/255$ and $\beta=2/255$, and use the 25--75\% layer range for all models.
% Detailed EOT and model-specific I2V settings are provided in the Appendix.
For a fair comparison, we set perturbation budget $\eta=8/255$ following I2VGuard~\cite{i2vguard} and details for baselines are provided in Appendix~\ref{sec:imple}.
% , step size = $1/255$, and iteration = 50 for EditShield~\cite{editshield}, AdvPaint~\cite{advpaint}, and I2VGuard~\cite{i2vguard}, and .

% Since official code for the I2V-specific baselines, I2VGuard~\cite{i2vguard} and Anti-I2V~\cite{antii2v}, is unavailable, we implement them as faithfully as possible based on the original papers and additional clarifications obtained through personal communication with the authors.
\vspace{-2mm}
\paragraph{Baselines.}
We first compare our method with state-of-the-art I2V immunization methods, including I2VGuard~\cite{i2vguard} and Anti-I2V~\cite{antii2v}, which are specifically designed to defend against malicious image-to-video generation.
We further include EditShield~\cite{editshield}, which operates on latent-level representations, and AdvPaint~\cite{advpaint}, which perturbs attention embedding representations.
Although originally developed for I2I settings, these methods are relevant to our study because they intervene on representation spaces closely associated with source-content preservation.
This selection allows us to evaluate our method against both task-specific I2V baselines and objective-relevant I2I baselines.

\vspace{-1mm}
\paragraph{Evaluation dataset and metrics.}
Since there is no benchmark designed for I2V immunization, we sample 100 image-prompt pairs from VBench~\cite{vbench}, widely used for I2V generation, including 80 human-related images and 20 non-human images.
For evaluation, we use \textit{I2V Subject} (IS) and \textit{Subject Consistency} (SC) from VBench~\cite{vbench, vbench++}, along with \textit{DINO Similarity} (DS)~\cite{oquab2024dinov} between the source image and generated frames.
These metrics assess how strongly the generated video preserves the source subject and visual content, while lower values indicate stronger immunization.
We also include quality metrics to monitor the perceptual quality of generated videos, \textit{aesthetic quality} (AQ) and \textit{imaging quality} (IQ) from VBench.
\iffalse
Since there is no publicly available benchmark designed for I2V immunization, we sample 100 images from VBench, one of the most widely used for I2V generation, including 80 human-related images and 20 non-human images.
For evaluation, we use \textit{I2V Subject} (IS) and \textit{Subject Consistency} (SC) from VBench~\cite{vbench, vbench++}, along with \textit{DINO Similarity} (DS)~\cite{oquab2024dinov} between the source image and generated frames.
These metrics assess how strongly the generated video preserves the source subject and visual content.
Lower values indicate stronger immunization.
% and therefore stronger immunization.
We also include quality metrics to show the perceptual quality of generated videos, \textit{aesthetic quality} (AQ) and \textit{imaging quality} (IQ) from VBench.
\fi
% Lower IS, SC, and DS indicate stronger immunization, while AQ and IQ are reported to show the perceptual quality of generated videos.
% Detailed descriptions are provided in Appendix~\ref{app:data}

% \begin{figure*}[t]
% \centering
%     \includegraphics[width=\linewidth]{fig/qual_1.pdf}
%     \vspace{-0.6cm}
%     \caption{Qualitative comparison with baselines on CogVideoX~\cite{cogvideox}. Our method leads to videos that fail to faithfully reflect the source visual content, rather than merely exhibiting temporal degradation.}

%     \label{fig/qual_1}
% \end{figure*}

\begin{figure*}[t]
\centering
    \includegraphics[width=0.93\linewidth]{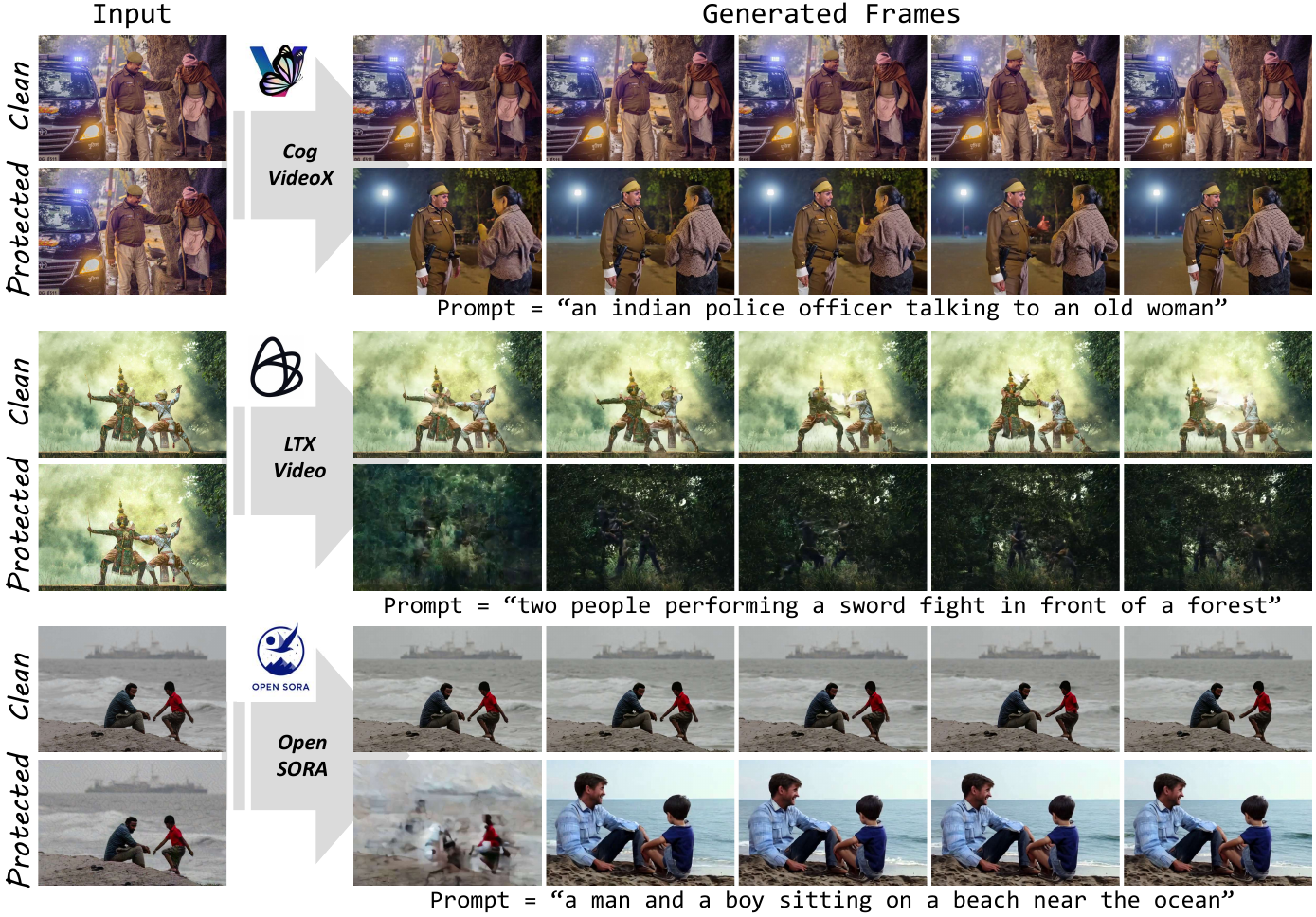}
    \vspace{-0.2cm}
    \caption{Qualitative results for our method across diverse diffusion-based I2V architectures~\cite{cogvideox, ltx, opensora}. %including CogVideoX~\cite{cogvideox}, LTX-Video~\cite{ltx}, and OpenSora1.2~\cite{opensora}.
    Regardless of the underlying model, our method consistently produces visually detached videos.}
    \label{fig/qual_2}
    \vspace{-0.3cm}
\end{figure*}
\vspace{-2mm}

\subsection{Main Results}
\vspace{-2mm}
\begin{table}[t]
\footnotesize
\centering
\setlength{\tabcolsep}{1.5pt}
\setlength{\dashlinedash}{2pt}
\setlength{\dashlinegap}{1pt}

\caption{White-box performance on CogVideoX~\cite{cogvideox}, LTX-Video~\cite{ltx}, and OpenSora 1.2~\cite{opensora}. \emph{Space} denotes the perturbation domain used for optimization. \emph{V/P} denotes that methods require a \emph{pre-generated video} and \emph{prompt} during perturbation optimization, respectively. * refers to the variant that incorporates purification-aware optimization. The best score is shown in \textbf{bold} and the second-best is \underline{underlined}.}
% \vspace{-2mm}
\resizebox{\linewidth}{!}{
\begin{tabular}{lcc|ccc;{0.2pt/1pt}cc|ccc;{0.2pt/1pt}cc|ccc;{0.2pt/1pt}cc}
\Xhline{0.35mm}

\multicolumn{3}{c|}{Model}
& \multicolumn{5}{c|}{CogVideoX} 
& \multicolumn{5}{c|}{LTX-Video}
& \multicolumn{5}{c}{OpenSora 1.2} \\

\hline
Method & Space & V/P
& IS $\downarrow$ & SC $\downarrow$ & DS $\downarrow$ & AQ  & IQ 
& IS $\downarrow$ & SC $\downarrow$ & DS $\downarrow$ & AQ  & IQ
& IS $\downarrow$ & SC $\downarrow$ & DS $\downarrow$ & AQ  & IQ \\

\hline
\rowcolor{gray!15}
Clean
& -
& -
& 97.28 & 94.83 & 89.17 & 60.46 & 70.03
& 97.12 & 94.22 & 81.04 & 57.85 & 67.60
& 95.73 & 95.17 & 87.30 & 56.95 & 66.10 \\

EditShield~\cite{editshield}
& RGB
& \ding{55}/ \ding{55}
& 95.97 & 93.77 & 86.09 & 60.72 & 65.41
& {92.63} & {86.85} & {69.62} & {51.04} & {57.04}
& 86.39 & 88.98 & 70.08 & 53.52 & 62.64 \\

AdvPaint~\cite{advpaint}
& RGB
& \ding{55}/ \ding{51}
& 93.27 & {89.90} & {78.42} & {57.11} & {63.64}
& 93.87 & 90.33 & 73.24 & 54.43 & 59.18
& {84.47} & {83.39} & {58.80} & {48.92} & {56.10} \\

I2VGuard~\cite{i2vguard}
& RGB
& \ding{51}/ \ding{51}
& 94.39 & 92.52 & 82.99 & 58.65 & {63.39}
& 94.10 & 89.67 & 73.37 & 54.47 & 61.08
& 92.20 & 92.68 & 77.68 & 56.10 & 62.19 \\

% \cdashline{1-18}
\rowcolor{green!15}
Ours
& RGB
& \ding{55}/ \ding{55}
& \underline{86.39} & \underline{80.63} & \underline{65.54} & {57.87} & 66.76
& \underline{89.85} & \underline{81.98} & \underline{60.88} & {50.35} & {58.51}
& \underline{79.62} & \underline{73.92} & \underline{52.10} & {48.95} & {52.38} \\

\cdashline{1-18}

Anti-I2V*~\cite{antii2v}
& Lab+DCT
& \ding{51}/ \ding{51}
& {93.13} & 92.42 & 82.34 & 60.97 & 65.38
& 93.80 & 89.50 & 75.12 & 53.40 & 63.77
& 91.64 & 91.97 & 77.33 & 55.30 & 62.03 \\

\rowcolor{green!15}
Ours*
& RGB+DCT
& \ding{55}/ \ding{55}
& \textbf{83.52} & \textbf{78.30} & \textbf{63.33} & 59.52 & 67.55
& \textbf{88.70} & \textbf{81.08} & \textbf{58.94} & {49.73} & 58.52
& \textbf{78.95} & \textbf{73.70} & \textbf{51.40} & 48.98 & {52.55} \\

\Xhline{0.35mm}
\end{tabular}
}
\vspace{-4mm}
\label{tab:white}
\end{table}
\paragraph{Qualitative comparison.}
We first qualitatively compare our method against the baselines on CogVideoX~\cite{cogvideox}.
As shown in Figure~\ref{fig/teaser}, \method produces visually source-detached videos, whereas the baselines mainly induce perceptual degradation while still retaining source-related content.
% As shown in Figure~\ref{fig/teaser}, the video frames generated from our method clearly fail to preserve the visual content of the input image, whereas the baselines mainly induce perceptual degradation while still retaining source-related content.
We also present qualitative results on diverse diffusion-based I2V models in Figure~\ref{fig/qual_2}.
While the resulting generation patterns differ across architectures, a common trend is clearly observed: all three models fail to faithfully preserve the visual content of the source image.
This indicates that the proposed perturbation consistently breaks source grounding, rather than merely inducing temporal degradation.
More qualitative comparisons and discussion are provided in Appendix~\ref{sec:qual_comparison_wb}.
\vspace{-0.2cm}

\paragraph{Quantitative comparison.}
\begin{table*}[t]
\footnotesize
\centering
\setlength{\tabcolsep}{5pt}
\caption{Human evaluation results. We evaluate each method in terms of \textit{Source Reflection} (Src. Ref.), \textit{Protection Assessment} (Pro. Ass.), and \textit{Overall Effectiveness} (Ovr. Eff.). ES, AP, IG, and AI denote EditShield~\cite{editshield}, AdvPaint~\cite{advpaint}, I2VGuard~\cite{i2vguard}, and Anti-I2V~\cite{antii2v}, respectively.}
\vspace{-2mm}
\resizebox{\linewidth}{!}{
\begin{tabular}{l|ccccc|ccccc|ccccc}
\Xhline{0.35mm}

 Model
& \multicolumn{5}{c|}{CogVideoX}
& \multicolumn{5}{c|}{LTX-Video}
& \multicolumn{5}{c}{OpenSora 1.2} \\

\cline{1-16}
  Method
& ES & AP & IG & AI & Ours
& ES & AP & IG & AI & Ours
& ES & AP & IG & AI & Ours \\

\Xhline{0.35mm}

 Src. Ref. $\downarrow$
& 4.34 & \underline{3.33} & 4.13 & 4.33 &  \cellcolor{green!15}{\textbf{1.46}}
& \underline{2.46} & 2.99 & 4.09 & 3.06 & \cellcolor{green!15}{\textbf{1.49}}
& 3.76 & \underline{2.94} & 4.04 & 3.86 & \cellcolor{green!15}{\textbf{1.64}}\\

 Pro. Ass. $\uparrow$
& 0.20 & 0.20 & 0.20 & \underline{0.27} & \cellcolor{green!15}{\textbf{4.12}}
& \underline{0.85} & 0.20 & \underline{0.85} & 0.45 & \cellcolor{green!15}{\textbf{2.65}}
& 0.32 & \underline{0.90} & 0.22 & 0.27 & \cellcolor{green!15}{\textbf{3.27}}\\

 Ovr. Eff. $\downarrow$
& 4.10 & \underline{2.50} & 3.05 & 4.00 & \cellcolor{green!15}{\textbf{1.35}}
& \underline{2.25} & 3.30 & 4.45 & 3.60 & \cellcolor{green!15}{\textbf{1.40}}
& 4.00 & \underline{2.10} & 3.95 & 3.55 & \cellcolor{green!15}{\textbf{1.40}}\\

% \textbf{}
% \ underline{}

\Xhline{0.35mm}
\end{tabular}
}
\vspace{-3mm}
\label{tab:human_llm_eval}
\end{table*}
We evaluate immunization performance across diverse diffusion-based I2V models.
As shown in Table~\ref{tab:white}, our method consistently outperforms the baselines by a substantial margin on consistency metrics (\ie, IS, SC, and DS), while requiring neither additional reference videos nor prompts, in contrast to the baselines~\cite{advpaint, i2vguard, antii2v}.
In addition, \method demonstrates strong generalizability across diverse attention architectures~\cite{cogvideox, ltx, opensora}.
We further evaluate an RGB+DCT variant, denoted as \emph{Ours*}, to validate  compatibility with robustness-oriented perturbation designs~\cite{dct, antii2v}.
Its improved performance suggests that our objective can be effectively extended beyond RGB-space perturbations to the DCT domain.
AQ/IQ trends differ by model, likely because each I2V model fails under immune perturbation in different ways.
CogVideoX~\cite{cogvideox} tends to preserve clean alternative videos, whereas LTX-Video~\cite{ltx} severely collapses and OpenSora 1.2~\cite{opensora} partially collapses while still producing source-deviated videos, but the key point is the broken source grounding rather than the quality score.
\vspace{-0.2cm}
% These results indicate that \method effectively disrupts source grounding across diverse model architectures, thereby preventing the generated videos from reliably preserving the visual cues of the source image.
% The AQ/IQ trends reflect model-specific modes
%Appendix~\ref{sec:qual_comparison_wb}.

\begin{table}[t]
\footnotesize
\centering
\setlength{\tabcolsep}{5pt}
\setlength{\dashlinedash}{2pt}
\setlength{\dashlinegap}{1pt}
\newcommand{\badchg}[1]{{\scriptsize\textcolor{red}{(#1)}}}
\newcommand{\goodchg}[1]{{\scriptsize\textcolor{blue}{(#1)}}}

\caption{Robustness evaluation against transformation and diffusion-based purification techniques. *~refers to the variant that incorporates purification-aware optimization. The best score is shown in \textbf{bold} and the second-best is \underline{underlined}.}
% \vspace{-2mm}
\resizebox{\linewidth}{!}{
\begin{tabular}{l|ccc|ccc|ccc|ccc}
% \begin{tabular}{c|ccc:cc|ccc:cc|ccc:cc}
\Xhline{0.35mm}

{Target Model}
& \multicolumn{3}{c|}{JPEG~\cite{jpeg}}
& \multicolumn{3}{c|}{Gaussian noise}
% & \multicolumn{3}{c|}{Crop \& Resize}
& \multicolumn{3}{c|}{Adverse Clean~\cite{advclean}}
% & \multicolumn{3}{c|}{Cond. DiffPure~\cite{condiffpure}}
& \multicolumn{3}{c}{GridPure~\cite{gridpure}}\\

\hline
Method
& IS $\downarrow$ & SC $\downarrow$ & DS $\downarrow$ 
& IS $\downarrow$ & SC $\downarrow$ & DS $\downarrow$ 
& IS $\downarrow$ & SC $\downarrow$ & DS $\downarrow$
& IS $\downarrow$ & SC $\downarrow$ & DS $\downarrow$  \\

\hline
% Clean
% & - & - & - & - & -
% & - & - & - & - & -
% & - & - & - & - & - \\
EditShield~\cite{editshield}
& 94.95 & 92.74 & 84.37 
& 96.03 & 94.19 & 86.55 
% & 94.77 & 93.77 & 84.98 crop
& 95.30 & 93.81 & 84.84 
& 96.30 & 94.36 & 87.00  \\

AdvPaint~\cite{advpaint}
& 94.42 & 92.30 & 83.17
& 94.14 & 96.03 & 86.55 
% & 94.58 & 93.45 & 84.29 crop
& 95.49 & 94.28 & 85.66 
& 95.53 & 93.50 & 85.48  \\

I2VGuard~\cite{i2vguard}
& 93.69 & 92.29 & 82.38 
& 94.19 & 92.72 & 82.85 
% & 93.71 & 93.14 & 83.09 crop
& 94.66 & 93.10 & 83.26 
& 94.55 & 93.12 & 83.27  \\

\rowcolor{green!15}
Ours
& \underline{90.21} & \underline{86.11} & \underline{74.26}
& \underline{87.57} & \underline{81.85} & \underline{67.54} 
% & \underline{92.43} & \underline{89.85} & \underline{79.91} crop
& \underline{90.31} & \underline{86.09} & \underline{73.62} 
& \underline{91.18} & \underline{86.12} & \underline{74.81} \\

\cdashline{1-13}

Anti-I2V*~\cite{antii2v}
& 93.33 & 92.32 & 83.08 
& 92.92 & 92.56 & 82.22 
% & 92.94 & 92.66 & 82.62 crop
& 94.23 & 92.88 & 84.49 
& 93.39 & 92.47 & 83.25  \\

\rowcolor{green!15}
Ours*
& \textbf{84.98} & \textbf{80.14} & \textbf{66.57}
& \textbf{83.80} & \textbf{78.89} & \textbf{64.32} 
% & \textbf{92.43} & \textbf{89.85} & \textbf{79.91} crop
& \textbf{88.10} & \textbf{83.92} & \textbf{70.51} 
& \textbf{85.06} & \textbf{79.60} & \textbf{64.83} \\

\Xhline{0.35mm}
\end{tabular}
}
\vspace{-3mm}
\label{tab:puri}
\end{table}

% EditShield~\cite{editshield}
% & 94.95 & 92.74 & 84.37 
% & 96.03 & 94.19 & 86.55 
% % & 94.77 & 93.77 & 84.98 crop
% & 95.30 & 93.81 & 84.84 
% & - & - & - 
% & 96.30 & 94.36 & 87.00  \\

% AdvPaint~\cite{advpaint}
% & 94.42 & 92.30 & 83.17
% & 94.14 & 96.03 & 86.55 
% % & 94.58 & 93.45 & 84.29 crop
% & 95.49 & 94.28 & 85.66 
% & - & - & - 
% & 95.53 & 93.50 & 85.48  \\

% I2VGuard~\cite{i2vguard}
% & 93.69 & \underline{92.29} & \underline{82.38} 
% & 94.19 & 92.72 & 82.85 
% % & 93.71 & 93.14 & 83.09 crop
% & 94.66 & 93.10 & \underline{83.26} 
% & - & - & - 
% & 94.55 & 93.12 & 83.27  \\

% \rowcolor{green!15}
% Ours
% & \textbf{90.21} & \textbf{86.11} & \textbf{74.26}
% & \textbf{87.57} & \textbf{81.85} & \textbf{67.54} 
% % & \textbf{92.43} & \textbf{89.85} & \textbf{79.91} crop
% & 90.31 & 86.09 & 73.62 
% & - & - & - 
% & \textbf{91.18} & \textbf{86.12} & \textbf{74.81} \\

% \cdashline{1-16}

% Anti-I2V*~\cite{antii2v}
% & \underline{93.33} & 92.32 & 83.08 
% & \underline{92.92} & \underline{92.56} & \underline{82.22} 
% % & \underline{92.94} & \underline{92.66} & \underline{82.62} crop
% & \underline{94.23} & \underline{92.88} & 84.49 
% & - & - & - 
% & \underline{93.39} & \underline{92.47} & \underline{83.25}  \\

% \rowcolor{green!15}
% Ours*
% & \textbf{84.98} & \textbf{80.14} & \textbf{66.57}
% & \textbf{83.80} & \textbf{78.89} & \textbf{64.32} 
% % & \textbf{92.43} & \textbf{89.85} & \textbf{79.91} crop
% & 88.10 & 83.92 & 70.51 
% & - & - & - 
% & \textbf{85.06} & \textbf{79.60} & \textbf{64.83} \\

% \Xhline{0.35mm}
% \end{tabular}
% }
% \vspace{0.1cm}
% \label{tab:puri}
% \end{table}
\paragraph{Human Evaluation.}
\label{sec:human}

We conduct human evaluation with 20 crowdsourced participants along three dimensions: 
(i) \textit{Source Reflection} (SR), how faithfully the generated video preserves the visual content of the source image on a 1--5 scale; 
(ii) \textit{Protection Assessment} (PA), select one method that best protects the source image from a protection perspective; and 
(iii) \textit{Overall Effectiveness} (OE), rank the methods according to their overall protection effectiveness.
For the human evaluation, we randomly sample 10 videos from each model and method.
All scores are converted to a 5-point scale.
Lower is better for SR and OE while higher scores indicate stronger protection for PA.
As shown in Table~\ref{tab:human_llm_eval}, the lowest SR indicates that human evaluators consistently find that \method least preserves the visual content of the source image across all models.
Furthermore, our method is most frequently selected in PA and receives the best ranking in OE, suggesting that source grounding detachment provides a suitable protection criterion for unauthorized I2V generation.
Details are provided in Appendix~\ref{app:human}.
\vspace{-0.3cm}

\subsection{Robustness Evaluation against Purification}
\vspace{-0.2cm}
We evaluate the robustness of \method against perturbation purification.
Specifically, we consider preprocessing-based purification, including JPEG compression~\cite{jpeg}, Gaussian noise, and Adverse Cleaner~\cite{advclean}, as well as diffusion-based GridPure~\cite{gridpure}.
% We further evaluate an RGB+DCT variant (\ie, \emph{Ours*}) to validate compatibility with robustness-oriented perturbation designs~\cite{dct, antii2v} under purification techniques.
As shown in Table~\ref{tab:puri}, our original RGB-space method already achieves the strongest protection performance under perturbation purification compared with baselines.
% This indicates that ourprovides a more suitable objective for image immunization against I2V generation than focusing primarily on temporal degradation.
Furthermore, the additional robustness of \emph{Ours*}, obtained by incorporating DCT perturbations, shows that our objective can be integrated with existing purification-aware techniques for improved robustness.
\vspace{-0.1cm}

\iffalse
We evaluate the robustness of our method against perturbation purification.
Specifically, we consider purification techniques that are effective at removing adversarial perturbations, including transformation-based methods such as JPEG compression~\cite{jpeg}, Gaussian noise, and Adverse Cleaner~\cite{advclean}, as well as GridPure~\cite{gridpure}, a diffusion-based purification method tailored to high-resolution adversarial images.
As shown in Table~\ref{tab:puri}, our method achieves the strongest immunization performance under perturbation purification.
While Anti-I2V~\cite{antii2v} is designed to consider purification robustness, our method still retains stronger protection effectiveness.
This suggests that directly disrupting visual grounding provides a more suitable objective for image immunization against I2V generation than focusing primarily on temporal degradation.
An additional integration with prior robustness-oriented methods is provided in the Appendix to further examine the extensibility of our approach.
\fi

\vspace{-1.mm}
\begin{table}[t!]
\footnotesize
\centering
\caption{Transferability evaluation on LTX-Video~\cite{ltx} and OpenSora 1.2~\cite{opensora}. CogVideoX~\cite{cogvideox} is used for victim model. The best score is shown in \textbf{bold} and the second-best is \underline{underlined}.}
% \vspace{-2mm}
\resizebox{0.9\linewidth}{!}{
\begin{tabular}{l|ccc;{0.2pt/1pt}cc|ccc;{0.2pt/1pt}cc}
% \begin{tabular}{c|ccc:cc|ccc:cc|ccc:cc}
\Xhline{0.35mm}

{Target Model}
& \multicolumn{5}{c|}{LTX-Video}
& \multicolumn{5}{c}{OpenSora 1.2} \\

\hline
Method
& IS $\downarrow$ & SC $\downarrow$ & DS $\downarrow$ & AQ & IQ 
& IS $\downarrow$ & SC $\downarrow$ & DS $\downarrow$ & AQ & IQ  \\

\hline
% Clean
% & - & - & - & - & -
% & - & - & - & - & - \\

EditShield~\cite{editshield}
& 94.42 & 91.23 & 77.72 & {53.09} & 62.52
& 94.86 & 94.51 & 84.51 & 54.85 & 63.07 \\

AdvPaint~\cite{advpaint}
& 94.67 & 91.41 & 76.02 & 53.93 & 62.30
& 94.69 & 94.28 & 83.46 & 54.93 & 62.99 \\

I2VGuard~\cite{i2vguard}
& 93.82 & 91.29 & 76.36 & 53.35 & {61.95}
& \underline{93.26} & \underline{94.08} & \underline{81.02} & {54.80} & {61.84} \\

Anti-I2V~\cite{antii2v}
& \underline{92.57} & \underline{90.21} & \underline{74.25} & 54.17 & 62.97
& 93.74 & 94.55 & 82.33 & 58.63 & 63.44 \\

\rowcolor{green!15}
Ours
& \textbf{92.06} & \textbf{88.26} & \textbf{72.30} & {51.23} & {60.28}
& \textbf{92.62} & \textbf{93.58} & \textbf{79.07} & {54.56} & {59.35} \\

\Xhline{0.35mm}
\end{tabular}
}
\vspace{-1mm}
\label{tab:black}
\end{table}
\subsection{Black-box Transferability}
\vspace{-1.mm}
We evaluate the black-box transferability between different diffusion architectures.
In particular, we use CogVideoX (full-attention)~\cite{cogvideox} as the victim model, and LTX-Video (cross-attention)~\cite{ltx} and OpenSora 1.2 (DiT-based)~\cite{opensora} as the black-box models.
As shown in Table~\ref{tab:black}, our method consistently maintains stronger image protection than competing methods, demonstrating strong black-box transferability to unseen diffusion models with different attention architectures.
Additional black-box evaluations, including experiments using LTX-Video and OpenSora 1.2 as surrogate models and transfer to large-scale and commercial models, are provided in Appendix~\ref{sec:black-box}.
\vspace{-0.1cm}

% \vspace{-1.mm}

\begin{table*}[t]
% \footnotesize
\centering
\setlength{\tabcolsep}{3.pt}
\caption{Imperceptibility comparison with baselines. The best score for each metric within each model is shown in \textbf{bold}.}
\vspace{-2mm}
\resizebox{\linewidth}{!}{
\begin{tabular}{l|cccc|cccc|cccc}
\Xhline{0.35mm}
Model
& \multicolumn{4}{c|}{CogVideoX~\cite{cogvideox}}
& \multicolumn{4}{c|}{LTX-Video~\cite{ltx}}
& \multicolumn{4}{c}{OpenSora 1.2~\cite{opensora}} \\
\hline
Method & PSNR $\uparrow$ & SSIM $\uparrow$ & LPIPS $\downarrow$ & DISTS $\downarrow$
& PSNR $\uparrow$ & SSIM $\uparrow$ & LPIPS $\downarrow$ & DISTS $\downarrow$
& PSNR $\uparrow$ & SSIM $\uparrow$ & LPIPS $\downarrow$ & DISTS $\downarrow$ \\
\hline
EditShield%~\cite{editshield}
& 31.54 & 0.779 & 0.174 & 0.210
& 31.55 & 0.786 & 0.388 & 0.181
& 31.43 & 0.778 & 0.400 & 0.232 \\

AdvPaint%~\cite{advpaint}
& 33.06 & 0.831 & 0.166 & 0.162
& 33.15 & 0.840 & 0.357 & 0.171
& 32.14 & 0.829 & 0.403 & 0.234 \\

I2VGuard%~\cite{i2vguard}
& 30.72 & 0.743 & 0.229 & 0.172
& 29.99 & 0.723 & 0.340 & 0.146
& 30.36 & 0.744 & 0.388 & 0.193 \\

Anti-I2V%~\cite{antii2v}
& 28.66 & 0.727 & 0.237 & 0.207
& 32.55 & 0.843 & 0.387 & 0.171
& 25.14 & 0.628 & \cellcolor{green!15}{\textbf{0.387}} & \cellcolor{green!15}{\textbf{0.191}} \\

Ours
& \cellcolor{green!15}{\textbf{33.49}} & \cellcolor{green!15}{\textbf{0.848}} & \cellcolor{green!15}{\textbf{0.155}} & \cellcolor{green!15}{\textbf{0.139}}
& \cellcolor{green!15}{\textbf{33.48}} & \cellcolor{green!15}{\textbf{0.853}} & \cellcolor{green!15}{\textbf{0.320}} & \cellcolor{green!15}{\textbf{0.145}}
& \cellcolor{green!15}{\textbf{32.61}} & \cellcolor{green!15}{\textbf{0.837}} & 0.397 & 0.214 \\
\Xhline{0.35mm}
\end{tabular}
}
\vspace{-4mm}
\label{tab:imper}
\end{table*}
\begin{table}[t!]
\footnotesize
\centering
\caption{Protection performance on OpenSora 1.2 under matched imperceptibility (DISTS $\approx$ 0.19). AdvPaint~\cite{advpaint} and I2V-DETACH are re-optimized using a perturbation budget of $6/255$.}
% \vspace{-2mm}
\resizebox{0.9\linewidth}{!}{
\begin{tabular}{c|l|ccccc}
\Xhline{0.35mm}

Category
& Metric
& ES ($8/255$)
& AP ($6/255$)
& IG ($8/255$)
& AI ($8/255$)
& Ours ($6/255$) \\
\hline

\multirow{2}{*}{Imperceptibility}
& LPIPS $\downarrow$
& 0.400
& 0.377
& 0.388
& 0.387
& \cellcolor{green!15}{ \textbf{0.354}} \\

& DISTS $\downarrow$
& 0.232
& 0.193
& 0.193
& 0.191
& \cellcolor{green!15}{ \textbf{0.187}} \\

\cdashline{1-7}

\multirow{3}{*}{Performance}
& IS $\downarrow$
& 86.39
& 88.08
& 92.20
& 91.64
& \cellcolor{green!15}{ \textbf{82.92}} \\

& SC $\downarrow$
& 88.98
& 86.03
& 92.68
& 91.97
& \cellcolor{green!15}{ \textbf{77.62}} \\

& DS $\downarrow$
& 70.08
& 66.81
& 77.68
& 77.33
& \cellcolor{green!15}{\textbf{56.51}} \\

\Xhline{0.35mm}
\end{tabular}
}
\vspace{-3mm}
\label{tab:opensora_matched_imperceptibility}
\end{table}
\begin{table*}[t!]
\footnotesize
\centering
\setlength{\tabcolsep}{5pt}
\caption{Protection performance of image-specific methods and their universal variants. Lower values indicate stronger protection. ES, AP, IG, and AI denote EditShield~\cite{editshield}, AdvPaint~\cite{advpaint}, I2VGuard~\cite{i2vguard}, and Anti-I2V~\cite{antii2v}, respectively.}
\vspace{-2mm}
\resizebox{\linewidth}{!}{
\begin{tabular}{c|cccccc|ccccc}
\Xhline{0.35mm}

\multirow{2}{*}{Method}
& \multicolumn{6}{c|}{\textit{Image-specific}}
& \multicolumn{5}{c}{\textit{UAP-variant}} \\
\cline{2-12}
& Clean & ES~\cite{editshield} & AP~\cite{advpaint} & IG~\cite{i2vguard} & AI~\cite{antii2v} & Ours
& ES~\cite{editshield} & AP~\cite{advpaint} & IG~\cite{i2vguard} & AI~\cite{antii2v} & Ours \\

\Xhline{0.35mm}

IS $\downarrow$
& 97.28 & 95.97 & 93.27 & 94.39 & \underline{93.13} & \cellcolor{green!15}{\textbf{86.39}}
& 96.32 & 95.58 & 95.37 & \underline{95.08} & \cellcolor{green!15}{\textbf{93.31}} \\

SC $\downarrow$
& 94.83 & 93.77 & \underline{89.90} & 92.52 & 92.42 & \cellcolor{green!15}{\textbf{80.63}}
& 94.45 & 94.01 & \underline{92.74} & 93.87 & \cellcolor{green!15}{\textbf{91.74}} \\

DS $\downarrow$
& 89.17 & 86.09 & \underline{78.42} & 82.99 & 82.34 & \cellcolor{green!15}{\textbf{65.54}}
& 87.57 & \underline{85.73} & 86.43 & 85.77 & \cellcolor{green!15}{\textbf{83.54}} \\

\Xhline{0.35mm}
\end{tabular}
}
\vspace{-3mm}
\label{tab:uap}
\end{table*}
\vspace{-1.mm}
\subsection{Imperceptibility Analysis}
\vspace{-1.mm}
We compare imperceptibility across methods using PSNR, DISTS~\cite{dists}, LPIPS~\cite{lpips}, and SSIM~\cite{ssim}, as a shared perturbation budget of $\eta=8/255$ does not guarantee comparable perceptual visibility.
As shown in Table~\ref{tab:imper}, \method achieves the best imperceptibility across all metrics on CogVideoX~\cite{cogvideox} and LTX-Video~\cite{ltx}.

On OpenSora 1.2~\cite{opensora}, however, \method yields slightly lower LPIPS and DISTS scores.
We therefore re-optimize \method with $\eta=6/255$, obtaining a DISTS score of approximately 0.19, comparable to I2VGuard~\cite{i2vguard} and Anti-I2V~\cite{antii2v}.
We also evaluate AdvPaint~\cite{advpaint}, the second-best baseline in Table~\ref{tab:opensora_matched_imperceptibility}, at a comparable perceptual distortion level.
As shown in Table~\ref{tab:opensora_matched_imperceptibility}, \method still demonstrates superior protection performance even when compared under similar perceptual imperceptibility.

\vspace{-1.mm}
\subsection{Extension to Universal Adversarial Perturbation}
\vspace{-1.mm}
\method introduces a novel source grounding detachment objective for I2V immunization, but image-specific optimization remains costly due to backpropagation through the I2V model.
To explore an extension to universal adversarial perturbations (UAPs)~\cite{uap} that protect different images without per-image optimization, we train UAP variants of \method and all baselines.
We train all UAP variants on 250 VBench~\cite{vbench} images disjoint from the evaluation set, using CogVideoX~\cite{cogvideox} as the surrogate with a shared $\ell_\infty$ budget of $8/255$.
Table~\ref{tab:uap} shows that \method leads all UAP variants across all three metrics, outperforming even image-specific EditShield~\cite{editshield} and I2VGuard~\cite{i2vguard}.
Despite a gap from its image-specific counterpart, these results support our objective's applicability to universal image protection.

\begin{figure*}[t]
\centering
    \includegraphics[width=\linewidth]{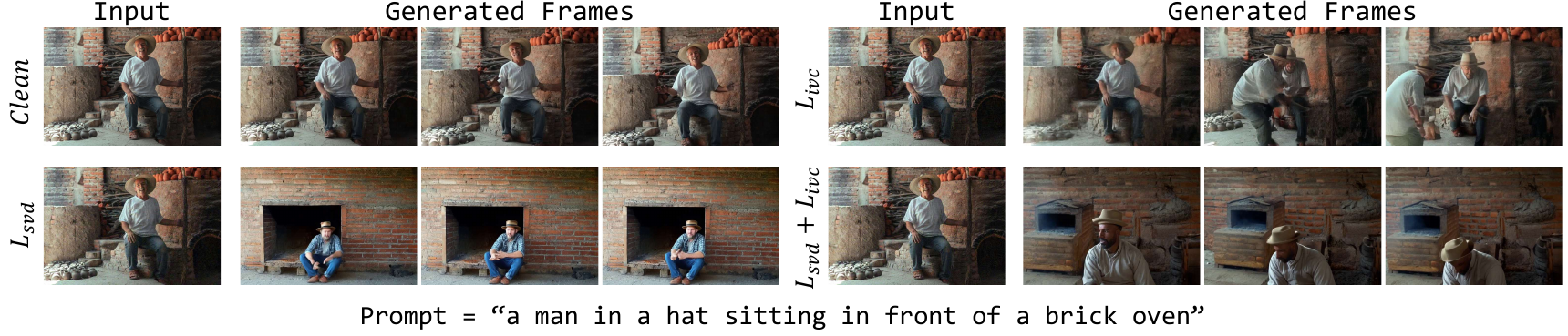}
    \vspace{-0.6cm}
    \caption{Qualitative ablation for the proposed two loss functions on CogVideoX~\cite{cogvideox}.}
    \label{fig/qual_loss}
    \vspace{-0.2cm}
\end{figure*}
\begin{table*}[t]
\footnotesize
\centering
\setlength{\tabcolsep}{5pt}
\caption{Ablation studies for the losses and video frames.
\emph{Black}, \emph{Source}, and \emph{Rand} denote repeated black frames, repeated source frames, and random noisy latents at the corresponding timestep, respectively, while \emph{SDV} denotes the proposed source-divergent video.}
\vspace{-2mm}
\resizebox{\linewidth}{!}{
\begin{tabular}{c|cc;{0.2pt/1pt}cc|cc;{0.2pt/1pt}cc|cc;{0.2pt/1pt}cc}
\Xhline{0.35mm}

Method
& \multicolumn{4}{c|}{$\mathcal{L}_{\mathrm{svd}}$}
& \multicolumn{4}{c|}{$\mathcal{L}_{\mathrm{ivc}}$}
& \multicolumn{4}{c}{$\mathcal{L}_{\mathrm{svd}}$+$\mathcal{L}_{\mathrm{ivc}}$} \\

\cline{1-13}
\multirow{2}{*}{Video}
& \multicolumn{2}{c;{0.2pt/1pt}}{\emph{Unrelated}}
& \multicolumn{2}{c|}{\emph{Related}}
% & \multicolumn{1}{c|}{\emph{Hybrid}}
& \multicolumn{2}{c;{0.2pt/1pt}}{\emph{Unrelated}}
& \multicolumn{2}{c|}{\emph{Related}}
% & \multicolumn{1}{c|}{\emph{Hybrid}}
& \multicolumn{2}{c;{0.2pt/1pt}}{\emph{Unrelated}}
& \multicolumn{2}{c}{\emph{Related}}\\
% & \multicolumn{1}{c}{\emph{Hybrid}}\\

\cline{2-13}
& Rand & Black & Source & SDV
& Rand & Black & Source & SDV
& Rand & Black & Source & SDV \\

\Xhline{0.35mm}

IS $\downarrow$
& 95.29 & 95.06 & \underline{92.83} & \cellcolor{green!15}{\textbf{91.83}} 
& 90.09 & \underline{89.39} & 90.38 & \textbf{\cellcolor{green!15}{89.31}}  
& 89.75 & 89.14 & \underline{88.45} & \cellcolor{green!15}{\textbf{86.39}}\\

SC $\downarrow$
& 91.59 & 90.98 & \underline{90.49} & \textbf{\cellcolor{green!15}{87.49}} 
& 85.80 & \textbf{85.07} & 86.50 & \underline{\cellcolor{green!15}{85.37}}
& 84.41 & 84.23 & \underline{83.54} & \cellcolor{green!15}{\textbf{80.63}} \\

DS $\downarrow$
& 83.16 & 82.55 & \underline{80.93} & \textbf{\cellcolor{green!15}{78.24}} 
& 74.77 & \underline{73.03} & 75.32 & \cellcolor{green!15}{\textbf{72.37}}  
& 72.62 & 72.60 & \underline{70.78} & \cellcolor{green!15}{\textbf{65.54}} \\

\Xhline{0.35mm}
\end{tabular}
}
\vspace{-4mm}
\label{tab:loss-video}
\end{table*}

\vspace{-1mm}
\subsection{Ablation Studies}
\vspace{-1mm}
We conduct ablation experiments to examine the effectiveness of the proposed components.
We use CogVideoX~\cite{cogvideox} as the victim model.
% and only report I2V Subject, Subject Consistency, and DINO similarity, which are aligned with our immunization objective.

\vspace{-1.5mm}
\paragraph{Impact of Loss Functions.}
\vspace{-0.5mm}
We ablate the two losses, $\mathcal{L}_{\mathrm{svd}}$ and $\mathcal{L}_{\mathrm{ivc}}$.
As shown in Table~\ref{tab:loss-video}, under our proposed source-divergent video (SDV) setting, each loss improves over the all baselines as in Table~\ref{tab:white}, while their combination achieves the best performance, confirming their complementary roles.
Qualitatively, Figure~\ref{fig/qual_loss} shows that $\mathcal{L}_{\mathrm{svd}}$ alone detaches the generated video from the source to some extent, but the result may still preserve source-like identity cues.
In contrast, $\mathcal{L}_{\mathrm{ivc}}$ alone increases video dynamics but does not detach the generation from the source.
The two losses, jointly designed for source grounding detachment, produce stronger source detachment when combined.
\vspace{-0.1cm}
% We analyze the immunization effectiveness of the proposed two loss functions, $\mathcal{L}_{\mathrm{svd}}$ and $\mathcal{L}_{\mathrm{ivc}}$.
% As shown by the \emph{SDV} results in Table~\ref{tab:loss-video}, optimizing either loss alone already improves over the baselines in Table~\ref{tab:white}, indicating that both source-video suppression and intra-video coupling contribute to immunization.
% Moreover, combining the two losses yields the strongest performance, showing that they address complementary aspects of source grounding detachment.
% These results support our design of a joint objective rather than relying on either term independently.
% We further provide qualitative ablation results and discussion in Appendix~\ref{sec:loss_abl_qual}.

% \input{table/decomposition}

% \input{table/video_frame}
\vspace{-1.5mm}
\paragraph{Video Latent Construction.}
We investigate the effect of video initialization on \method.
As shown in Table~\ref{tab:loss-video}, $\mathcal{L}_{\mathrm{svd}}$ is effective mainly with source-related videos, whereas $\mathcal{L}_{\mathrm{ivc}}$ is relatively insensitive to the video type.
% Compared with using $\mathcal{L}_{\mathrm{ivc}}$, furthermore, $\mathcal{L}_{\mathrm{svd}}$+$\mathcal{L}_{\mathrm{ivc}}$ yields a large gain with source-related videos, but only marginal improvements with source-unrelated videos.
Moreover, $\mathcal{L}_{\mathrm{svd}}+\mathcal{L}_{\mathrm{ivc}}$ substantially improves over $\mathcal{L}_{\mathrm{ivc}}$ with source-related videos, but yields only marginal gains with source-unrelated videos.
This is expected, since our objective is not to drive generation toward source-unrelated videos, but to detach it from source grounding.
This indicates that source-related videos provide the necessary source-grounding signals for $\mathcal{L}_{\mathrm{svd}}$ to meaningfully suppress source reliance, while our source-divergent videos offer more diverse video-side signals than simple repeated source images.

\vspace{-2mm}
\section{Conclusion}
\vspace{-2mm}
We introduced a novel I2V immunization objective \emph{source grounding detachment}, shifting the goal from temporal degradation to disrupting source-guided visual grounding.
We presented \method, which prevents unauthorized video generation by suppressing source-video responses and amplifying intra-video interactions, thereby weakening the source-dominant attention pathway.
We further introduced source-divergent video construction, which provides diverse source-oriented signals for our objective.
Extensive experiments across diverse diffusion-based I2V models demonstrate that our method consistently achieves stronger immunization performance than existing baselines.
% Unlike prior approaches that primarily induce temporal degradation or generic corruption, our method 
% To this end, our method effectively disrupts the source-dominant attention pathway by weakening source-video responses and amplifying intra-video interactions.
% The source-divergent videos further support that source grounding detachment is a key mechanism for effective I2V immunization.
% To this end, we introduce two complementary objectives: \svloss, which suppresses source-image dependence in video latents, and \vvloss, which reinforces video-token interactions to prevent denoising from returning to source-guided visual grounding.
% We further constructed pseudo-video frames from the source image, enabling optimization without pre-generated reference videos while providing source-originated latent variations.

% 

{
\small
\bibliographystyle{plain}
\bibliography{references} 
}
\newpage

\addtocontents{toc}{\protect\setcounter{tocdepth}{2}}
\newpage 
\appendix
\begin{center}
\makeatletter
\@toptitlebar
    {\bfseries\Large \method:Preventing Unauthorized Image-to-Video Generations via source grounding Detachment\par}
    % \vspace{-1mm}
    {\large\itshape -Supplementary Materials-\par}
    % \vspace{-2mm}
    \@bottomtitlebar
    % \textcolor{blue!90}{
% \large \textit{We have provided generated videos from images immunized by both the baselines and our method in the supplementary material for a comprehensive comparison. \\ We hope these examples would be helpful.}
% }
\makeatother
\end{center}

\begingroup
\renewcommand{\clearpage}{}
\linespread{0.8}\selectfont
\addtocontents{toc}{\protect\vspace{-6mm}}

{\small
\makeatletter
\@starttoc{toc}
\makeatother
}
\endgroup

% \makecontents
\newpage

\section{Experimental Details}
\label{sec:exp_detail}

\subsection{Implementation Setup}
\label{sec:imple}
% \paragraph{Immunization Settings.}
For our method, we set the step sizes for $\delta$ and $\Delta$ to $1/255$ and $2/255$, respectively.
We use the $25$--$75\%$ layer range across all I2V models.
Since our objective targets video-token interactions commonly present in I2V models, we apply it to the corresponding attention module of each architecture without additional model-specific design: full attention in CogVideoX~\cite{cogvideox}, self-attention in LTX-Video~\cite{ltx}, and temporal attention in OpenSora 1.2~\cite{opensora}.
For a fair comparison, we set the perturbation budget to $\eta=8/255$ and use $50$ attack iterations.
We conduct all experiments on an NVIDIA A100 GPU, and the optimization time is approximately 10 minutes with 62~GB memory.

We sample diffusion timesteps from the first $10\%$ of the denoising trajectory, where coarse video structure is established under strong source-image guidance~\cite{alg}.
By intervening before the denoising trajectory becomes stabilized, \method more effectively disrupts source grounding and prevents the generated video from faithfully preserving the source-image content.

\subsection{Threat Models}
\label{app:model}
We consider I2V models with diverse attention architectures--MMDit full-attention, MMDit cross-attention, and Dit-based--to evaluate the generality of \method.
We select representative models from each architecture that can be reasonably used for immunization on a single NVIDIA A100 80GB GPU, considering factors such as model size and the number of frames required for high-quality video generation.

\paragraph{Full-attention.}
In full-attention I2V models, condition, text, and video tokens are jointly processed within a unified attention module, allowing direct interaction across all modalities.
Representative models include Hunyuan-Video~\cite{hunda} and OpenSora 2.0~\cite{opensora2}.
In this category, we use CogVideoX-5B~\cite{cogvideox} as the primary model.

For CogVideoX-5B~\cite{cogvideox}, text tokens precede visual tokens.
Let $N_{\mathrm{text}}$ denote the number of text tokens, $s_p$ the number of tokens per latent frame, and $F$ the number of latent frames.
We instantiate the losses as follows:

\begin{itemize}[leftmargin=10pt]
    \item \textbf{Source-video decoupling.}
    We use video-side queries from frames $1,\dots,F-1$ and source/condition keys from frame $0$:
    \[
    Q_{\mathcal{V}} = q_{\ell,h}[N_{\mathrm{text}}+s_p:N_{\mathrm{text}}+F s_p],
    \quad
    K_{\mathcal{C}} = k_{\ell,h}[N_{\mathrm{text}}:N_{\mathrm{text}}+s_p].
    \]

    \item \textbf{Intra-video coupling.}
    We use both queries and keys from the video-side frames $1,\dots,F-1$:
    \[
    Q_{\mathcal{V}} = q_{\ell,h}[N_{\mathrm{text}}+s_p:N_{\mathrm{text}}+F s_p],
    \quad
    K_{\mathcal{V}} = k_{\ell,h}[N_{\mathrm{text}}+s_p:N_{\mathrm{text}}+F s_p].
    \]
\end{itemize}

\paragraph{Cross-attention.}
Cross-attention-based I2V models separate video latent interactions from conditioning signals, where self-attention organizes video latents while cross-attention injects source and text information.
Representative models include Wan~\cite{wan} family.
We use LTX-Video-2B~\cite{ltx} in our experiments.

For LTX-Video-2B~\cite{ltx}, we apply our losses to self-attention over frame-major visual tokens.
Let $s_p$ denote the number of tokens per latent frame and $F$ the number of latent frames.
We instantiate the losses as follows:
\begin{itemize}[leftmargin=10pt]
    \item \textbf{Source-video decoupling.}
    We use video-side queries from frames $1,\dots,F-1$ and source/condition keys from frame $0$:
    \[
    Q_{\mathcal{V}} = q_{\ell,h}[s_p:F s_p],
    \quad
    K_{\mathcal{C}} = k_{\ell,h}[0:s_p].
    \]

    \item \textbf{Intra-video coupling.}
    We use both queries and keys from the video-side frames $1,\dots,F-1$:
    \[
    Q_{\mathcal{V}} = q_{\ell,h}[s_p:F s_p],
    \quad
    K_{\mathcal{V}} = k_{\ell,h}[s_p:F s_p].
    \]
\end{itemize}
\paragraph{DiT-based spatio-temporal attention.}
DiT-based I2V models often decompose spatial and temporal processing into separate attention modules, alternating between spatial and temporal blocks.
Representative models include OpenSora variants.
In this paper, we use OpenSora 1.2~\cite{opensora}, where spatial attention captures within-frame structure, temporal attention propagates information across frames, and cross-attention incorporates conditioning signals.

For OpenSora 1.2~\cite{opensora}, we compute the losses on temporal attention.
Let $T$ denote the number of latent frames.
We instantiate the losses as follows:
\begin{itemize}[leftmargin=10pt]
    \item \textbf{Source-video decoupling.}
    We use video-side queries $1,\dots,F-1$ and source/condition keys from frame $0$:
    \[
    Q_{\mathcal{V}} = q_{\ell,h}[1:F],
    \quad
    K_{\mathcal{C}} = k_{\ell,h}[0:1].
    \]

    \item \textbf{Intra-video coupling.}
    We use both queries and keys from the video-side frames $1,\dots,T-1$:
    \[
    Q_{\mathcal{V}} = q_{\ell,h}[1:F],
    \quad
    K_{\mathcal{V}} = k_{\ell,h}[1:F].
    \]
\end{itemize}
\subsection{Evaluation Dataset and Metrics}
\label{app:data}
\paragraph{Dataset.}
To ensure a reliable evaluation setting and improve the credibility of the measured performance, we build our evaluation set using 100 image-prompt pairs sampled from VBench~\cite{vbench}, consisting of 80 human-related images and 20 non-human images such as animals or natural scenes.
Since our evaluation metrics are also adopted from VBench series~\cite{vbench, vbench++}, using samples from the same benchmark provides a more consistent evaluation protocol.

\paragraph{Metrics.}
We evaluate immunization performance using five metrics: \textit{I2V Subject} (IS), \textit{Subject Consistency} (SC), DINO similarity (DS), \textit{Aesthetic Quality} (AQ), and \textit{Imaging Quality} (IQ), where IS, SC, AQ, and IQ are oriented from VBench.
Lower IS, SC, and DS indicate stronger immunization, while AQ and IQ are reported to monitor the perceptual quality of generated videos.

\textbf{I2V Subject (IS).}
IS measures how strongly the generated video preserves the subject of the source image while maintaining temporal coherence.
For a source image $x$ and generated frames $\{v_t\}_{t=1}^{T}$, we compute the DINO-based cosine similarities between the source image and each frame,
\begin{align}
c_t = \max\!\big(0,\cos(\phi(x),\phi(v_t))\big),
\end{align}
and between consecutive frames,
\begin{align}
r_t = \max\!\big(0,\cos(\phi(v_{t-1}),\phi(v_t))\big), \qquad t=2,\dots,T.
\end{align}
The final IS score is defined as
\begin{align}
\mathrm{IS}
=
0.4 \cdot \max_t c_t
+
0.3 \cdot \frac{1}{T-1}\sum_{t=2}^{T} r_t
+
0.3 \cdot \min_t r_t.
\end{align}

\textbf{Subject Consistency (SC).}
SC measures temporal subject coherence within the generated video.
We compute the average, over all frames, of the mean cosine similarity to the first frame and to the immediately preceding frame:
\begin{align}
\mathrm{SC}
=
\frac{1}{T-1}\sum_{t=2}^{T}
\frac{
\max\!\big(0,\cos(\phi(v_1),\phi(v_t))\big)
+
\max\!\big(0,\cos(\phi(v_{t-1}),\phi(v_t))\big)
}{2}.
\end{align}

\textbf{DINO Similarity (DS).}
DS directly measures how strongly the generated frames preserve the visual content of the source image.
It is computed as the average cosine similarity between the source image and each generated frame in the DINO feature space:
\begin{align}
\mathrm{DS}
=
\frac{1}{T}\sum_{t=1}^{T}
\max\!\big(0,\cos(\phi(x),\phi(v_t))\big).
\end{align}

\textbf{Aesthetic Quality (AQ).}
AQ measures the perceptual aesthetic quality of the generated video. Following VBench, we compute frame-level aesthetic scores using the LAION aesthetic predictor and average them over all generated frames:
\begin{align}
\mathrm{AQ}
=
\frac{1}{T}\sum_{t=1}^{T} a(v_t),
\end{align}
where $a(\cdot)$ denotes the aesthetic scoring model.

\textbf{Imaging Quality (IQ).}
IQ measures the low-level visual quality of the generated video.
Following VBench, we compute MUSIQ scores for each generated frame and average them over the video:
\begin{align}
\mathrm{IQ}
=
\frac{1}{T}\sum_{t=1}^{T} q(v_t),
\end{align}
where $q(\cdot)$ denotes the MUSIQ-based image quality predictor.

\subsection{Baseline Selection and Implementation Details}

Image immunization has been actively studied in I2I generation~\cite{raising, advcross, advpaint, blurguard, semantic, myface, diffguard}.
Among prior methods, we adapt EditShield~\cite{editshield} and AdvPaint~\cite{advpaint} as representative baselines, since they effectively disrupt how diffusion models interpret the source image at the encoder and attention levels, respectively.
In the I2V immunizations, we select two state-of-the-art I2V baselines, I2VGuard~\cite{i2vguard} and Anti-I2V~\cite{antii2v}.
Detailed explanations are provided below.
 
\paragraph{EditShield.}
EditShiled~\cite{editshield} first introduced an image protection method against instruction-guided image editing models such as InstructPix2Pix~\cite{ip2p}.
Since such models repeatedly use the latent of the conditioning image during the editing process, EditShield~\cite{editshield} maximizes the $\ell_2$ distance between the clean latent and the perturbed latent to disrupt faithful source-image interpretation.
To improve robustness, it further incorporates EOT~\cite{eot} during perturbation optimization.
The perturbation is updated by PGD~\cite{pgd}, following their original implementation~\cite{editshield}.

\paragraph{AdvPaint.}
To mitigate the risks of misusing image inpainting, AdvPaint~\cite{advpaint} proposed immunization method against diffusion-based inpainting.
It perturbs attention-related features of the diffusion denoiser so that the model misinterprets the source image during generation.
Specifically, we perturb the query, key, and value features of the condition image in CogVideoX~\cite{cogvideox}; the condition-image self-attention query, key, and value as well as the cross-attention query in LTX-Video~\cite{ltx}; and the query, key, and value features in spatial blocks for Open-Sora~\cite{opensora}.
In our implementation, following the original setup, we use a word representing the source image (\eg, \textit{man} or \textit{woman}) as the prompt when computing the cross-attention query and for residual update to deeper layers.
The perturbation is updated by PGD~\cite{pgd}, following their original implementation~\cite{advpaint}.

\paragraph{I2VGuard.}
I2VGuard~\cite{i2vguard} is one of the first methods to explicitly address image immunization against I2V generation.
It optimizes perturbations using multiple objectives, including an encoder attack, alignment of the perturbed attention map with that of a fixed target video (\ie, a black video), and alignment of the denoised video latent with the same target, in order to induce temporal degradation and reduce generation fidelity.
At the same time, it employs a contrastive objective that pushes the denoised video latent of the perturbed input away from that of the pre-generated clean outputs.
The perturbation is updated by gradient descent algorithm, following their original implementation~\cite{i2vguard}.

\paragraph{Anti-I2V.} 
To prevent malicious I2V generation, Anti-I2V~\cite{antii2v} introduced IRC loss, which encourages shallow intermediate representations to collapse toward those of deeper layers, thereby inducing information loss, and IRA loss, which aligns the latent and deep features with those of a target to promote temporal degradation and reduce generation fidelity.
To improve robustness against purification, it further proposes dual-space perturbation, which jointly optimizes perturbations in both the Lab and DCT domains.
The perturbation is updated by AdamW optimizer, following their original implementation~\cite{antii2v}.
% \subsection{Transformation parameters}
% Table~\ref{tab:eot_params} summarizes the transformations used to construct source-like video latents.
% To preserve source-related content while avoiding overly strong distortion, we set the range of each transformation conservatively.
% \input{table/eot}

\subsection{Details of Purification}
We consider two categories of purification methods.
The first consists of transformation-based methods, including JPEG compression~\cite{jpeg}, Gaussian noise, and Adverse Cleaner~\cite{advclean}, which are commonly used to suppress high-frequency adversarial perturbations.
The second is diffusion-based purification method that directly target adversarial perturbations in diffusion models.
In this category, we use GridPure~\cite{gridpure}, an effective diffusion-based purification method designed for high-resolution images.
% Since GridPure is designed for high-resolution images, while the original DiffPure~\cite{diffpure} may undesirably damage the original content of perturbed images, we adopt GridPure instead of the original DiffPure in our evaluation.

\paragraph{JPEG Compression.}
JPEG compression~\cite{jpeg} is originally designed to reduce image storage and transmission costs through quantization in the DCT domain.
This quantization process can cause information loss in carefully crafted perturbations, particularly those concentrated in high-frequency components.

\paragraph{Gaussian Noise}
Gaussian noise is one of the simplest image degradation methods, but it can still weaken adversarial perturbations by directly corrupting the crafted perturbation pattern with random noise, thereby partially overwriting the perturbation signal embedded in the image.

\paragraph{Adverse Cleaner}
% As used in prior works~\cite{dct, diffguard}, crop and resize is an effective technique which only crops the boundary regions so that most of the original visual content is preserved, and the cropped image is then resized back to the original resolution.
% Because this resizing step involves interpolation, perturbation signals are partially attenuated or removed.
Adverse cleaner~\cite{advclean} is an effective purifier that removes high-frequency artifacts by repeatedly applying bilateral filtering and then refining the result with guided filtering, while preserving the main visual structure of the original image.

% \paragraph{Conditional DiffPure}
% Conditional DiffPure~\cite{condiffpure} extends DiffPure~\cite{diffpure} to diffusion-targeted adversarial perturbations by incorporating text conditioning, whereas the original DiffPure uses an unconditional diffusion process.
% This conditional design helps restore the original image semantics during purification.

\paragraph{GridPure}
GridPure~\cite{gridpure} is designed to remove adversarial perturbations from high-resolution images by dividing the image into grids, performing slicing and merging, and applying small step DiffPure~\cite{diffpure} iteratively to each grid.

\subsection{Details of Human Evaluation}
\label{app:human}

We provide additional details for the human evaluation described in Section~\ref{sec:human}.
We conduct the evaluation on crowdsourcing with 20 participants.
For each I2V model and each method, we randomly sample 10 generated videos for evaluation.
The evaluation is designed to assess not only whether the generated videos preserve the source image, but also whether the resulting failure mode is effective from a protection perspective.

\begin{figure*}[h]
\centering
    \includegraphics[width=\linewidth]{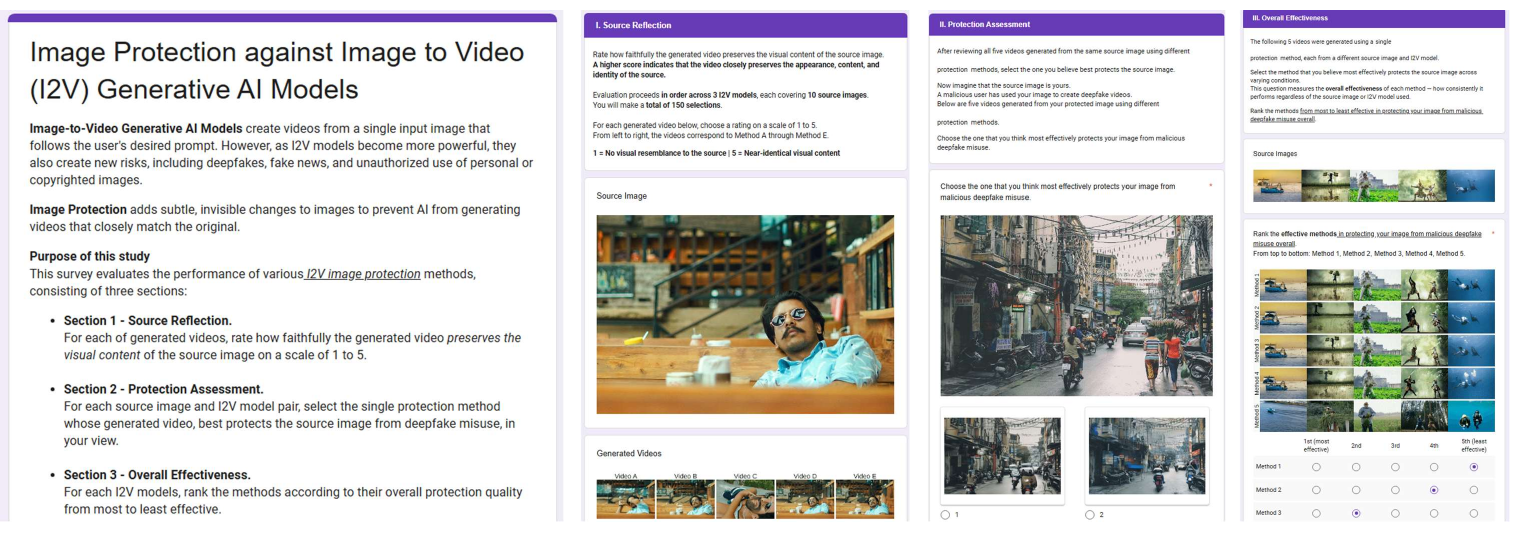}
    \vspace{-5mm}
    \caption{Screenshots of the conducted human evaluation.}
    \label{fig/human}
\end{figure*}

Evaluators assess the results along the following three dimensions:
\begin{itemize}
    \item {Source Reflection}: Evaluators rate how faithfully the generated video preserves the visual content of the source image on a 1--5 scale. A lower score indicates weaker source-image reflection and therefore stronger protection.

    \item {Protection Assessment}: Evaluators select the generated output that is most effective from a protection perspective. This question is designed to compare different immunization objectives from a protection perspective, assessing whether an output that suppresses source-image reflection is preferred over one that mainly exhibits temporal degradation or fidelity reduction.

    \item {Overall Effectiveness}: Evaluators rank the methods according to their overall effectiveness as image protection methods. This dimension evaluates which method provides the strongest protection when considering the generated results as a whole.
\end{itemize}
Importantly, since {protection assessment} and {overall effectiveness} require participants' subjective judgments, we did not provide any instruction that could favor our method, such as asking them to ``select the video that differs most from the source image''. 
Snapshots of the human evaluation interface is shown in Figure~\ref{fig/human}.

\section{Additional Analysis}
\subsection{Attention Mass Shift}
\label{app/att-mass}
% =========================
% CogVideoX
% =========================
\begin{table}[h]
\footnotesize
\centering
\caption{Quantification of attention mass on CogVideoX~\citep{cogvideox}.
\emph{Source-video} attention mass is measured between video queries and source keys, while \emph{video-video} attention mass is measured among video-side tokens.}
% \vspace{-3mm}
\resizebox{0.75\linewidth}{!}{
\begin{tabular}{c|c;{0.2pt/1pt}ccccc}
\Xhline{0.35mm}
Attention
& Clean & ES & AP & IG & AI & \method \\
\hline

{Source-video} &
0.087 & 0.085 & 0.090 & 0.080 & 0.063 &
\textbf{0.015} \\

{video-video} &
0.666 & 0.671 & 0.637 & 0.679 & 0.660 &
\textbf{0.710} \\

\Xhline{0.35mm}
\end{tabular}
}
\vspace{-0.3cm}
\label{tab:att-mass-cog}
\end{table}

% =========================
% LTX-Video
% =========================
\begin{table}[h]
\footnotesize
\centering
\caption{Quantification of attention mass on LTX-Video~\citep{ltx}.}
% \vspace{-3mm}
\resizebox{0.75\linewidth}{!}{
\begin{tabular}{c|c;{0.2pt/1pt}ccccc}
\Xhline{0.35mm}
Attention
& Clean & ES & AP & IG & AI & \method \\
\hline

{Source-video} &
0.083 & 0.073 & 0.076 & 0.081 & 0.075 &
\textbf{0.051} \\

{video-video} &
0.917 & 0.927 & 0.924 & 0.919 & 0.925 &
\textbf{0.949} \\

\Xhline{0.35mm}
\end{tabular}
}
\vspace{-0.3cm}
\label{tab:att-mass-ltx}
\end{table}

% =========================
% OpenSora 1.2
% =========================
\begin{table}[h]
\footnotesize
\centering
\caption{Quantification of attention mass on OpenSora 1.2~\citep{opensora}.}
% \vspace{-3mm}
\resizebox{0.75\linewidth}{!}{
\begin{tabular}{c|c;{0.2pt/1pt}ccccc}
\Xhline{0.35mm}
Attention
& Clean & ES & AP & IG & AI & \method \\
\hline

{Source-video} &
0.237 & 0.220 & 0.282 & 0.246 & 0.233 &
\textbf{0.178} \\

{video-video} &
0.763 & 0.780 & 0.718 & 0.754 & 0.767 &
\textbf{0.822} \\

\Xhline{0.35mm}
\end{tabular}
}
% \vspace{0.1cm}
\label{tab:att-mass-opensora}
\end{table}
We quantitatively measure the shift in attention mass from video queries to condition-image keys and among video-side tokens using a single denoising pass with video latents generated from the clean source image, following the visualization setting in Fig.~\ref{fig/attention}.

As shown in Table~\ref{tab:att-mass-cog},~\ref{tab:att-mass-ltx}, and~\ref{tab:att-mass-opensora}, \method substantially reduces the source-video attention mass compared with the clean image and baseline methods.
At the same time, \method effectively increases the video-video attention mass, indicating that denoising video latents rely less on condition-image keys and more on video-side interactions, even on clean-generated video.
These results quantitatively support our objective: the protected condition image shifts denoising toward video-side interactions, preventing the trajectory from being re-anchored to the source image.

\subsection{Analysis of Source-divergent Video}
\paragraph{Black.}
\emph{Black} provides video latents that are clearly separated from the conditioning image.
While this can make the source-video response weak, the resulting signal is less meaningful because the video-side latents are not source-related states induced by the protected image.
Moreover, although {\vvloss} loss can still amplify interactions among black video latents, these interactions arise from source-uninformative states rather than source-conditioned video-side continuations.
This explains why, under \emph{Black}, combining both losses performs similarly to using {\vvloss} loss alone in Table~\ref{tab:loss-video}: {\svloss} loss provides limited additional benefit when the video-side latents are already far from the source.

\paragraph{Source.}
\emph{Source} uses video latents that are well aligned with the conditioning image, providing meaningful source-related states for {\svloss} loss.
This makes it more suitable than \emph{Black} for suppressing source-video grounding within the source-conditioned I2V process, as shown in Table~\ref{tab:loss-video}.
However, simply repeating the source image produces redundant video-side states with little diversity.
As a result, {\vvloss} loss can mainly reinforce source-copy interactions rather than non-trivial video-side continuations, limiting the overall effectiveness compared with source-divergent video (SDV).

\paragraph{Source-divergent Video.}
\emph{SDV} better satisfies the requirements of both losses.
It starts from a source-oriented state, making it suitable for {\svloss} loss, while gradually introducing video-side diversity, making it more effective for {\vvloss} loss.
This balance allows SDV to preserve meaningful source-video grounding signals for suppression while providing sufficiently diverse video-side states for amplifying intra-video interactions.
Consequently, SDV achieves the strongest performance under the full objective in Table~\ref{tab:loss-video}.

\section{Additional Experiments}

\subsection{Effect of Layer Selection}
In the main experiments, we use the 25--75\% layer range, which covers an intermediate subset of layers with sufficient representational depth.
To provide a more comprehensive analysis of layer selection, we additionally evaluate two alternative ranges, 0--50\% and 50--100\%.
As shown in Table~\ref{tab:layer}, the intermediate 25--75\% range achieves the strongest immunization performance, yielding the lowest IS, SC, and DS scores.
In contrast, using earlier layers only (0--50\%) degrades performance, while using later layers only (50--100\%) further weakens the immunization effect.
These results suggest that \method is most effective when the intervention spans a moderate range of layers.
Rather than relying exclusively on shallow or deep layers, using both moderately shallow and moderately deep layers better captures the hierarchical interactions involved in source-conditioned video generation.
Accordingly, we use the 25--75\% range as our main setting in all experiments.
\begin{table}[h]
\centering
\begin{minipage}[h]{0.4\linewidth}
\footnotesize
\centering
\setlength{\tabcolsep}{4pt}
\caption{Ablation studies on layer ratio. * denotes our main setting.}
\resizebox{0.95\linewidth}{!}{
\begin{tabular}{l|ccc}
\Xhline{0.35mm}
Method 
& \multicolumn{3}{c}{Ours} \\
\hline
Layer Ratio
& IS $\downarrow$ & SC $\downarrow$ & DS $\downarrow$ \\
\hline

25--75\% *
& \textbf{86.39} & \textbf{80.63} & \textbf{65.54} \\

0--50\%
& 87.91 & 82.71 & 69.36 \\

50--100\%
& 89.43 & 84.85 & 72.95 \\
\Xhline{0.35mm}
\end{tabular}
}
\vspace{0.1cm}
\label{tab:layer}
\end{minipage}
\hfill
\begin{minipage}[h]{0.50\linewidth}
\footnotesize
\centering
\setlength{\tabcolsep}{4pt}
\caption{Ablation experiments on perturbation budgets, where * denotes the main setting.}
\resizebox{0.8\linewidth}{!}{
\begin{tabular}{l|ccc}
\Xhline{0.35mm}
Method 
& \multicolumn{3}{c}{Ours} \\
\hline
Budget
& IS $\downarrow$ & SC $\downarrow$ & DS $\downarrow$ \\
\hline
$\eta=4/255$
& 92.30 & 88.00& 77.68 \\

$\eta=8/255$ *
& 86.39 & 80.63 & 65.54 \\

$\eta=16/255$
& 81.24 & 76.77 & 61.63 \\
\Xhline{0.35mm}
\end{tabular}
}
\vspace{0.1cm}
\label{tab:budget}
\end{minipage}
\end{table}
\begin{table}[h]
\footnotesize
\centering
\caption{Ablation study on source-deviated video latent optimization strategies. \textit{Ours-P} denotes pre-optimized source-divergent video latents, Ours-I optimizes all $F-1$ tail latents independently, and Ours* optimizes a single source-deviated tail latent and replicates it across the remaining positions.}
% \vspace{-2mm}
\resizebox{0.55\linewidth}{!}{
\begin{tabular}{l|ccc;{0.2pt/1pt}c}
\Xhline{0.35mm}
Method
% & \multicolumn{4}{c}{Ours} \\
% \hline
& IS $\downarrow$ & SC $\downarrow$& DS $\downarrow$& Latency(sec) \\
\hline

Ours-P 
 & 88.72 & 83.42 & 70.73 & 475.94 \\
 
Ours-I 
 & 86.30 & 80.18 & 64.65 & 1,344.50 \\

Ours* 
 & 86.39 & 80.63 & 65.54 & 620.72 \\

\Xhline{0.35mm}
\end{tabular}
}
\vspace{0.1cm}
\label{tab:efficiency}
\end{table}
\subsection{Perturbation Budgets}
We conduct experiments under different perturbation budgets to analyze the sensitivity of our method to the perturbation magnitude.
The results in Table~\ref{tab:budget} show that \method achieves competitive immunization performance even with a small budget of $\eta=4/255$, compared with existing methods in Table~\ref{tab:white}.
As the perturbation budget increases, the immunization performance improves consistently.
% The qualitative results in Figure~\ref{fig:eps} further show that larger budgets lead to stronger disruption of source-image grounding.
In particular, we use $\eta=8/255$ as the default setting, our method provides the strongest protection at the commonly acceptable image-specific adversarial budget of $\eta=16/255$.
This allows the perturbation budget to be flexibly chosen according to the characteristics of the image to be protected, such as its frequency content, and the desired protection goal, as suggested in prior work~\cite{nearly}.

\subsection{Source-divergent Video Optimization}
\label{app:pseudo}
In our main setting, for efficiency, we optimize only a single source-divergent video latent and replicate it across the remaining video-side positions, rather than optimizing all $F-1$ source-divergent video independently, as mentioned in Section~\ref{sec:sdv}.
This design is motivated by the role of the source-divergent video in our framework: it is not intended to produce a fully optimized video itself, but to provide a source-originated video-side signal for Eq.~\ref{eq:loss_svd}.

To verify whether this simplification is sufficient, we compare it with a more expensive variant that optimizes all $F-1$ source-divergent video frames independently.
We also evaluate \textit{Ours-P}, which uses source-divergent videos pre-optimized with the full steps, to test whether our small-step iterative construction better matches the objective than a fixed pre-optimized signal.
As shown in Table~\ref{tab:efficiency}, optimizing each source-divergent video frame separately provides only marginal improvement over the replicated variant, suggesting that our objective mainly requires a source-related video-side latent signal rather than position-wise optimization.
In contrast, \textit{Ours-P} slightly degrades performance, likely because fully pre-optimized videos remain fixed source-oriented signals with limited diversity, rather than providing progressively source-divergent guidance.
Therefore, we use the single-latent optimization and replication strategy as the default setting in all experiments.
% As shown in Table~\ref{tab:efficiency}, optimizing each source-divergent video frame separately provides only marginal improvement over the replicated variant.
% This result suggests that our objective mainly benefits from establishing a source-related video-side latent signal, rather than requiring position-wise optimization.
% Therefore, we use the single-latent optimization and replication strategy as the default setting in all experiments.

% \input{table/layer}
\begin{figure*}[h]
\centering
    \includegraphics[width=0.45\linewidth]{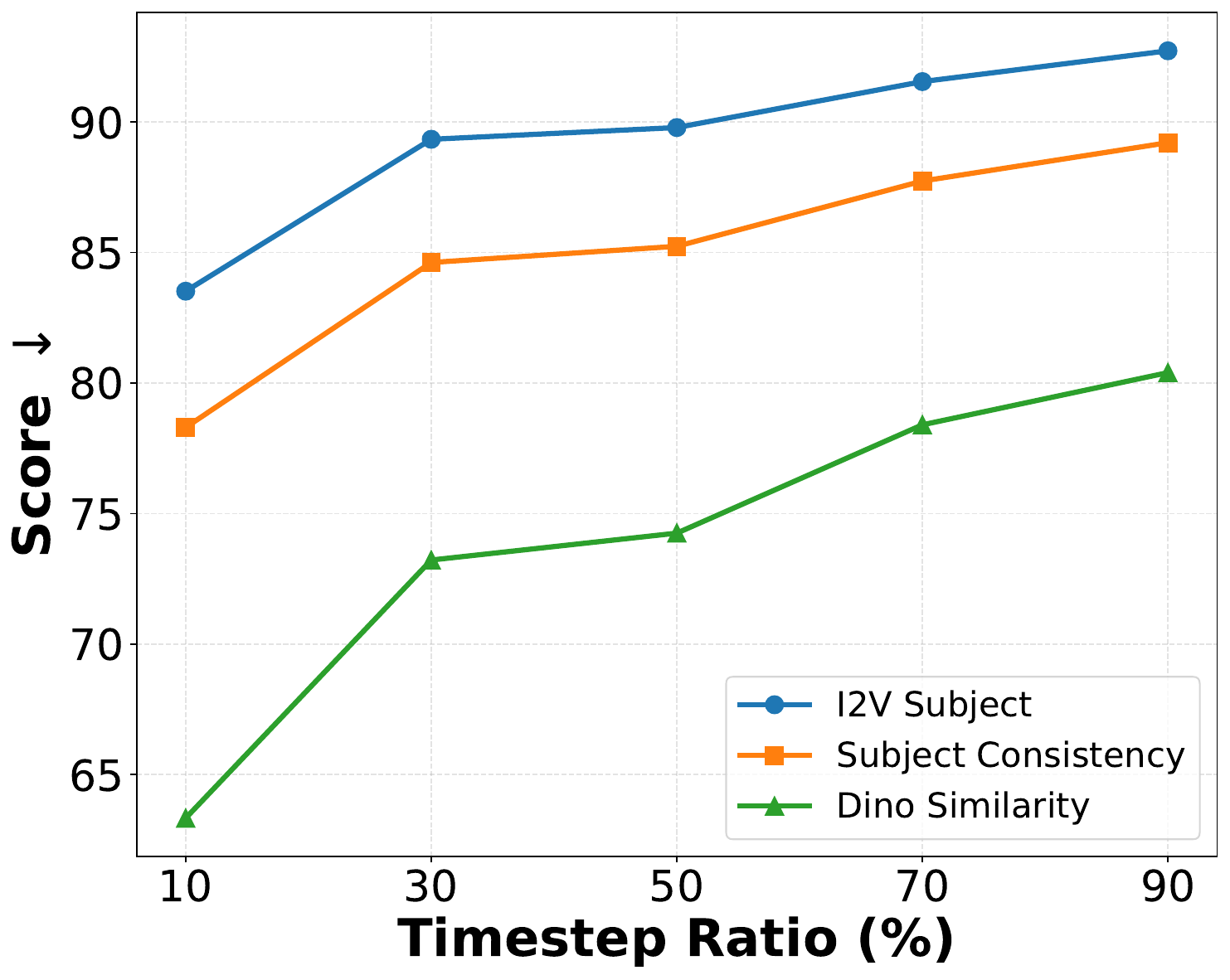}
    \vspace{-1mm}
    \caption{Ablation study on timestep sampling. As the sampled timestep range becomes larger, immunization performance degrades.}
    \label{fig/timestep}
\end{figure*}
\subsection{Range of Timestep Sampling}
We optimize perturbations using the first 10\% of the denoising steps, where the coarse video structure is established under strong source guidance, by uniformly sampling timesteps within this range.
To study the effect of timestep selection, we additionally conduct experiments with progressively larger timestep ranges.
As shown in Figure~\ref{fig/timestep}, immunization performance decreases as the sampled timestep range expands.
This suggests that source grounding detachment is particularly effective in early denoising stages, where coarse semantic structure is formed under strong source guidance, and is consistent with our method intervening in this formation process.

% \input{table/dct}
% \subsection{Optimization on DCT Domain}
% The primary goal of this paper is to establish an effective objective for image immunization against diffusion-based image-to-video generation.
% However, robustness to purification has also become increasingly important from a real-world application perspective~\cite{blurguard, dct, antii2v}.
% To examine the extensibility of our method to the DCT domain, which has been adopted in prior purification-robust designs, we further optimize perturbations in the DCT space using Eq.~\ref{eq:objective}, following DCT-Shield~\cite{dct} and Anti-I2V~\cite{antii2v}.
% As shown in Table~\ref{tab:dct}, incorporating DCT perturbations consistently improves robustness against purification, even accounting for the slight performance gain already observed without purification.
% These results suggest that our method is not confined to the RGB space, but can be naturally extended to purification-aware formulations used in prior robust protection methods.

\begin{table}[t]
\footnotesize
\centering
\setlength{\tabcolsep}{5pt}
\caption{Effect of video latent construction variations.}
\label{tab:video_frame}
\resizebox{0.55\linewidth}{!}{
\begin{tabular}{c|ccccc}
\Xhline{0.35mm}
Videos & Source & Trans. & Gen. & Gen.+G. & SDV \\
\Xhline{0.35mm}
IS $\downarrow$ & 88.45 & 88.12 & \underline{87.49} & 87.96 & \cellcolor{green!15}{\textbf{86.39}} \\
SC $\downarrow$ & 83.54 & 83.07 & \underline{82.54} & 82.56 & \cellcolor{green!15}{\textbf{80.63}} \\
DS $\downarrow$ & 70.78 & 70.09 & {67.43} & \underline{67.30} & \cellcolor{green!15}{\textbf{65.54}} \\
\Xhline{0.35mm}
\end{tabular}
}
\vspace{0.1cm}
\label{tab:video_frame2}
\end{table}

% \begin{table}[t]
% \footnotesize
% \centering
% \setlength{\tabcolsep}{5pt}
% \caption{Effect of video latents. \textit{Noise} uses the scheduler-derived noisy latent at the target timestep; \textit{Black} and \textit{Source} repeat a black image and the source image, respectively; \textit{Trans.} uses transformed source images; and \textit{Gen.} and \textit{PVF} use generated video frames and the proposed pseudo-video frames.}
% % \vspace{-2mm}
% \resizebox{0.65\linewidth}{!}{
% \begin{tabular}{c|cc|cccc}
% \Xhline{0.35mm}
% \multirow{2}{*}{Videos} & \multicolumn{2}{c|}{\textit{Source-unrelated}} & \multicolumn{4}{c}{\textit{Source-related}} \\
% \cline{2-7}
% & Noise & Black & Source & Trans. & Gen. & SDV \\
% \Xhline{0.35mm}
% IS $\downarrow$ & 89.75 & 89.14 & 88.45 & 88.12 & \underline{87.49} & \textbf{86.39} \\
% SC $\downarrow$ & 84.41 & 84.23 & 83.54 & 83.07 & \underline{82.54} & \textbf{80.63} \\
% DS $\downarrow$ & 72.62 & 72.60 & 70.78 & 70.09 & \underline{67.43} & \textbf{65.54} \\
% \Xhline{0.35mm}
% \end{tabular}
% }
% \vspace{0.1cm}
% \label{tab:video_frame}
% \end{table}
\subsection{Optimization with Diverse Video}
% \note{
We further explore diverse source-related video constructions.
A simple source repetition preserves the visual content of the protected image, but it does not account for temporal variation in I2V generation.
We therefore consider additional source-related variants: \textit{Trans.}, which uses transformed source images; \textit{Gen.}, which uses pre-generated videos following the conventional setup of previous I2V immunization methods~\cite{i2vguard, antii2v}; and \textit{Gen.+G.}, which adds Gaussian noise to the pre-generated videos with the same step size as our source-deviated video construction.
In particular, \textit{Gen.+G.} is designed to verify whether adding a simple stochastic deviation to generated video latents can satisfy both source-relatedness and source deviation, reflecting the motivation behind our source-divergent video construction.

As shown in Table~\ref{tab:video_frame2}, \textit{Trans.} slightly improves over the repeated \textit{Source} baseline, indicating that source-preserving variation is more effective than static source repetition.
Using generated frames in \textit{Gen.} further improves performance, suggesting that latents drawn from the I2V model better reflect its source-conditioned generation distribution than simple image transformations.
However, \textit{Gen.+G.} achieves performance similar to \textit{Gen.} or even degraded.
This suggests that simply injecting stochastic noise into generated latents is insufficient to obtain useful source-diversified video-side states.
In contrast, \textit{SDV} consistently achieves the best results across all metrics, indicating that our optimized construction provides a more suitable more \method.
\subsection{Additional Black-box Transferability}
\label{sec:black-box}
% }

\paragraph{Cross-model Transferability.}
Following transferability evaluation optimized on CogVideoX~\citep{cogvideox} in Table~\ref{tab:black}, we additionally conduct cross-transferability experiments in Table~\ref{tab:black_transfer_supp} using LTX-Video~\citep{ltx} and OpenSora 1.2~\citep{opensora} as surrogate models.
Across these transfer settings, \method consistently achieves the strongest overall protection performance, demonstrating that its effectiveness is not limited to a specific surrogate model.

\paragraph{Commercial and Large-scale Models.}
Evaluating image immunization on commercial and large-scale I2V models is particularly important for assessing its practical effectiveness, yet remains highly challenging due to limited access to model internals, including their architectures, training data, and image preprocessing pipelines.
To investigate how effective current I2V immunization methods remain under such realistic black-box conditions, we further evaluate transferability to the commercial Luma Ray 3.2~\citep{lumaray32} and the large-scale Wan 2.1-14B~\citep{wan} using randomly selected 30 images in our evaluation dataset.
Despite these substantial uncertainties and distributional differences, the result in Table~\ref{tab:commercial_large_transfer} shows that \method consistently achieves the strongest protection performance among the compared approaches on both models, demonstrating its effectiveness beyond the open-source models used during optimization.

\begin{table*}[t]
\footnotesize
\centering
\caption{Additional black-box transferability evaluation on LTX-Video~\cite{ltx} and OpenSora 1.2~\cite{opensora}.}
\vspace{-2mm}
\resizebox{\linewidth}{!}{
\begin{tabular}{l|ccc|ccc||ccc|ccc}
\Xhline{0.35mm}
Surrogate Model
& \multicolumn{6}{c||}{LTX-Video}
& \multicolumn{6}{c}{OpenSora 1.2} \\
\hline
Target Model
& \multicolumn{3}{c|}{CogVideoX}
& \multicolumn{3}{c||}{OpenSora 1.2}
& \multicolumn{3}{c|}{CogVideoX}
& \multicolumn{3}{c}{LTX-Video} \\
\hline
Metric
& IS $\downarrow$ & SC $\downarrow$ & DS $\downarrow$
& IS $\downarrow$ & SC $\downarrow$ & DS $\downarrow$
& IS $\downarrow$ & SC $\downarrow$ & DS $\downarrow$
& IS $\downarrow$ & SC $\downarrow$ & DS $\downarrow$ \\
\hline
EditShield%~\cite{editshield}
& 95.02 & 93.31 & 85.22
& 93.78 & 93.29 & 81.82
& 94.52 & 94.31 & 84.29
& 93.32 & 90.73 & 75.52 \\

AdvPaint%~\cite{advpaint}
& 94.74 & 93.65 & 84.63
& 94.28 & 94.42 & 83.00
& 93.38 & 93.03 & 81.84
& 92.03 & 90.05 & 72.10 \\

I2VGuard%~\cite{i2vguard}
& 96.64 & 94.21 & 88.13
& 95.61 & 95.18 & 87.06
& 94.31 & 93.27 & 83.20
& 92.93 & 91.19 & 75.34 \\

Anti-I2V%~\cite{antii2v}
& 96.68 & 94.52 & 88.24
& 95.44 & 94.84 & 85.95
& 92.69 & 91.83 & 82.03
& 92.11 & 90.73 & 74.48 \\

\rowcolor{green!15}
Ours
& \textbf{93.47} & \textbf{92.95} & \textbf{82.15}
& \textbf{92.89} & \textbf{92.01} & \textbf{80.09}
& \textbf{91.33} & \textbf{90.80} & \textbf{79.83}
& \textbf{90.96} & \textbf{87.92} & \textbf{69.74} \\
\Xhline{0.35mm}
\end{tabular}
}
% \vspace{-5mm}
\label{tab:black_transfer_supp}
\end{table*}
\begin{table}[t!]
\footnotesize
\centering
\caption{Transferability evaluation on commercial and large-scale models.}
% \vspace{-4mm}
\resizebox{0.65\linewidth}{!}{
\begin{tabular}{l|ccc|ccc}
\Xhline{0.35mm}
Model
& \multicolumn{3}{c|}{Luma Ray}
& \multicolumn{3}{c}{Wan2.1-14B} \\
\hline
Method
& IS $\downarrow$ & SC $\downarrow$ & DS $\downarrow$
& IS $\downarrow$ & SC $\downarrow$ & DS $\downarrow$ \\
\hline
% \multicolumn{7}{l}{\textit{Non-optimized}} \\
Clean
& 95.52 & 94.72 & 83.55
& 97.28 & 96.20 & 92.37 \\
% Gauss.
% & 94.86 & 93.81 & 82.57
% & 96.48 & 95.90 & 91.89 \\
% \hline
% \multicolumn{7}{l}{\textit{Optimized}} \\
EditShield%~\cite{editshield}
& 93.90 & 92.84 & 79.92
& 95.19 & 95.95 & 88.25 \\
AdvPaint%~\cite{advpaint}
& 94.39 & 93.52 & 81.60
& 95.39 & 96.19 & 88.70 \\
I2VGuard%~\cite{i2vguard}
& 93.79 & 92.87 & 79.53
& 94.59 & 95.65 & 86.85 \\
Anti-I2V%~\cite{antii2v}
& 93.75 & 90.99 & 81.52
& 93.90 & 95.46 & 85.38 \\
\rowcolor{green!15}
Ours
& \textbf{93.42} & \textbf{90.65} & \textbf{78.61}
& \textbf{92.45} & \textbf{94.02} & \textbf{83.25} \\
\Xhline{0.35mm}
\end{tabular}
}
% \vspace{-5mm}
\label{tab:commercial_large_transfer}
\end{table}

\section{Qualitative Results}
\label{sec:qual_comparison}
\subsection{Comparison with Baselines}
\label{sec:qual_comparison_wb}
\begin{figure*}[t]
\centering
    \includegraphics[width=\linewidth]{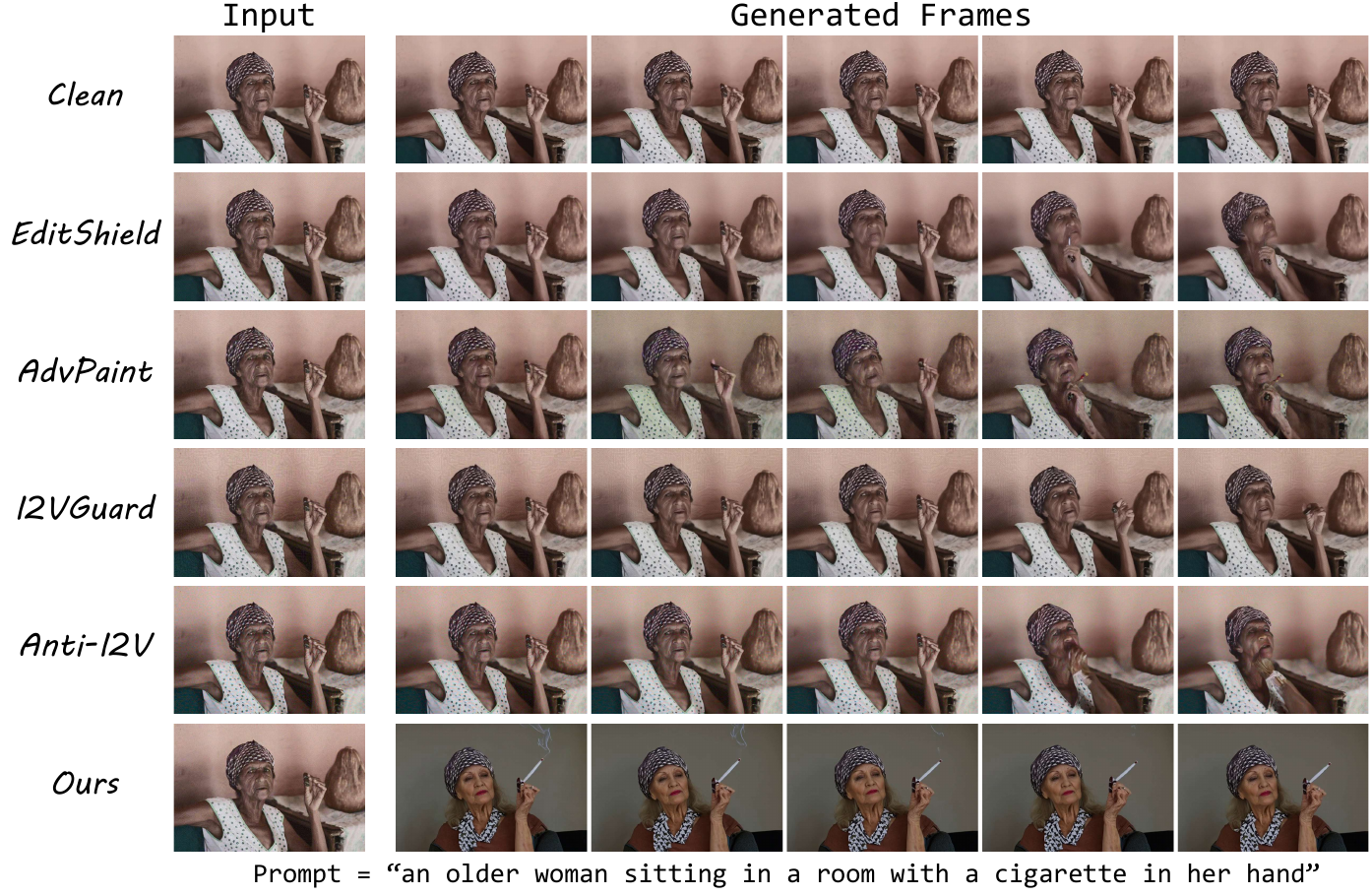}
    
    \includegraphics[width=\linewidth]{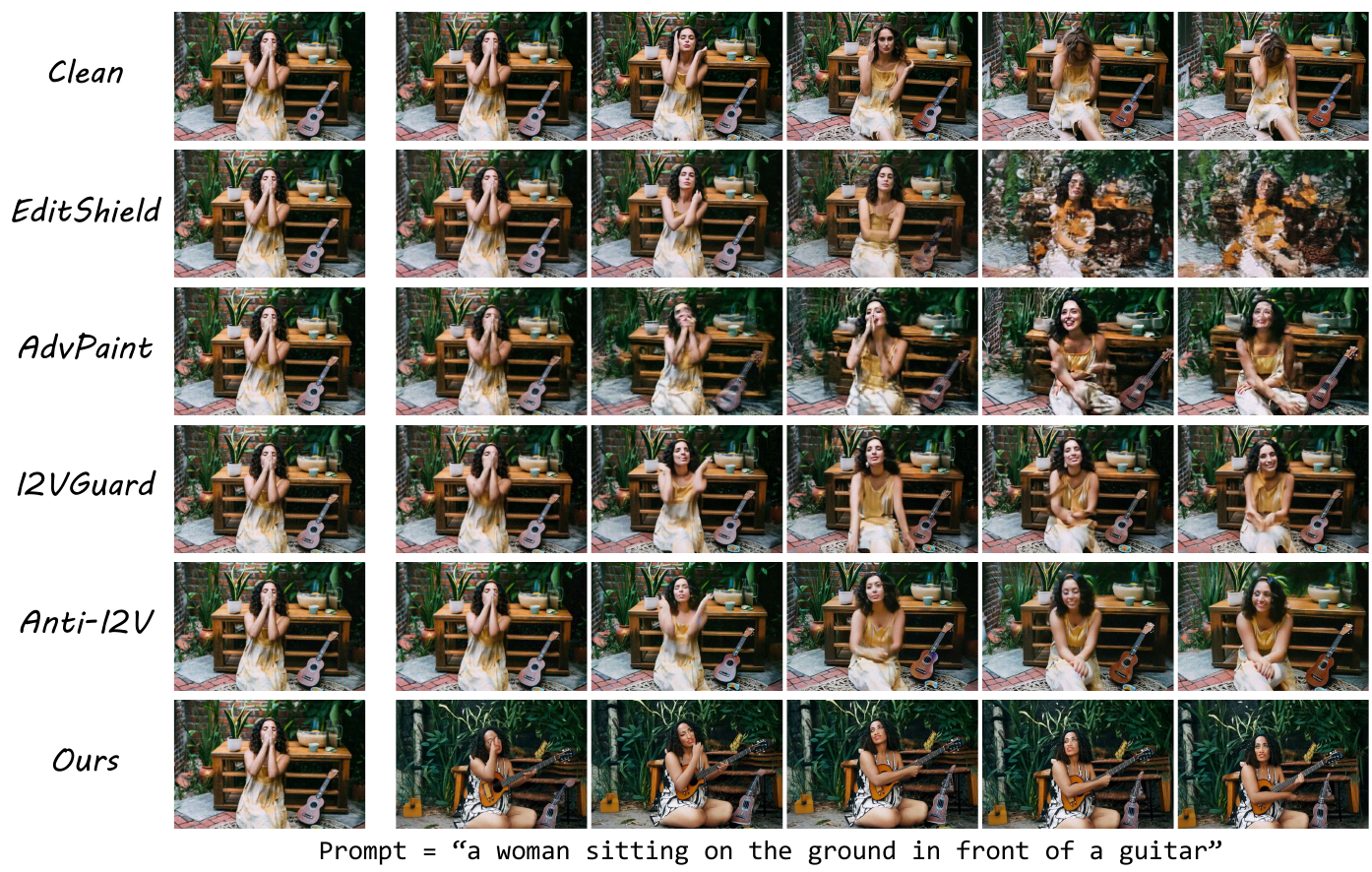}
    \vspace{-0.6cm}
    \caption{Qualitative comparison with baselines on CogVideoX~\cite{cogvideox}.}

    \label{fig/cog_wb}
\end{figure*}

\begin{figure*}[t]
\centering
    \includegraphics[width=\linewidth]{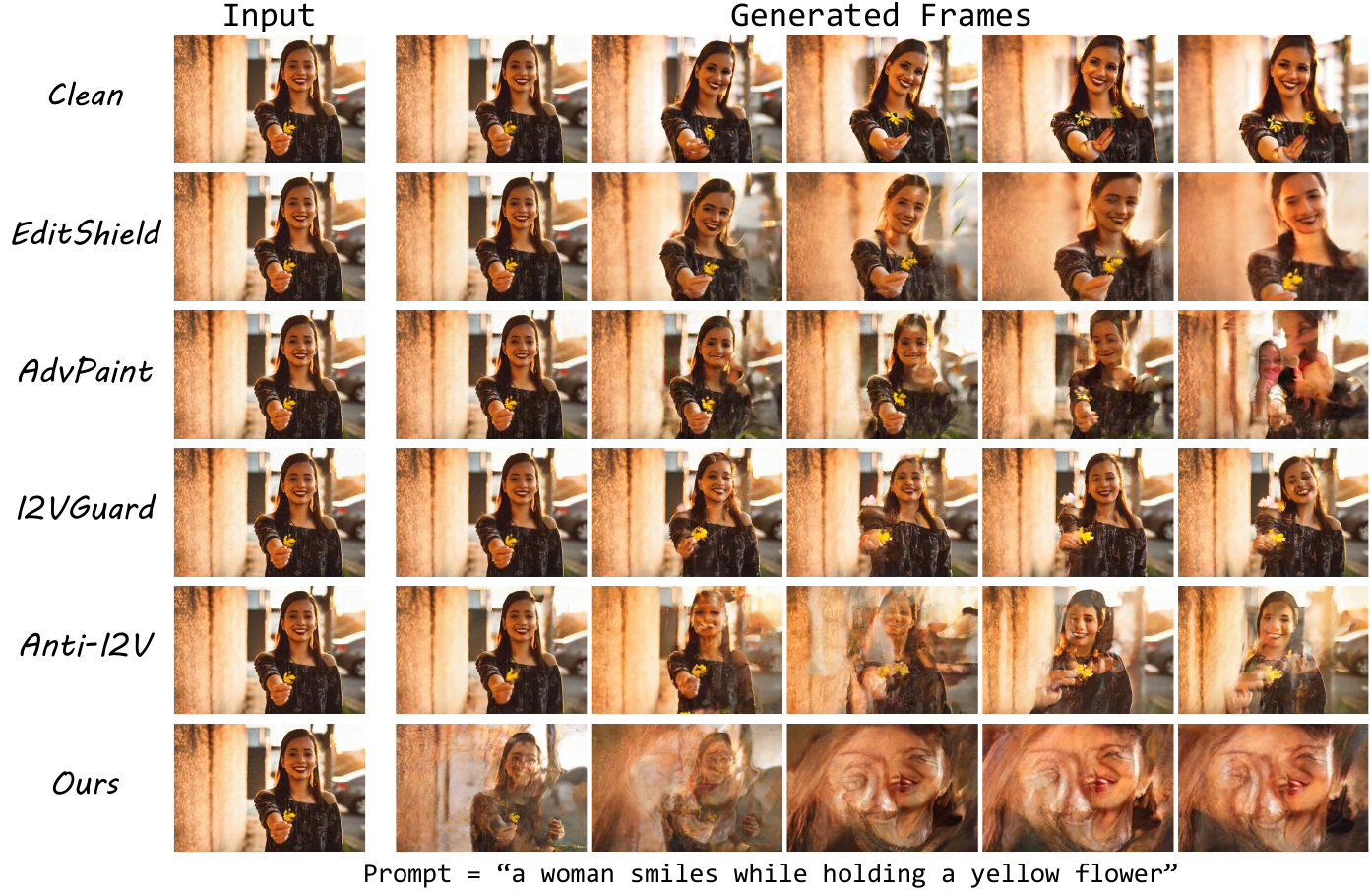}
    
    \includegraphics[width=\linewidth]{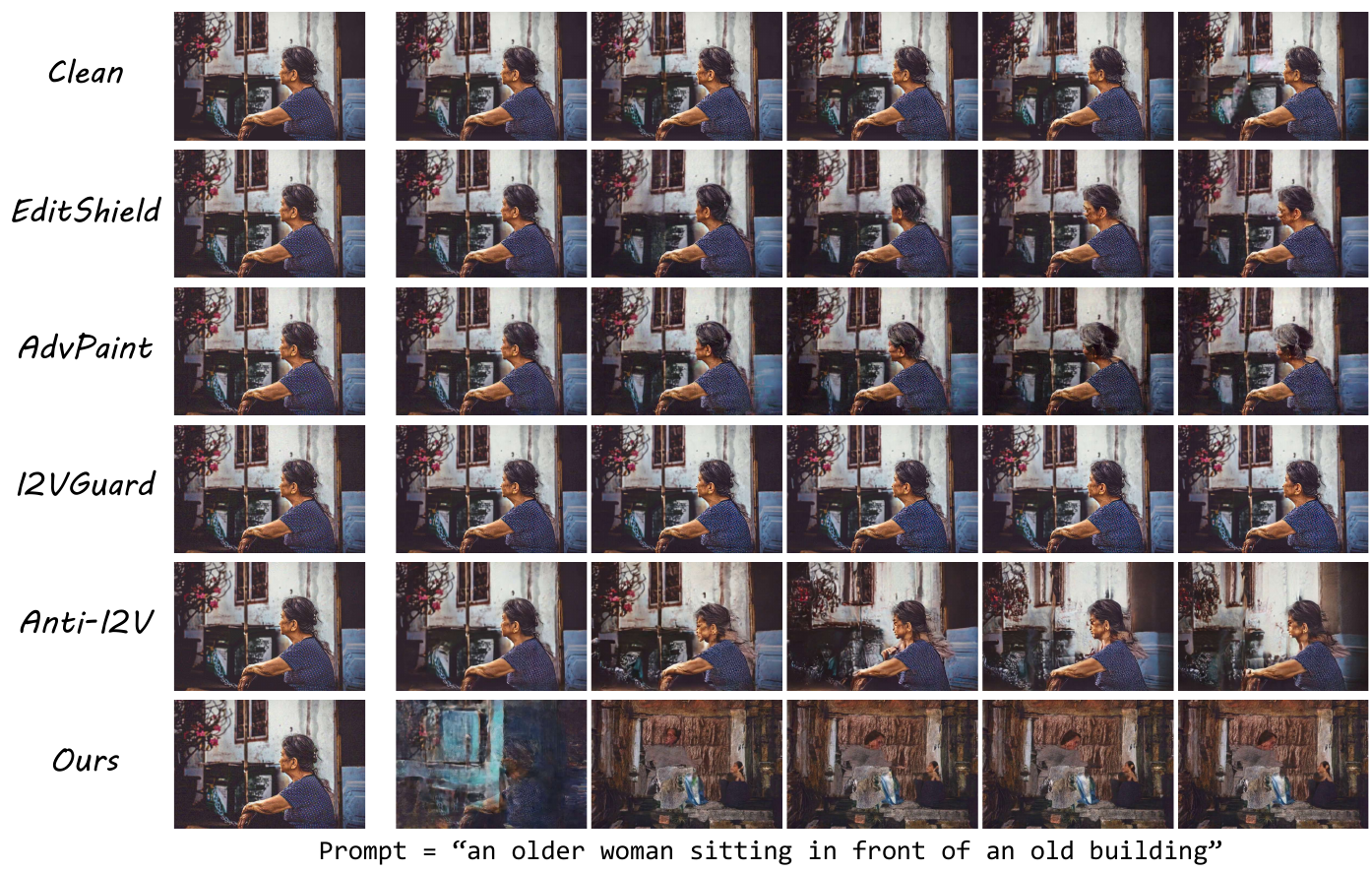}
    \vspace{-0.6cm}
    \caption{Qualitative comparison on LTX-Video~\cite{ltx}.}

    \label{fig/ltx_wb}
\end{figure*}

\begin{figure*}[t]
\centering
    \includegraphics[width=\linewidth]{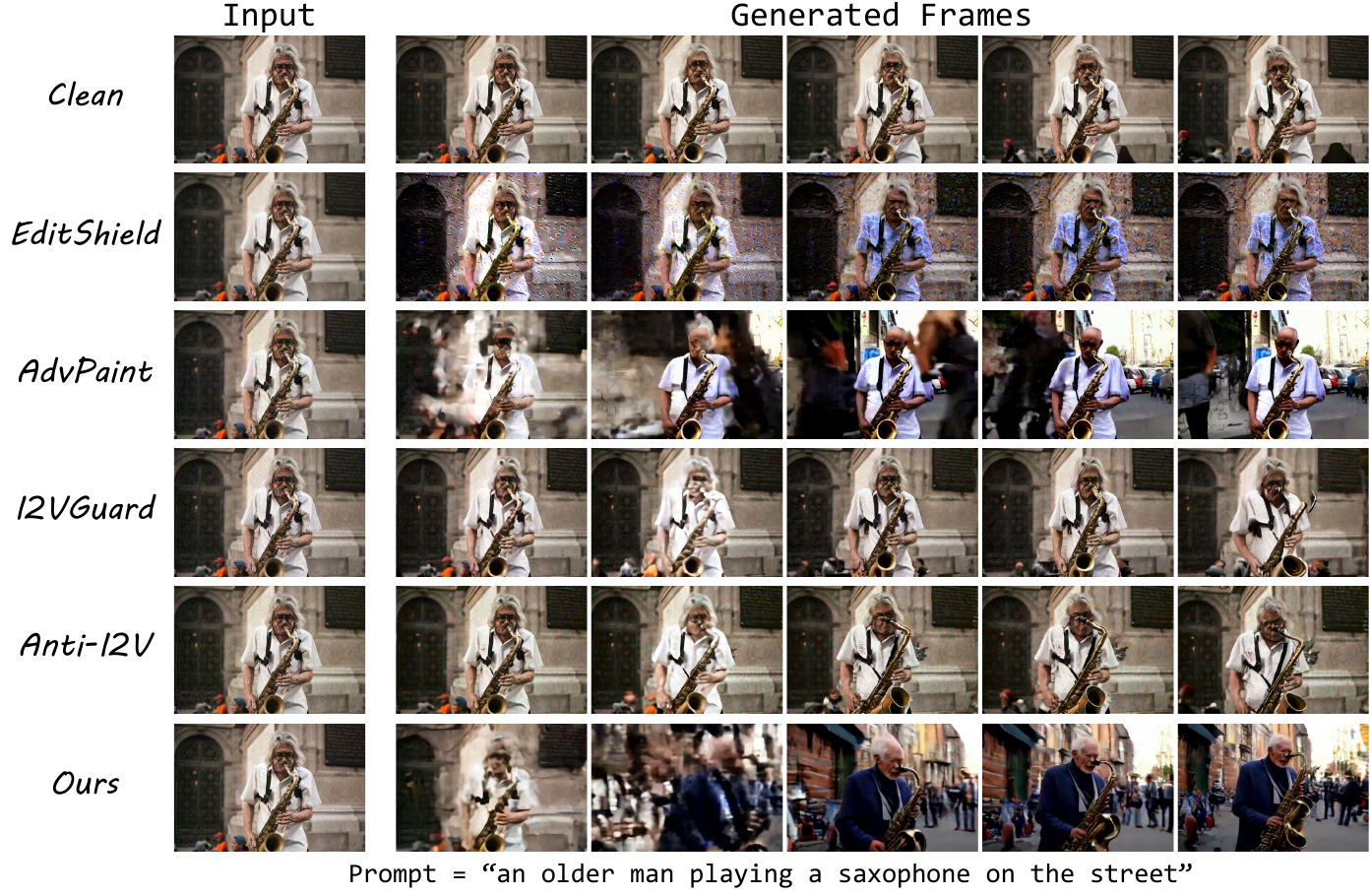}
    
    \includegraphics[width=\linewidth]{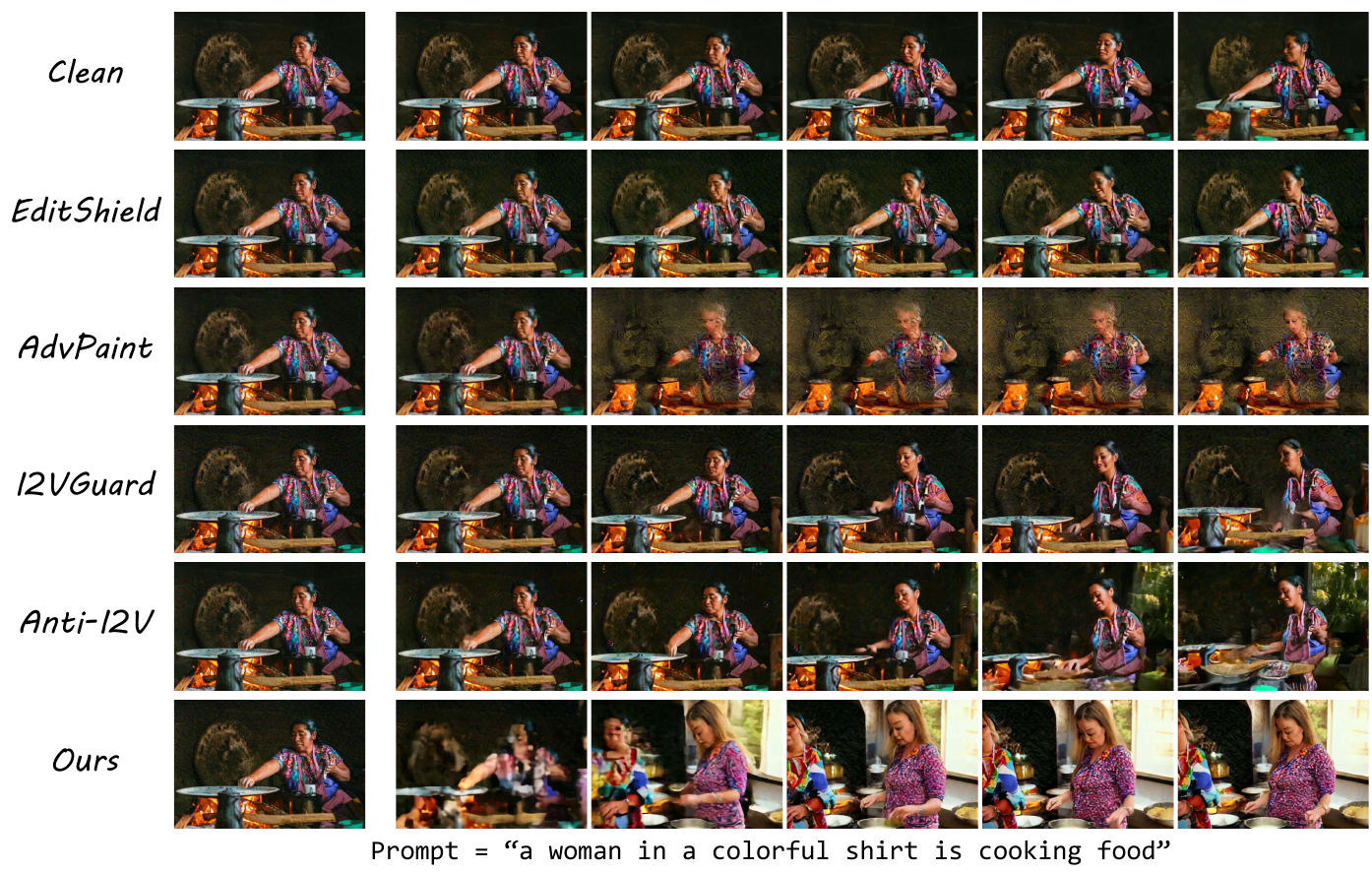}
    \vspace{-0.6cm}
    \caption{Qualitative comparison on OpenSora 1.2~\cite{opensora}.}

    \label{fig/sora_wb}
\end{figure*}

We provide qualitative comparisons with other immunization methods on CogVideoX~\cite{cogvideox}, LTX-Video~\cite{ltx}, and OpenSora 1.2~\cite{opensora}.
% Figures~\ref{fig/cog_wb},~\ref{fig/ltx_wb}, and~\ref{fig/sora_wb} show that, while the baselines effectively induce degradation in the generated videos, they often still preserve source-related visual content.
Figures~\ref{fig/cog_wb},~\ref{fig/ltx_wb}, and~\ref{fig/sora_wb} show that, although the baselines effectively induce degradation in the generated videos, they often fail to prevent the source visual content from remaining aligned with the generated frames.
In contrast, \method directly blocks such alignment, so that the generated video no longer retains recognizable visual cues from the source image, rather than merely exhibiting degraded quality.

Interestingly, the resulting failure patterns differ across architectures, which we attribute to the different roles of the targeted attention modules in each model.
In CogVideoX~\cite{cogvideox}, our objective mainly affects visual-token interactions, including condition-image tokens, weakening source-to-video continuity while preserving prompt-level semantics and generating prompt-consistent but source-detached videos.
In OpenSora 1.2~\cite{opensora}, we intervene on the temporal attention branch, which is followed by cross-attention and mainly governs information flow across frames rather than spatial structure. As a result, the perturbation often leads to protection patterns similar to those observed in CogVideoX.
In contrast, for LTX-Video~\cite{ltx}, we target self-attention layers that directly organize video-latent interactions across both spatial and temporal dimensions, so weakening it more readily causes global collapse rather than only continuity failure.
Despite these architecture-dependent differences, all such patterns share the same underlying outcome: the generated videos are no longer reliably grounded in the source image.
% 여기에 ltx랑 cog open이 다른 이유 적기. 공격하는 block이 달라서 그런 거 같긴한데

\subsection{Qualitative Results Beyond Human-Related Images}
\begin{figure*}[t]
\centering
    \includegraphics[width=\linewidth]{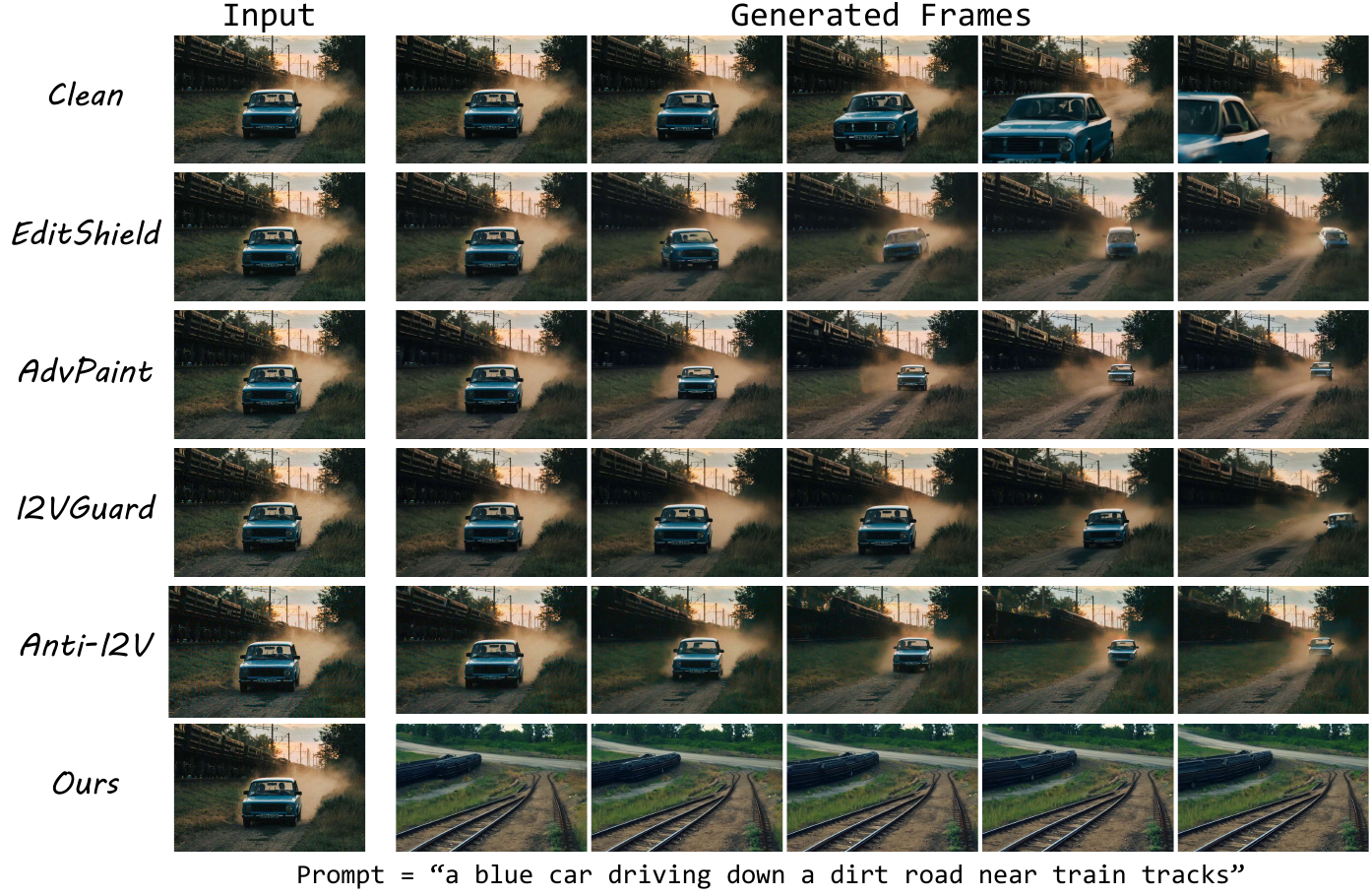}
    
    \includegraphics[width=\linewidth]{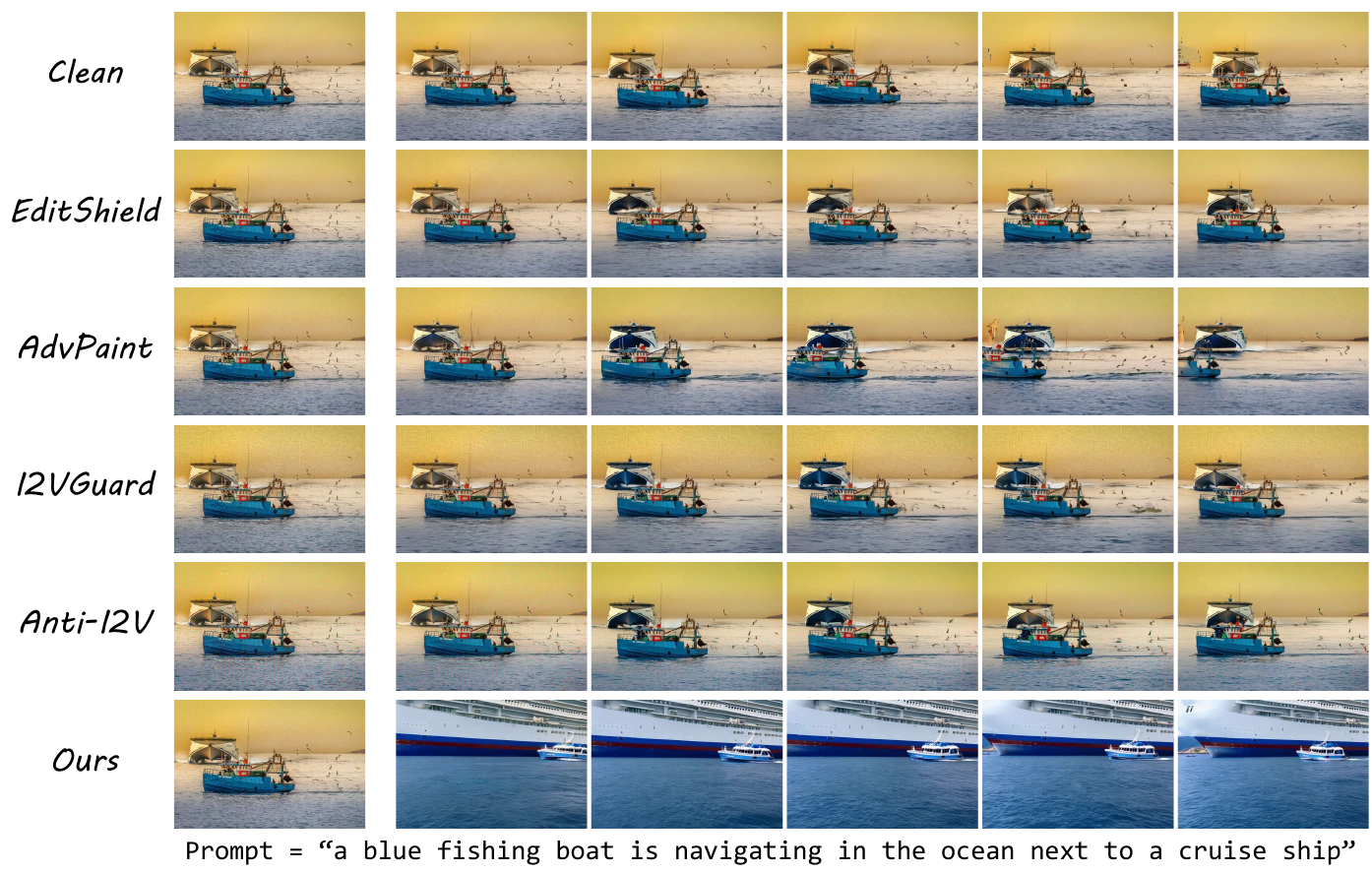}
    \vspace{-0.6cm}
    \caption{Qualitative comparison with baselines on CogVideoX~\cite{cogvideox}.}

    \label{fig/cog_wb_2}
\end{figure*}

\begin{figure*}[t]
\centering
    \includegraphics[width=\linewidth]{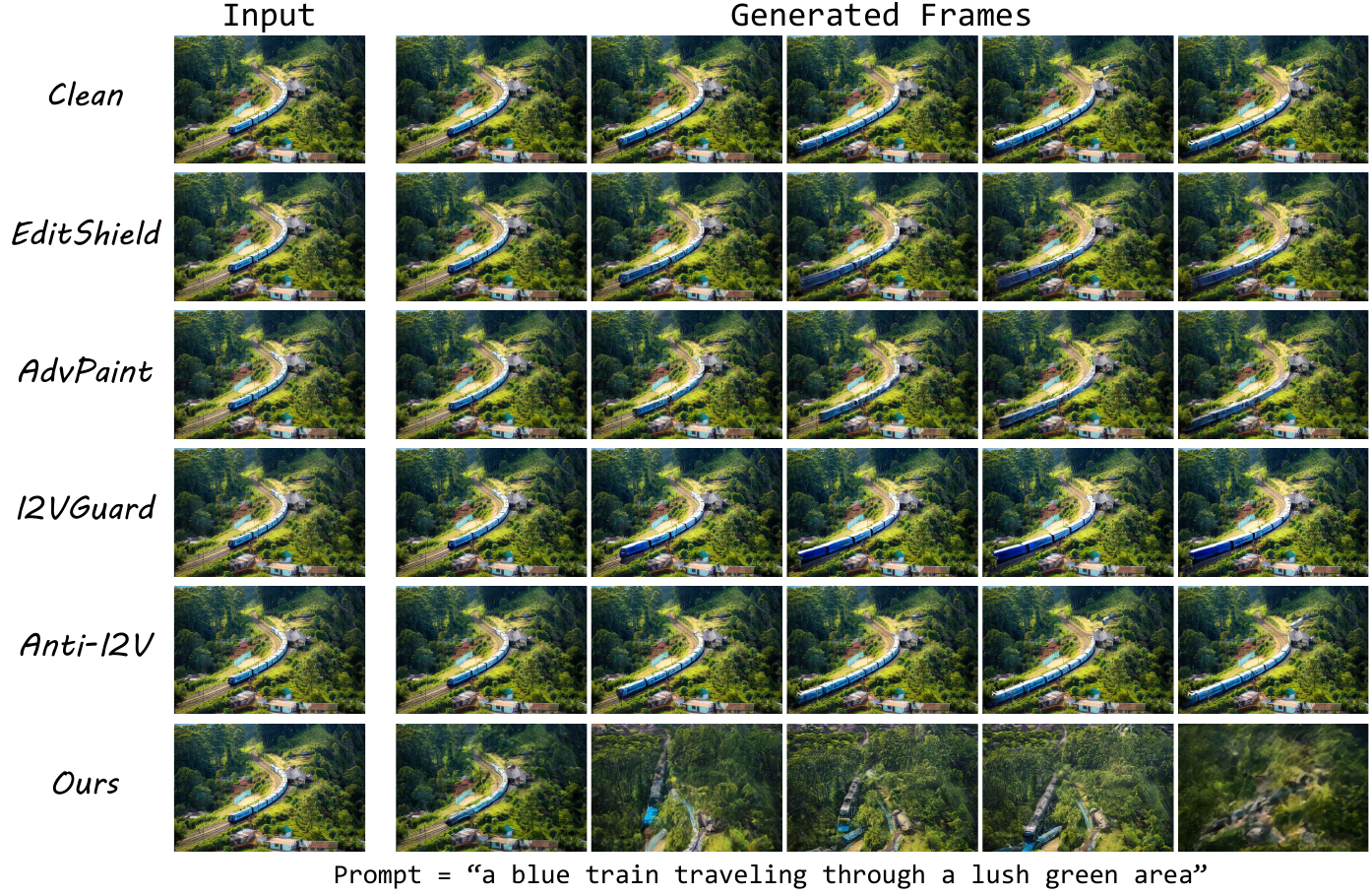}
    
    \includegraphics[width=\linewidth]{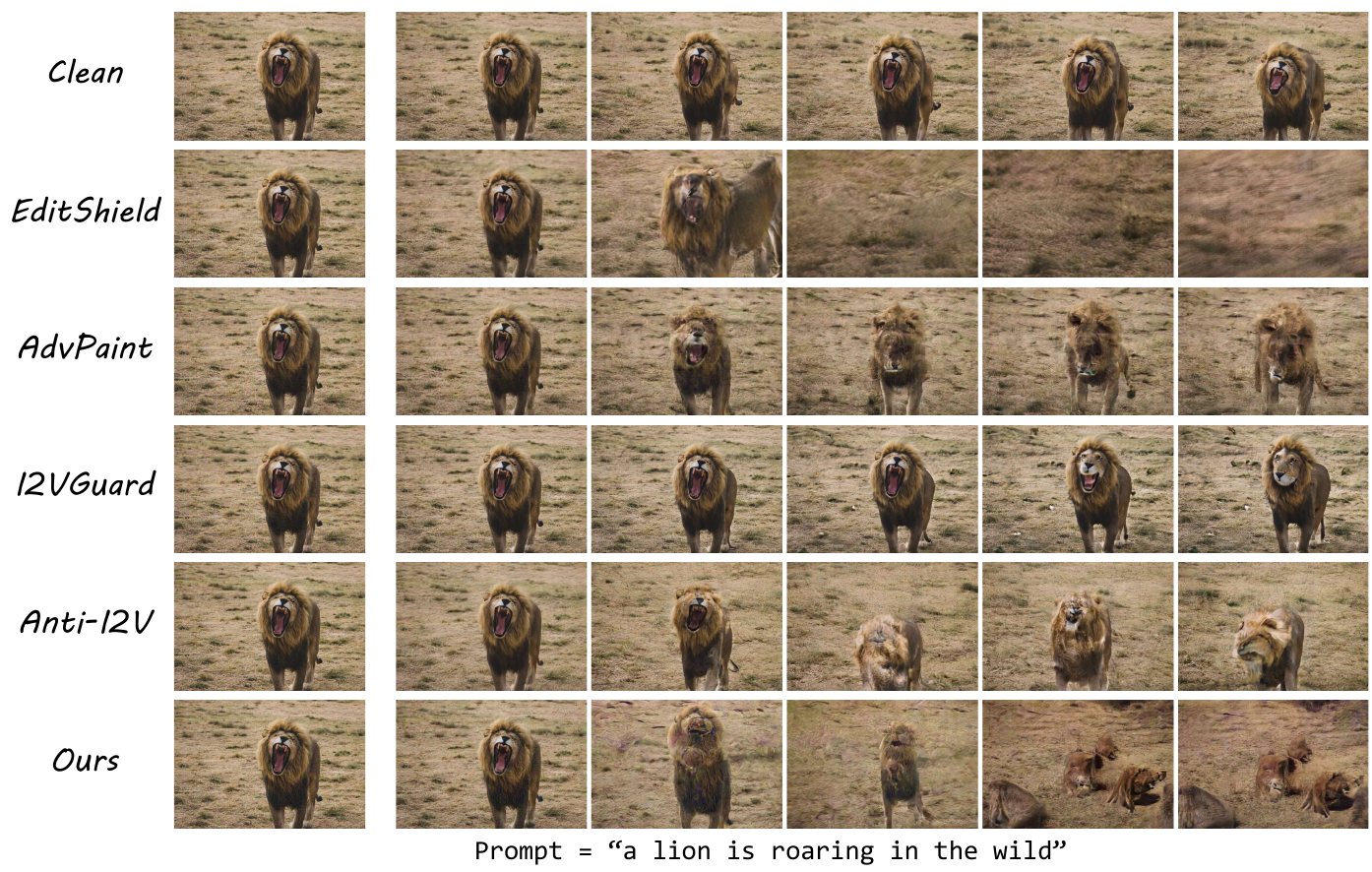}
    \vspace{-0.6cm}
    \caption{Qualitative comparison with baselines on LTX-Video~\cite{ltx}.}

    \label{fig/ltx_wb_2}
\end{figure*}

\begin{figure*}[t]
\centering
    \includegraphics[width=\linewidth]{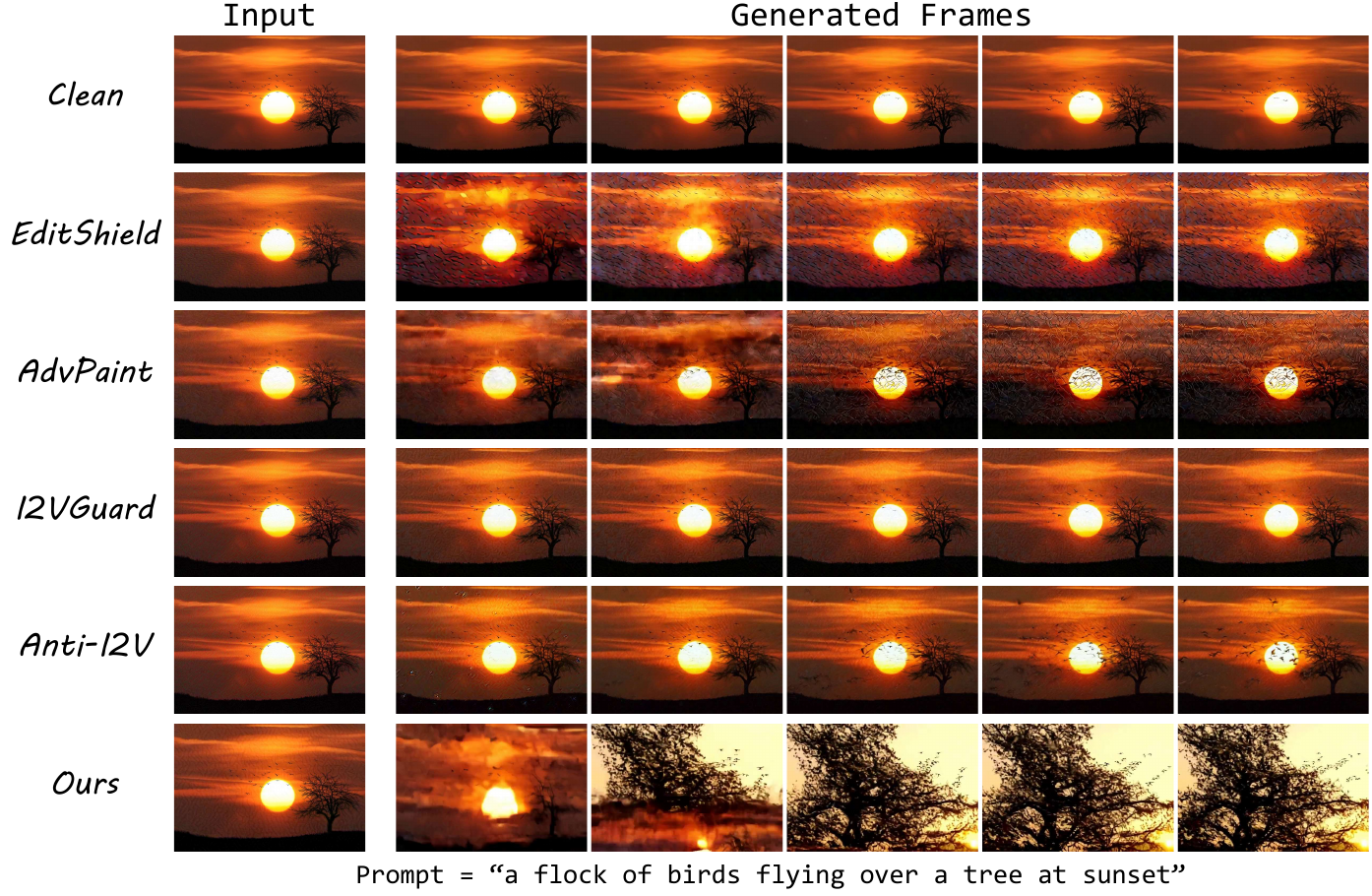}
    
    \includegraphics[width=\linewidth]{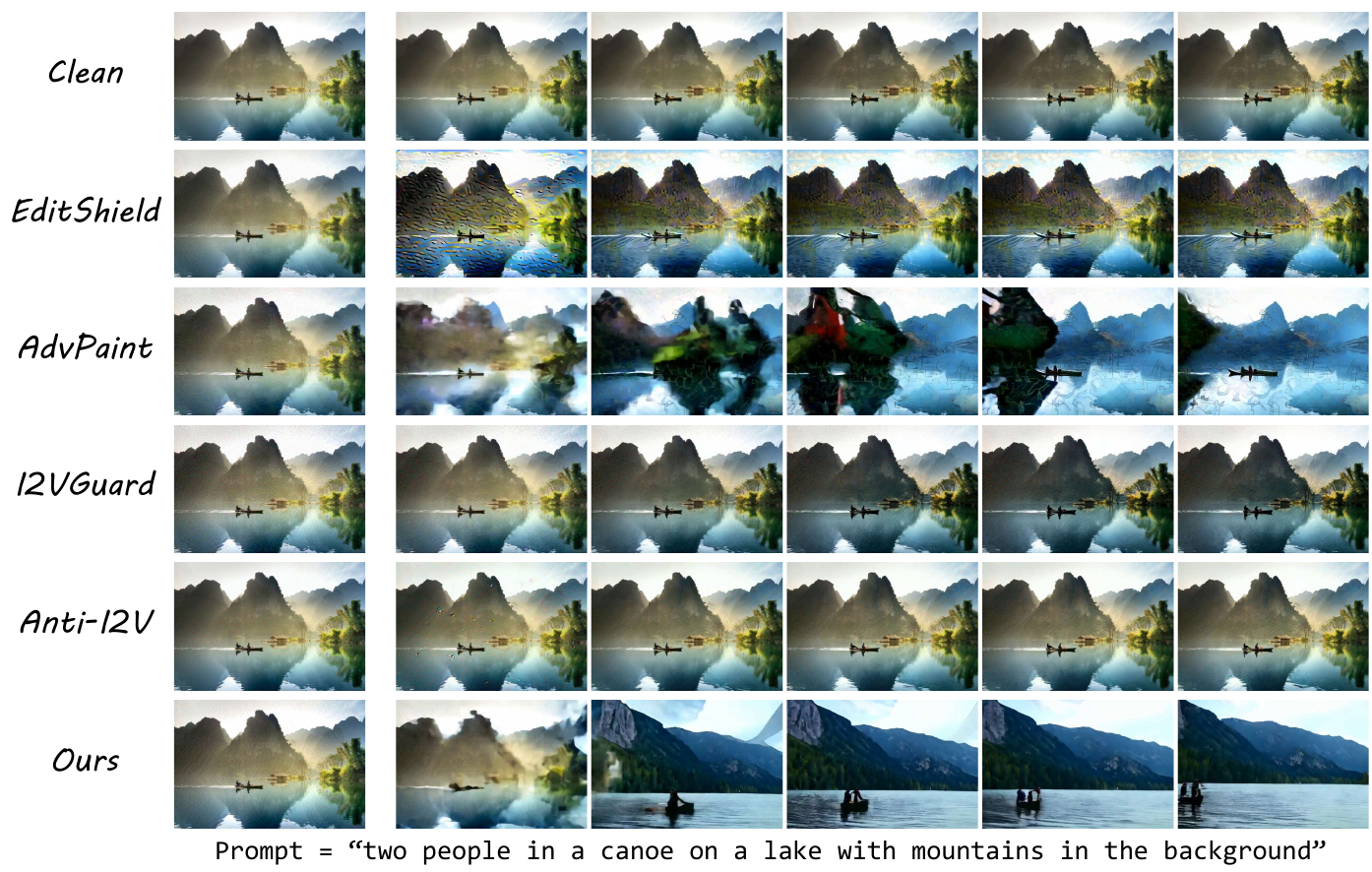}
    \vspace{-0.6cm}
    \caption{Qualitative comparison with baselines on OpenSora 1.2~\cite{opensora}.}

    \label{fig/sora_wb_2}
\end{figure*}

Although portrait misuse is a major concern in I2V generation, unauthorized reuse is not limited to human-related images. Non-human visual content, including copyrighted or otherwise protected images, can also be exploited for unauthorized video generation.
To examine this broader applicability, we present qualitative results on diverse non-human images.
As shown in Figures~\ref{fig/cog_wb_2}, \ref{fig/ltx_wb_2}, and \ref{fig/sora_wb_2}, \method consistently disrupts source grounding across diverse object categories, indicating that its protection effect is not limited to human-related images. These results suggest that our method can also prevent unauthorized I2V generation for non-human visual content, including objects or images that may be subject to copyright or other ownership concerns.

\subsection{Black-box Results}
\begin{figure*}[t]
\centering
    \includegraphics[width=\linewidth]{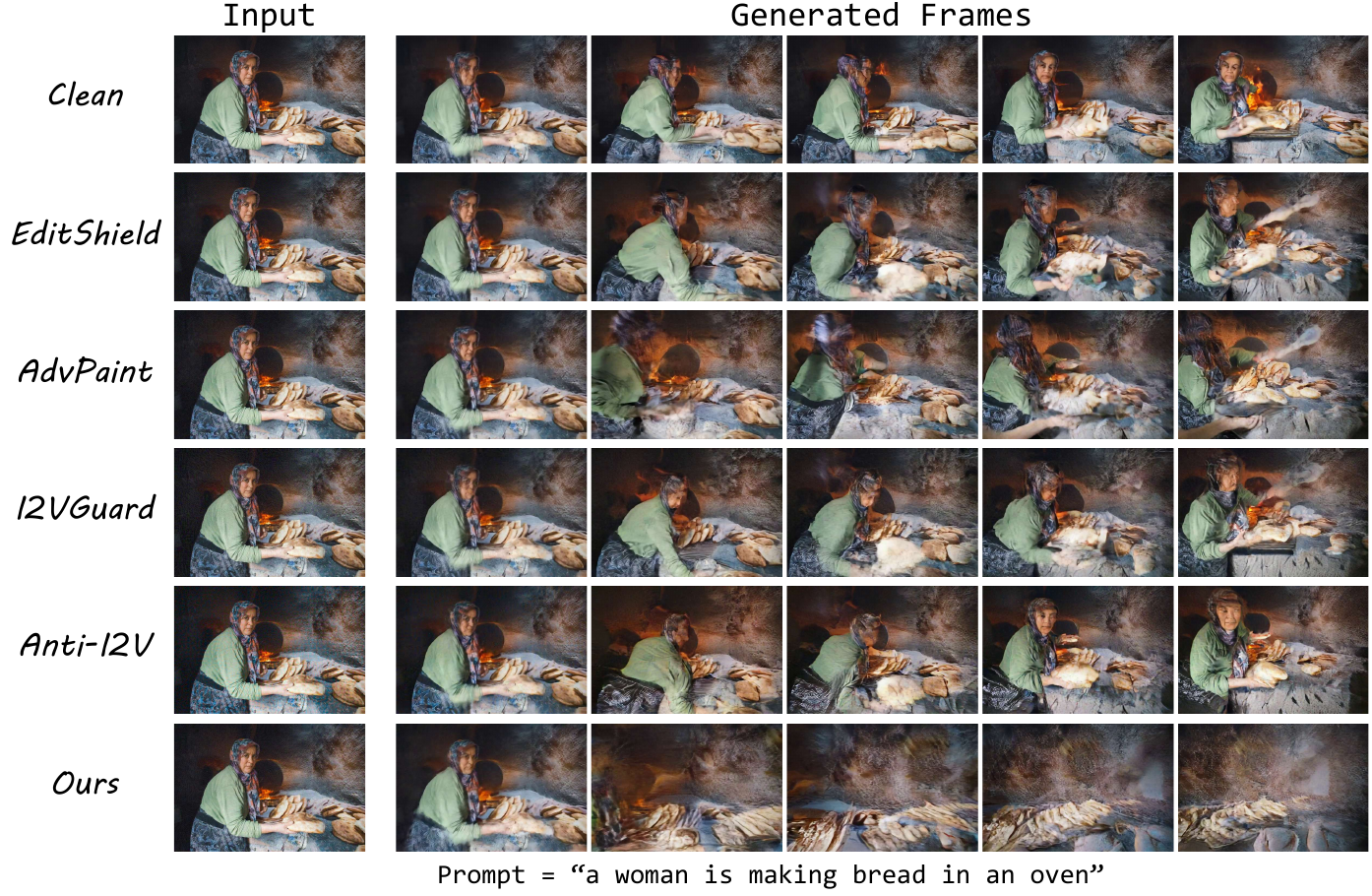}
    
    \includegraphics[width=\linewidth]{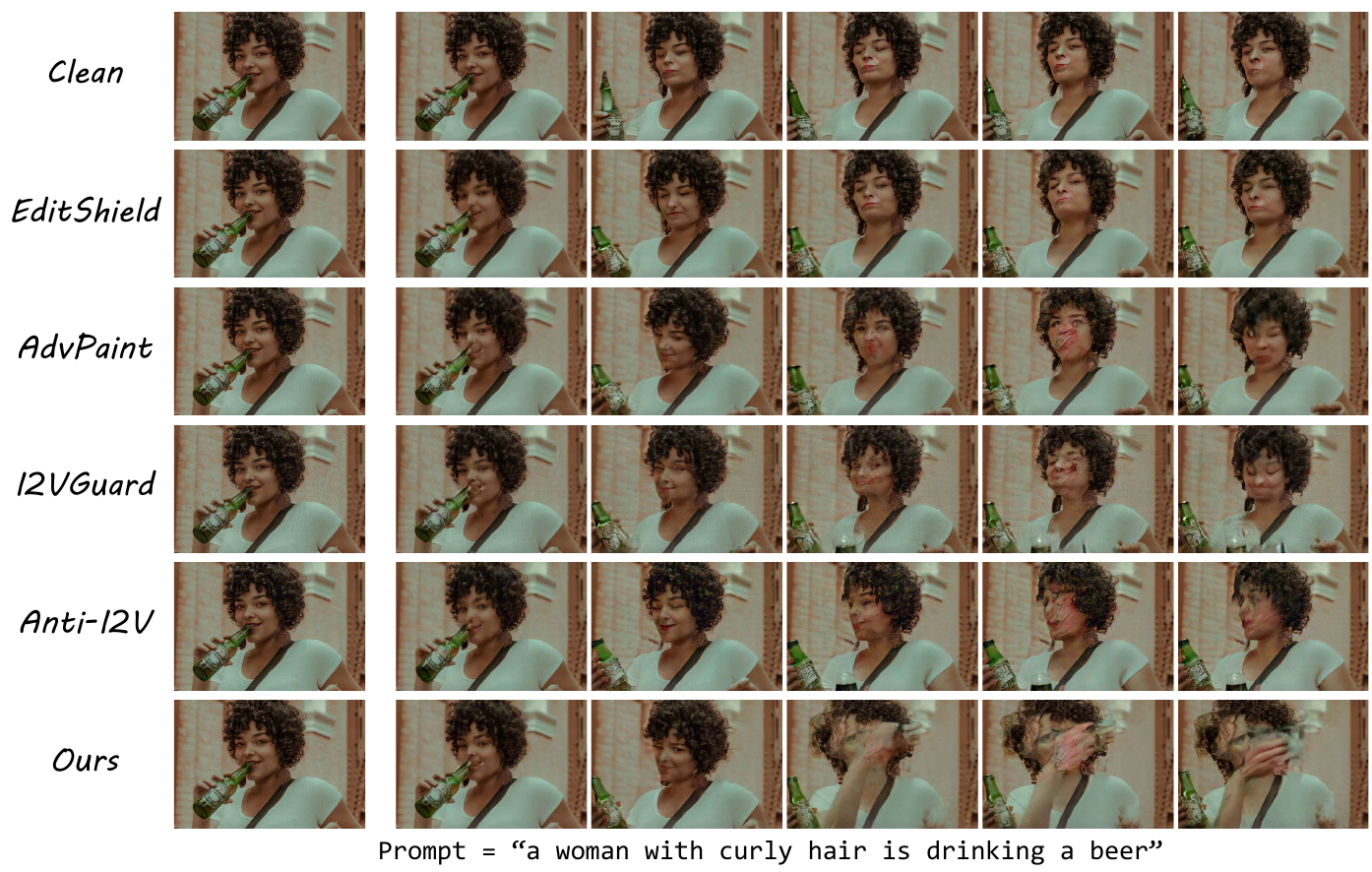}
    \vspace{-0.6cm}
    \caption{Qualitative results for black-box scenario on LTX-Video~\cite{ltx}. CogVideoX~\cite{cogvideox} is used for victim model.}

    \label{fig/ltx_bb}
\end{figure*}

\begin{figure*}[t]
\centering
    \includegraphics[width=\linewidth]{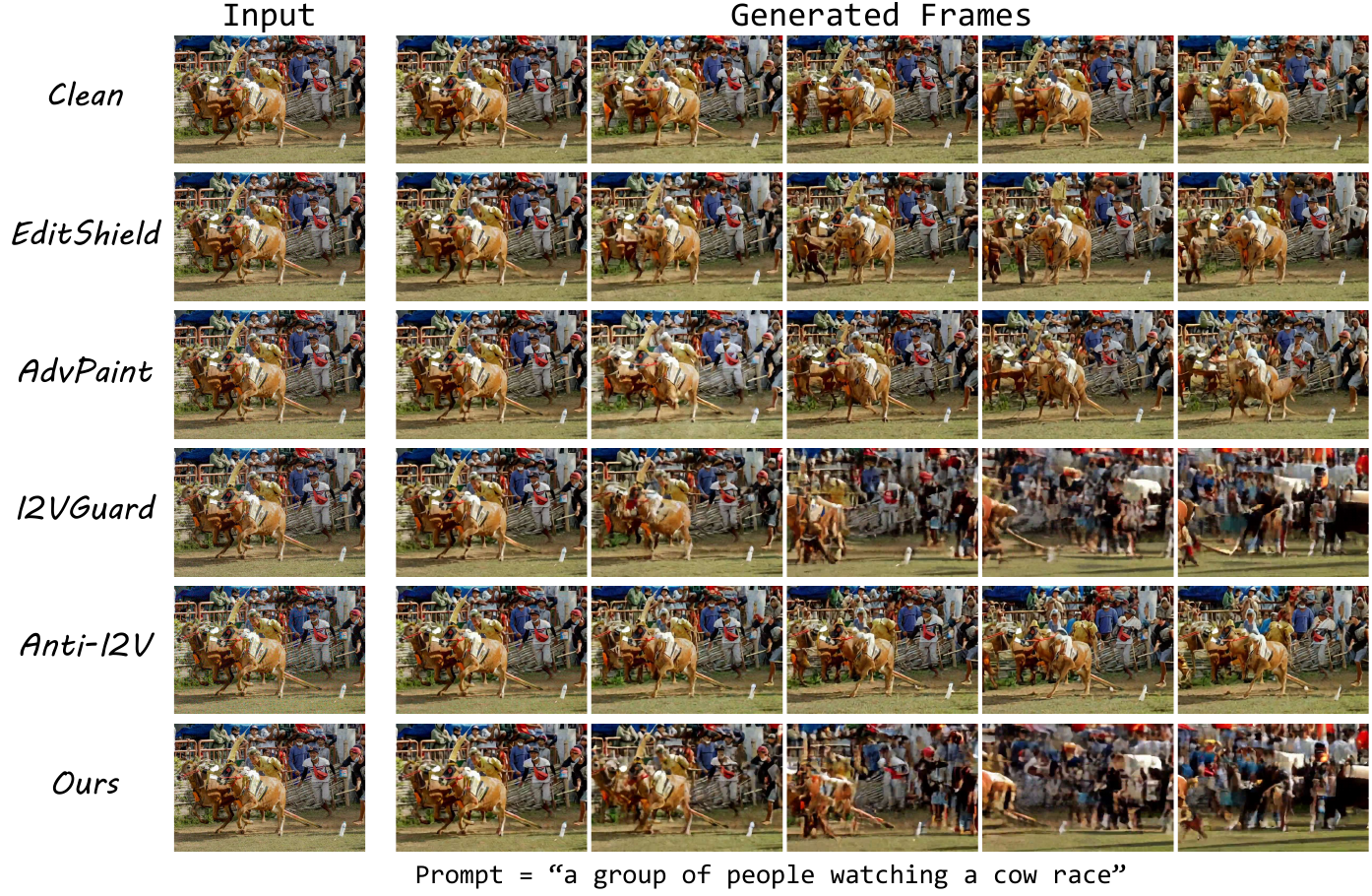}
    
    \includegraphics[width=\linewidth]{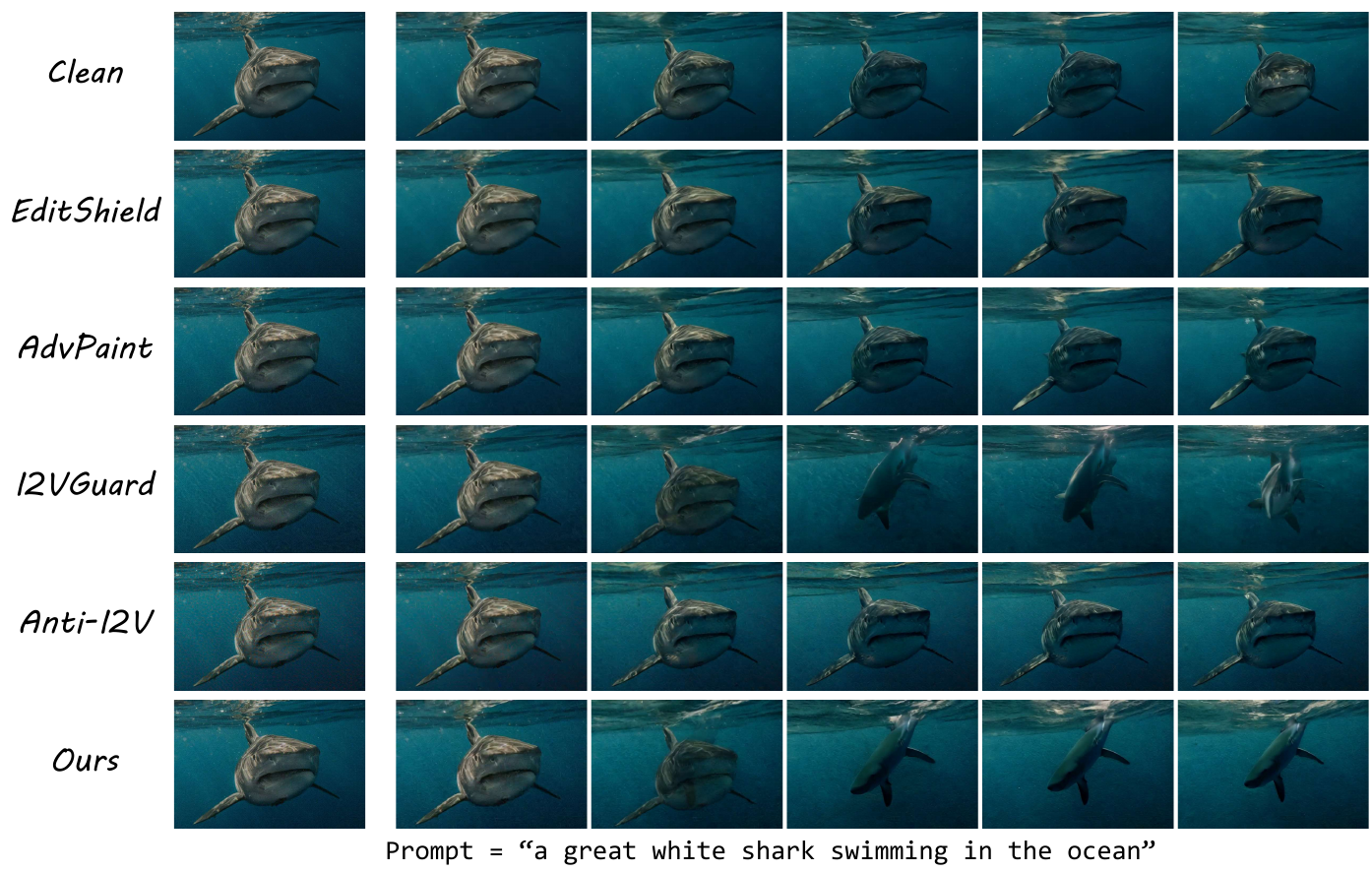}
    \vspace{-0.6cm}
    \caption{Qualitative results for black-box scenario on OpenSora 1.2~\cite{opensora}. CogVideoX~\cite{cogvideox} is used for victim model.}

    \label{fig/sora_bb}
\end{figure*}

We additionally provide qualitative comparisons in the black-box setting, where CogVideoX~\cite{cogvideox} is used as the victim model.
As shown in Figures~\ref{fig/ltx_bb} and~\ref{fig/sora_bb}, the degree of source grounding detachment is reduced compared with the white-box setting, reflecting the architectural gap between the victim and target I2V models.
Even under this challenging transfer setting, our method induces stronger generation failure and fidelity degradation than the baselines.
Together with Table~\ref{tab:black}, these results show that \method offers stronger black-box protection across both quantitative and qualitative evaluations.

% \subsection{Qualitative Ablation for Loss Functions}
% \label{sec:loss_abl_qual}
% The quantitative effects of the proposed {\svloss} loss and {\vvloss} video reinforcement loss are reported in Table~\ref{tab:loss-video} in the manuscript.
% To further examine how each loss influences the generated videos, we additionally conduct a qualitative ablation study in Figure~\ref{fig/loss_abl}.
% When only the {\svloss} loss is used, the generated videos often exhibit changes in background details or overall visual tone, suggesting that the video latents are encouraged to rely less on the conditioning image, but not enough to fully break source-image grounding.
% On the other hand, using only the {\vvloss}loss tends to produce stronger temporal degradation, yet the source content is still not fully detached from the generated video.
% When both losses are jointly applied, however, the generated videos no longer remain visually grounded in the source image, providing successful protection against unauthorized I2V generation.
% These results indicate that the advantage of jointly optimizing both losses is consistently observed in both quantitative and qualitative evaluations.

\section{Failure Cases}
\label{sec:fail}
Illustration for several failure cases of our method is in Figure~\ref{fig/fail}, denoted as \emph{Ours}.
In these examples, the generated videos do not fully break source-image grounding, but instead exhibit only partial deviations, such as changes in background details or shifts in the overall visual tone.
In other words, the source image is still partially reflected in the generated video, even though the resulting outputs are not fully faithful reproductions of the source content.

From the perspective of I2V immunization, these cases can still provide partial protection, since the generated videos no longer preserve the source image in a fully reliable manner.
However, they fall short of our intended objective of preventing the source visual content from serving as a stable grounding signal throughout generation.
% Interestingly, such failure patterns are qualitatively similar to the results obtained when only the {\svloss} loss is used as shown in Figure~\ref{fig/qual_loss}).
More discussion is provided in Appendix~\ref{sec:limit}

\section{Limitations and Discussion} 
\label{sec:limit}
\begin{figure*}[t]
\centering
    \includegraphics[width=\linewidth]{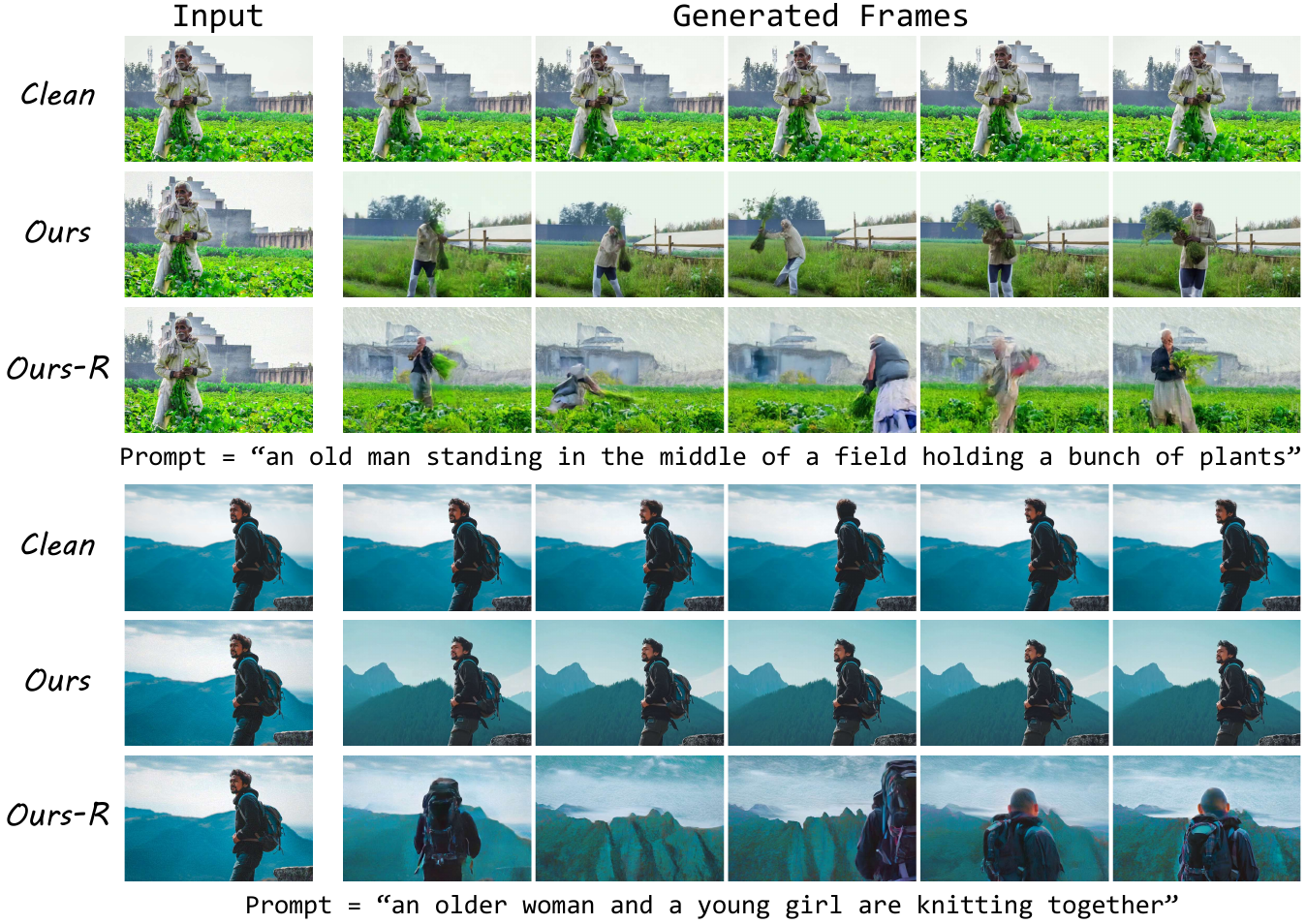}
    \vspace{-0.6cm}
    \caption{Failure case visualization. \emph{Ours} and \emph{Ours-R} denote the proposed method without and with loss reweighting, respectively.}

    \label{fig/fail}
\end{figure*}

Despite the overall effectiveness of our method, several limitations remain.
First, as discussed in Appendix~\ref{sec:fail}, our method still exhibits failure cases in which the generated videos deviate from the source image only partially.
% , for example through changes in background details or overall visual tone, rather than through a full disruption of source grounding.
% Notably, these cases are qualitatively similar to those observed when only the source-video decoupling loss is used.
This failure case indicates insufficient source grounding detachment.
As shown by \emph{Ours-R} in Fig.~\ref{fig/fail}, increasing the weight of the {\svloss} term can further suppress source grounding and produce stronger detachment when needed.

Second, our method still requires iterative optimization over a large I2V model to obtain a perturbation for each source image, which leads to substantial computational cost in practice.
Since modern diffusion-based I2V models are considerably larger and more expensive than typical image-generation models, this optimization process can be time-consuming and may limit practical scalability.
We believe this highlights an important direction for future research: extending more efficient protection paradigms, such as universal adversarial perturbations~\cite{uap} or learned immunization models~\cite{diffvax}, to the I2V setting.
Such approaches could improve the practicality of image protection by reducing the cost of per-image optimization while maintaining strong immunization performance.

\section{Broader Impact}
\label{app:broader}
\method serves as a proactive image immunization technique against unauthorized and malicious I2V generation.
Rather than merely degrading video quality, our method aims to make generated videos visually detached from the source image, reducing the extent to which protected visual content is preserved or reflected in the output.
% By effectively operating across I2V models with diverse architectures, \method can help prevent the misuse of personal or copyrighted images before source-related videos are generated.
This is particularly relevant in light of increasing risks such as deepfakes, fabricated news, identity misuse, and copyright infringement, where a single source image can be exploited to synthesize misleading or damaging video content.
In this sense, \method provides a preventive protection mechanism that helps users retain control over how their visual content is used in generative video applications.

However, such protection may also introduce potential risks if misused.
For example, if the method is applied to images that are legitimately licensed or authorized for video generation, it may interfere with intended creative or commercial uses.
% A malicious actor could also distribute immunized versions of images without consent, preventing others from generating videos from images they are otherwise allowed to use.
Therefore, image immunization techniques should be deployed with careful consideration of ownership, consent, and usage rights.
We believe that responsible use of such methods requires clear user control and transparent disclosure when protected images are shared or distributed.

\section{Future Work}
In this paper, we proposed \method, a novel image immunization framework that introduces a source grounding detachment objective and effectively prevents diffusion-based image-to-video generation across diverse diffusion architectures without architecture-specific methodological tuning.
Our method outperforms prior approaches on consistency-related metrics that assess how faithfully generated videos preserve the source image, while qualitatively inducing stronger visual detachment from the protected source.
Moreover, \method demonstrates strong black-box transferability and can be further combined with purification-aware strategies, suggesting its compatibility with existing robustness-oriented defenses.

A remaining limitation is the optimization cost of image-specific perturbations, particularly for large-scale I2V models.
We believe a promising future direction is to improve scalability by developing universal adversarial perturbations or pretraining an efficient immunizer that produces protective perturbations in a feed-forward manner.
This could enable practical image protection at lower computational cost as diffusion-based I2V models continue to scale.

\section{Reproducibility}
% Our code will be publicly released if our paper is accepted.
To support reproducibility, we include the immunization code in the supplementary materials, provide pseudo-code in Algorithm~\ref{alg:sigguard}, and report implementation details and hyperparameters (\ie, timestep set and layer selection) in Appendix~\ref{sec:imple}.

We also carefully consider the reproducibility of baseline implementations. 
Unfortunately, the official code for representative I2V immunization methods, including I2VGuard~\cite{i2vguard} and Anti-I2V~\cite{antii2v}, is not publicly available. 
To ensure an objective and reliable evaluation of \method against these key baselines, we contacted the original authors for implementation details. 
Although they were unable to release the code due to internal company policy, they kindly shared important implementation details, including hyperparameters, timestep sampling, perturbation optimization strategy, and selected layers. 
Based on this detailed information, we reimplemented the baselines as faithfully as possible. 
% \note{Upon acceptance, we will also release our reimplemented baseline code together with the code for \method.}

%%%%%%%%%%%%%%%%%%%%%%%%%%      cogvideox whitebox      %%%%%%%%%%%%%%%%%%%%%%%%%%%%%%%%%%%%%%%%%%%%%%%%%%%

%%%%%%%%%%%%%%%%%%%%%%%%%%%%%%%%     ltx   whitebox     %%%%%%%%%%%%%%%%%%%%%%%%%%%%%%%%%%%%%%%%%%%%%

%%%%%%%%%%%%%%%%%%%%%%%%%%%%%%%%     opensora   whitebox     %%%%%%%%%%%%%%%%%%%%%%%%%%%%%%%%%%%%%%%%%%%%%

%%%%%%%%%%%%%%%%%%%%%%%%%%%%%%%%     non human     %%%%%%%%%%%%%%%%%%%%%%%%%%%%%%%%%%%%%%%%%%%%%

%%%%%%%%%%%%%%%%%%%%%%%%%%%%%%%%     black-box    %%%%%%%%%%%%%%%%%%%%%%%%%%%%%%%%%%%%%%%%%%%%%

%%%%%%%%%%%%%%%%%%%%%%%%%%%%%%%%    loss ablation     %%%%%%%%%%%%%%%%%%%%%%%%%%%%%%%%%%%%%%%%%%%%%

% \begin{figure*}[t]
% \centering
%     \includegraphics[width=\linewidth]{fig/abl_1.pdf}
    
%     \includegraphics[width=\linewidth]{fig/abl_2.pdf}
%     \vspace{-0.6cm}
%     \caption{Qualitative ablation of the proposed loss functions. \emph{Loss 1} and \emph{Loss 2} denote \emph{source-like decoupling} loss and \emph{source-deviated video enhancement} loss, respectively.} 

%     \label{fig/loss_abl}
% \end{figure*}

%%%%%%%%%%%%%%%%%%%%%%%%%%%%%%%%     failure     %%%%%%%%%%%%%%%%%%%%%%%%%%%%%%%%%%%%%%%%%%%%%

\clearpage
\section*{NeurIPS Paper Checklist}

%%% BEGIN INSTRUCTIONS %%%
The checklist is designed to encourage best practices for responsible machine learning research, addressing issues of reproducibility, transparency, research ethics, and societal impact. Do not remove the checklist: {\bf The papers not including the checklist will be desk rejected.} The checklist should follow the references and follow the (optional) supplemental material.  The checklist does NOT count towards the page
limit. 

Please read the checklist guidelines carefully for information on how to answer these questions. For each question in the checklist:
\begin{itemize}
    \item You should answer \answerYes{}, \answerNo{}, or \answerNA{}.
    \item \answerNA{} means either that the question is Not Applicable for that particular paper or the relevant information is Not Available.
    \item Please provide a short (1--2 sentence) justification right after your answer (even for \answerNA). 
   % \item {\bf The papers not including the checklist will be desk rejected.}
\end{itemize}

{\bf The checklist answers are an integral part of your paper submission.} They are visible to the reviewers, area chairs, senior area chairs, and ethics reviewers. You will also be asked to include it (after eventual revisions) with the final version of your paper, and its final version will be published with the paper.

The reviewers of your paper will be asked to use the checklist as one of the factors in their evaluation. While \answerYes{} is generally preferable to \answerNo{}, it is perfectly acceptable to answer \answerNo{} provided a proper justification is given (e.g., error bars are not reported because it would be too computationally expensive'' or ``we were unable to find the license for the dataset we used''). In general, answering \answerNo{} or \answerNA{} is not grounds for rejection. While the questions are phrased in a binary way, we acknowledge that the true answer is often more nuanced, so please just use your best judgment and write a justification to elaborate. All supporting evidence can appear either in the main paper or the supplemental material, provided in appendix. If you answer \answerYes{} to a question, in the justification please point to the section(s) where related material for the question can be found.

IMPORTANT, please:
\begin{itemize}
    \item {\bf Delete this instruction block, but keep the section heading ``NeurIPS Paper Checklist"},
    \item  {\bf Keep the checklist subsection headings, questions/answers and guidelines below.}
    \item {\bf Do not modify the questions and only use the provided macros for your answers}.
\end{itemize}

%%% END INSTRUCTIONS %%%

\begin{enumerate}

\item {\bf Claims}
    \item[] Question: Do the main claims made in the abstract and introduction accurately reflect the paper's contributions and scope?
    \item[] Answer: \answerYes{} % Replace by \answerYes{}, \answerNo{}, or \answerNA{}.
    \item[] Justification: In both the abstract and the introduction, we claim our scope as a image immunization against unauthorized image-to-video generation. We are the first to propose source grounding detachment objective to make the generated videos do not faithfully reflect visual content of the source image. 

\item {\bf Limitations}
    \item[] Question: Does the paper discuss the limitations of the work performed by the authors?
    \item[] Answer: \answerYes{} % Replace by \answerYes{}, \answerNo{}, or \answerNA{}.
    \item[] Justification: We provide limitation and discussion in Appendix~\ref{sec:limit}

\item {\bf Theory assumptions and proofs}
    \item[] Question: For each theoretical result, does the paper provide the full set of assumptions and a complete (and correct) proof?
    \item[] Answer: \answerNA{} % Replace by \answerYes{}, \answerNo{}, or \answerNA{}.
    \item[] Justification: Our paper does not include theoretical results.

    \item {\bf Experimental result reproducibility}
    \item[] Question: Does the paper fully disclose all the information needed to reproduce the main experimental results of the paper to the extent that it affects the main claims and/or conclusions of the paper (regardless of whether the code and data are provided or not)?
    \item[] Answer: \answerYes{} % Replace by \answerYes{}, \answerNo{}, or \answerNA{}.
    \item[] Justification: We provide the detailed loss formulations, the complete pseudo-code in Algorithm~\ref{alg:sigguard} to ensure algorithmic reproducibility, targeted component of each model~\ref{app:model}, and code in supplementary materials.

\item {\bf Open access to data and code}
    \item[] Question: Does the paper provide open access to the data and code, with sufficient instructions to faithfully reproduce the main experimental results, as described in supplemental material?
    \item[] Answer: \answerYes{} % Replace by \answerYes{}, \answerNo{}, or \answerNA{}.
    \item[] Justification: We provide our training code and instruction in supplemental material.

\item {\bf Experimental setting/details}
    \item[] Question: Does the paper specify all the training and test details (e.g., data splits, hyperparameters, how they were chosen, type of optimizer) necessary to understand the results?
    \item[] Answer: \answerYes{} % Replace by \answerYes{}, \answerNo{}, or \answerNA{}.
    \item[] Justification: We provide experiment details in Section~\ref{sec:exp_detail} including threat models, datasets, and hyperparameters.

\item {\bf Experiment statistical significance}
    \item[] Question: Does the paper report error bars suitably and correctly defined or other appropriate information about the statistical significance of the experiments?
    \item[] Answer: \answerNo{} % Replace by \answerYes{}, \answerNo{}, or \answerNA{}.
    \item[] Justification: The paper includes extensive quantitative experiments, but the reported tables and figures do not provide error bars, confidence intervals, or statistical significance tests. The main claims are supported by consistent trends across multiple I2V models, evaluation metrics, and ablation studies.

\item {\bf Experiments compute resources}
    \item[] Question: For each experiment, does the paper provide sufficient information on the computer resources (type of compute workers, memory, time of execution) needed to reproduce the experiments?
    \item[] Answer: \answerYes{} % Replace by \answerYes{}, \answerNo{}, or \answerNA{}.
    \item[] Justification: We mentioned GPU memory and time-cost in Appendix~\ref{sec:exp_detail} and Table~\ref{tab:efficiency}.
    
\item {\bf Code of ethics}
    \item[] Question: Does the research conducted in the paper conform, in every respect, with the NeurIPS Code of Ethics \url{https://neurips.cc/public/EthicsGuidelines}?
    \item[] Answer: \answerYes{} % Replace by \answerYes{}, \answerNo{}, or \answerNA{}.
    \item[] Justification: We have reviewed the NeurIPS Code of Ethics and confirm that our research conforms to its principles in all respects.

\item {\bf Broader impacts}
    \item[] Question: Does the paper discuss both potential positive societal impacts and negative societal impacts of the work performed?
    \item[] Answer: \answerYes{} % Replace by \answerYes{}, \answerNo{}, or \answerNA{}.
    \item[] Justification: We mentioned broader impacts of the proposed method in Appendix~\ref{app:broader}.
    
\item {\bf Safeguards}
    \item[] Question: Does the paper describe safeguards that have been put in place for responsible release of data or models that have a high risk for misuse (e.g., pre-trained language models, image generators, or scraped datasets)?
    \item[] Answer: \answerNA{} % Replace by \answerYes{}, \answerNo{}, or \answerNA{}.
    \item[] Justification: Our work focuses on preventing the misuse of Image-to-Video models and does not release any high-risk artifacts.

\item {\bf Licenses for existing assets}
    \item[] Question: Are the creators or original owners of assets (e.g., code, data, models), used in the paper, properly credited and are the license and terms of use explicitly mentioned and properly respected?
    \item[] Answer: \answerYes{} % Replace by \answerYes{}, \answerNo{}, or \answerNA{}.
    \item[] Justification: We mentioned and cited the original owners of all assets we used in this paper.

\item {\bf New assets}
    \item[] Question: Are new assets introduced in the paper well documented and is the documentation provided alongside the assets?
    \item[] Answer: \answerYes{} % Replace by \answerYes{}, \answerNo{}, or \answerNA{}.
    \item[] Justification: We include the details of the code and the dataset in the supplementary.

\item {\bf Crowdsourcing and research with human subjects}
    \item[] Question: For crowdsourcing experiments and research with human subjects, does the paper include the full text of instructions given to participants and screenshots, if applicable, as well as details about compensation (if any)? 
    \item[] Answer: \answerYes{} % Replace by \answerYes{}, \answerNo{}, or \answerNA{}.
    \item[] Justification: We include detailed description about human evaluation in Appendix~\ref{app:human} and screenshots in Figure~\ref{fig/human}.

\item {\bf Institutional review board (IRB) approvals or equivalent for research with human subjects}
    \item[] Question: Does the paper describe potential risks incurred by study participants, whether such risks were disclosed to the subjects, and whether Institutional Review Board (IRB) approvals (or an equivalent approval/review based on the requirements of your country or institution) were obtained?
    \item[] Answer: \answerNo{} % Replace by \answerYes{}, \answerNo{}, or \answerNA{}.
    \item[] Justification: The paper includes a human evaluation with crowdworkers, but does not report IRB approval. The evaluation involved only visual rating/preference tasks on generated videos and did not collect sensitive personal information from participants.

\item {\bf Declaration of LLM usage}
    \item[] Question: Does the paper describe the usage of LLMs if it is an important, original, or non-standard component of the core methods in this research? Note that if the LLM is used only for writing, editing, or formatting purposes and does \emph{not} impact the core methodology, scientific rigor, or originality of the research, declaration is not required.
    %this research? 
    \item[] Answer: \answerNA{} % Replace by \answerYes{}, \answerNo{}, or \answerNA{}.
    \item[] Justification: We do not use LLM in this work.

\end{enumerate}
\end{document}